%% file: main.tex
\documentclass{article}

\PassOptionsToPackage{numbers, compress}{natbib}

\usepackage[main, final]{neurips_2026}

\usepackage[utf8]{inputenc}
\usepackage[T1]{fontenc}
\usepackage{hyperref}
\hypersetup{
    colorlinks=true,
    linkcolor=black, 
    citecolor=black, 
    urlcolor=blue   
}
\usepackage{url}
\usepackage{booktabs}
\usepackage{amsfonts}
\usepackage{microtype}
\usepackage{xcolor}
\usepackage{tikz}

\usepackage{amsmath, amssymb}
\usepackage{amsthm}
\usepackage{algorithm}
\usepackage{algpseudocode}
\usepackage{wrapfig}
\usepackage{enumitem}
\usepackage{bm}
\usepackage{caption}
\usepackage{xpatch}

\theoremstyle{plain}
\newtheorem{theorem}{Theorem}
\newtheorem{corollary}{Corollary}
\newtheorem{proposition}{Proposition}
\newtheorem{lemma}{Lemma}
\theoremstyle{definition}
\newtheorem{definition}{Definition}
\newtheorem{remark}{Remark}
\usepackage{cancel}
\usepackage[capitalise]{cleveref}
\usepackage{mathtools}
\usepackage{nccmath}
\usepackage{accents}
\usepackage{empheq}
\usetikzlibrary{backgrounds, arrows.meta, calc}
\makeatletter
\def\theHALG@line{\thealgorithm.\arabic{ALG@line}}
\DeclareRobustCommand{\cev}[1]{
  {\mathpalette\do@cev{#1}}
}
\newcommand{\do@cev}[2]{
  \vbox{\offinterlineskip
    \sbox\z@{$\m@th#1 x$}
    \ialign{##\cr
      \hidewidth\reflectbox{$\m@th#1\vec{}\mkern4mu$}\hidewidth\cr
      \noalign{\kern-\ht\z@}
      $\m@th#1#2$\cr
    }
  }
}
\DeclareMathOperator*{\Log}{Log}
\DeclareMathOperator*{\Arg}{Arg}
\DeclareMathOperator*{\Tr}{Tr}
\DeclareMathOperator{\Cov}{Cov}
\DeclareMathOperator*{\minimize}{minimize}
\newcommand*{\bC}{\mathbb{C}}
\newcommand*{\bD}{\mathbb{D}}
\newcommand*{\bP}{\mathbb{P}}
\newcommand*{\bQ}{\mathbb{Q}}
\newcommand*{\bR}{\mathbb{R}}
\newcommand*{\cR}{\mathcal{R}}
\newcommand*{\cL}{\mathcal{L}}
\newcommand*{\cH}{\mathcal{H}}
\newcommand*{\cC}{\mathcal{C}}
\newcommand*{\cZ}{\mathcal{Z}}
\newcommand*{\cE}{\mathcal{E}}
\newcommand*{\cF}{\mathcal{F}}
\newcommand*{\cP}{\mathcal{P}}
\newcommand*{\cQ}{\mathcal{Q}}
\newcommand*{\cN}{\mathcal{N}}
\newcommand*{\cS}{\mathcal{S}}
\newcommand*{\bZ}{\mathbb{Z}}
\newcommand*{\bE}{\mathbb{E}}
\newcommand*{\rd}{\mathrm{d}}
\newcommand*{\sA}{\mathsf{A}}
\newcommand*{\sT}{\mathsf{T}}
\newcommand*{\vw}{\mathbf{w}}
\newcommand*{\vv}{\mathbf{v}}
\newcommand*{\vI}{\mathbf{I}}
\newcommand*{\vJ}{\mathbf{J}}
\newcommand*{\vp}{\mathbf{p}}
\newcommand*{\vx}{\mathbf{x}}
\newcommand*{\vy}{\mathbf{y}}
\newcommand*{\vz}{\mathbf{z}}
\newcommand*{\vzero}{\mathbf{0}}
\newcommand*{\vD}{\mathbf{D}}
\newcommand*{\KL}{\mathrm{KL}}
\newcommand*{\Aut}{\operatorname{Aut}}
\newcommand{\dd}{\mathop{}\!\mathrm{d}}
\renewcommand*{\ss}{{\mkern-1.2mu s}}
\newcommand*{\ssT}{{\scriptscriptstyle\mathsf{T}}}
\DeclareMathSymbol{\shortminus}{\mathord}{AMSa}{"39}
\newcommand*{\im}{\sqrt{\mspace{-1mu}\shortminus\mspace{-1.4mu}1}\mspace{1.2mu}}
\newcommand*{\QED}{\null\nobreak\hfill\ensuremath{\square}}%

\newcommand*{\ie}{\textit{i.e.}}
\newcommand*{\eg}{\textit{e.g.}}
\newcommand*{\one}{{\raisebox{.53pt}{\small\textcircled{\raisebox{-.23pt} {\hskip.06pt\scriptsize1}}}}}
\newcommand*{\two}{{\raisebox{.53pt}{\small\textcircled{\raisebox{-.2pt} {\hskip.1pt\scriptsize2}}}}}
\newcommand*{\three}{{\raisebox{.53pt}{\small\textcircled{\raisebox{-.25pt} {\hskip.15pt\scriptsize3}}}}}

\title{Complex Schr\"odinger Bridges}
\author{
  Dong-Sig Han\quad Tolga Birdal\\ Department of Computing\\
  Imperial College London
}

\begin{document}
\maketitle
\begin{abstract}
  Diffusion bridges have emerged as versatile frameworks for facilitating transformations between arbitrary data distributions. However, these approaches are hindered by an immense optimization space and its underlying high-order structures. This paper explores a variant of diffusion bridges for solving real-world problems in complex vector spaces governed by scaling and rotation laws, such as Hamiltonian structures. To address this, we introduce conformal optimal transport (COT), a complex-valued transportation framework designed for physical systems with geometric and dissipative dynamics under a geometric lifting. Building on this, we propose a new class of dynamical diffusion bridges, called complex Schr\"odinger bridges (CSBs), whose derived algorithms can handle diffusion-based tasks using a conformal geometric prior. For scalability, we propose image-to-image CSB (I$^2$CSB), a tractable diffusion-bridge technique with an analytic spiral posterior. Finally, we implement a physics-informed CSB algorithm within simulations of quantum electrodynamics, mimicking the logical behavior of quantum logic gates. Our extensive results validate the underlying theory while highlighting advantages of complex-valued diffusion. Our project page is available \href{https://circle-group.github.io/research/CSB}{here}.
\end{abstract}

\input{text/1_intro.tex}

\input{text/2_prelim.tex}

\input{text/3_csb.tex}

\input{text/4_algo.tex}

\input{text/5_expr.tex}

\input{text/6_work.tex}

\input{text/7_concl.tex}

\vspace{-1mm}
\section*{Acknowledgments}
The authors would like to thank the anonymous reviewers for their helpful discussion and comments. T. Birdal was supported by a UKRI Future Leaders Fellowship [grant number MR/Y018818/1]. The authors acknowledge support from the UK AI Research Resource (AIRR Isambard AI) through grant 0251-4584-0945-1 - TopoFound.

\medskip {
  \small
  \bibliographystyle{apalike}
  \bibliography{main.bib}
}

\clearpage\appendix
\vbox{
  \hsize\textwidth
  \linewidth\hsize
  \vskip 0.1in
  \vskip 0.08in
  \hrule height 3pt
  \vskip 0.218in
  \vskip -\parskip
  \centering
  {\large\bf Appendices for\par}
  \vskip -2pt
  {\Large\bf \@title\par}
  \vskip 0.258in
  \vskip -\parskip
  \hrule height 1pt
  \vskip 0.09in
}
\vspace{-2mm}
\section*{Abbreviation and Notation}
\vspace{-3mm}
\begingroup\centering
\tikzset{inner sep=0, outer sep=0}
\begin{tikzpicture}[tight background] 
  \begin{scope}
    \node[scale=0.85] at (0, 0) {
      \begin{tabular}{ll}
        \toprule
          Abbreviation&Expansion\\
        \midrule
          SB&Schr\"odinger Bridge\\
          SBP&Schr\"odinger Bridge Problem\\
          CSB&Complex Schr\"odinger Bridge\\
          I$^2$CSB&Image-to-Image CSB\\
          BM&Brownian motion\\
          OT&Optimal Transport\\
          EOT&Entropy-regularized OT\\
          COT&Conformal OT\\
          SOC&Stochastic Optimal Control\\
          SDE&Stochastic Differential Equation\\
          CVSDE&Complex-Valued SDE\\
          FBSDE&Forward--Backward SDE\\
          PDE&Partial Differential Equation\\
          MFG&Mean-Field Game\\
          CFT&Conformal Field Theory\\
          OU&Ornstein--Uhlenbeck\\
          KL&Kullback--Leibler\\
          IPF&Iterative Proportional Fitting\\
          TD&Temporal Difference\\
          FID&Fr\'echet Inception Distance\\
          RK4&Fourth-order Runge--Kutta\\
          CFM&Conditional Flow Matching\\
          EMD&Earth Mover's Distance\\
          cQED&circuit Quantum Electrodynamics\\
          JJ&Josephson Junction\\
        \bottomrule
      \end{tabular}};
  \end{scope}
  \begin{scope}[shift={(6.584651, -1.450075)}]
    \node[scale=0.85] at (0, 0) {
      \begin{tabular}{ll}
        \toprule
          Notation&Usage\\
        \midrule
          $t,s$&forward and reverse time indices\\
          $Z_t,\bar{Z}_\ss$&conjugate stochastic states\\
          $\vz_t,\bar{\vz}_\ss$&complex state realizations\\
          $\vx_t,\vv_t$&position and momentum\\
          $W_t,\bar{W}_s$&complex Brownian motions\\
          $\vw_t,\hat{\vw}_t$&real Brownian motions\\
          $\bC^d, \bR^{2d}$&complex and real state spaces\\
          $\cZ,\cS$&sample space, state domain\\
          $\Omega$&continuous-path space\\
          $\cP(\Omega)$&path-measure space\\
          $\bP,\bP_t$&path measure and marginal\\
          $\bQ_\vw,\bQ_W$&reference measures\\
          $\bQ^\gamma$&conformal reference measure\\
          $\rho_t$&time density function\\
          $\mu,\nu$&endpoint marginals $(\rho_{\scriptscriptstyle 0}, \rho_{\scriptscriptstyle \mkern-1.2mu T}\mkern-1.2mu)$\\
          $\pi\in\Pi(\mu,\nu)$&coupling and coupling set\\
          $\sigma_t$&diffusion coefficient\\
          $h_\gamma$&conformal map\\
          $c_\gamma,\hat{c}_\gamma$&conformal coefficients\\
          $L_\gamma$&conformal geodesic length\\
          $H,\hat{H}$&MF Hamiltonian and its operator\\
          $\psi,\varphi$&log-Schr\"odinger potentials\\
          $\Psi,\bar{\Psi}$&conjugate wavefunctions\\
          $\Phi,\hat{\Phi}$&complex Hopf--Cole transform\\
          $\nabla_{\!\vz},\!\nabla_{\!\bar{\vz}}$&Wirtinger gradients\\
          $f,F$&base drift, MF cost\\
          $Y_t,\widehat{Y}_t$&CVSDE log-potentials\\
          $\chi,\eta$&parameterized potentials\\
          $v_t,\hat{v}_t,\check{v}_t,\bar{v}_t$&scaled potential gradients\\
          $\cE^{+}_{T-\tau},\cE^{-}_{\tau}$&accumulated IPF errors\\
          $\cC\cN$&complex normal distribution\\
          $\bar{\mu}_s,\Sigma_s$&posterior mean and covariance\\
          $\omega_s^2,\hat{\omega}_s^2$&integrated diffusion variances\\
        \bottomrule
      \end{tabular}};
  \end{scope}
 \end{tikzpicture}
\par\endgroup

\input{text/A_cop.tex}

\input{text/B_cmap.tex}

\input{text/C_proof.tex}

\input{text/D_detail.tex}

\input{text/E_expr.tex}

\end{document}

%% file: text/1_intro.tex
\begin{wrapfigure}[14]{r}{0.3498\textwidth}
  \vskip-30pt  
  \centering
  \tikzset{inner sep=0, outer sep=0}
  \begin{tikzpicture}[tight background]
    \begin{scope}
      \node (B) at (0,-0.01) {\includegraphics[height=120pt]{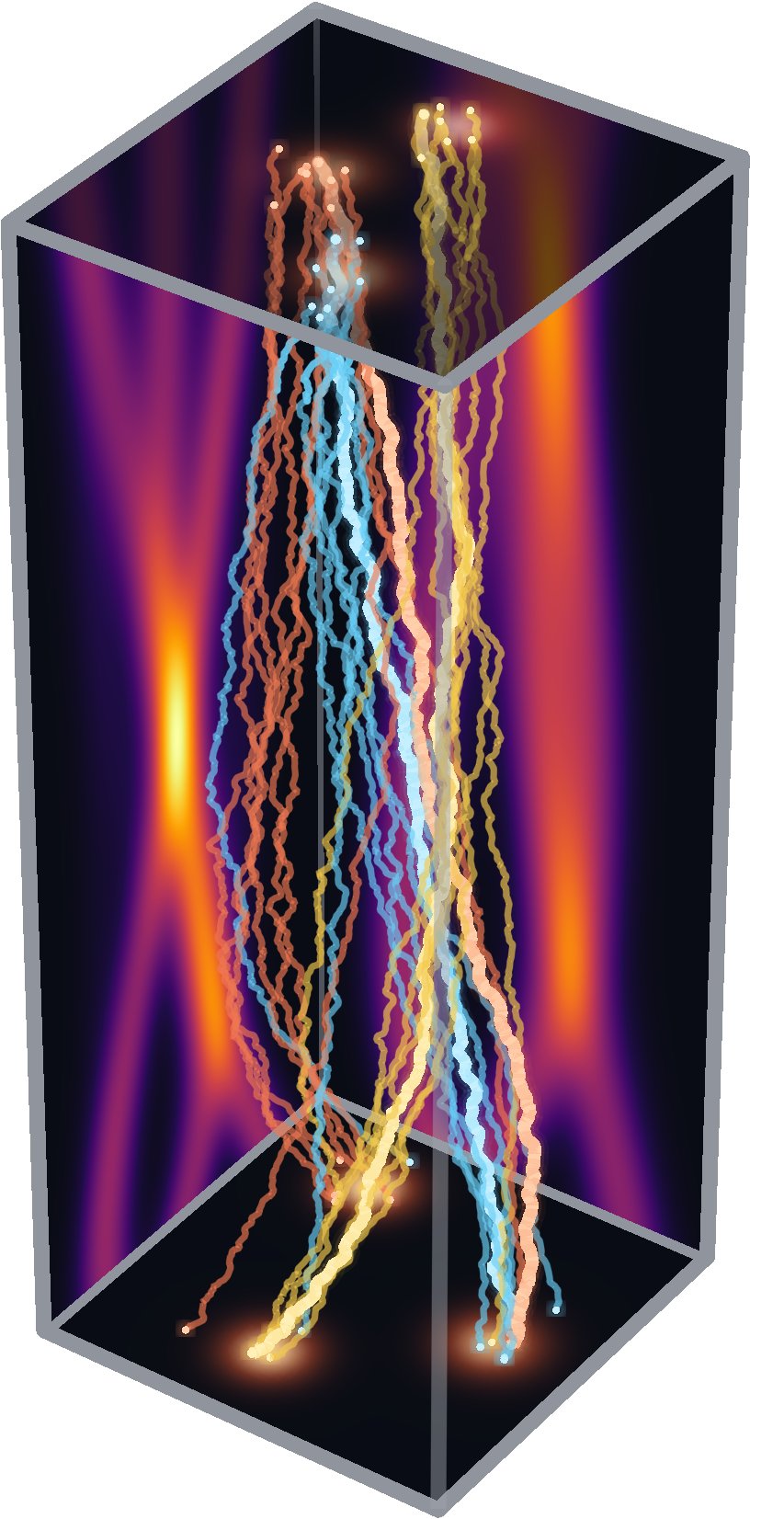}};
      \node[anchor=north,font=\fontsize{10}{10}\selectfont]
        at ([yshift=7pt]B.north) {$\rho_{\scriptscriptstyle0}$};
      \node[anchor=south,font=\fontsize{10}{10}\selectfont]
        at ([yshift=-6.5pt]B.south) {$\rho_{\scriptscriptstyle T}$};
    \end{scope}
    \begin{scope}[shift={(1.13, 0)}]
      \begin{scope}[shift={(0.03,0.35)}]
        \draw[transform shape, black!60, line width=7pt, -{Triangle[width=11pt, length=5.5pt, line width=0pt]}] (-0.1, 0) -- (0.3,0);
        \draw[transform shape, black!20, line width=6pt, -{Triangle[width=8.586pt, length=4.293pt, line width=0pt]}] (-0.08243, 0) -- (0.27515, 0);
        \node[font=\fontsize{5}{6}\selectfont,text=black!70] at (0.08,0.31) {\bfseries\textsf{COT}};
      \end{scope}
      \begin{scope}[shift={(0,-0.35)}]
        \draw[transform shape, black!60, line width=7pt, -{Triangle[width=11pt, length=5.5pt, line width=0pt]}] (0.3, 0) -- (-0.1,0);
        \draw[transform shape, black!20, line width=6pt, -{Triangle[width=8.586pt, length=4.293pt, line width=0pt]}] (0.28243, 0) -- (-0.07515, 0);
        \node[font=\fontsize{5}{6}\selectfont,text=black!70] at (0.1,-0.31) {\bfseries\textsf{Bridge}};
      \end{scope}
    \end{scope}
    \begin{scope}[shift={(2.7, 0)}]
      \node (C) at (0,0) {\includegraphics[height=134pt]{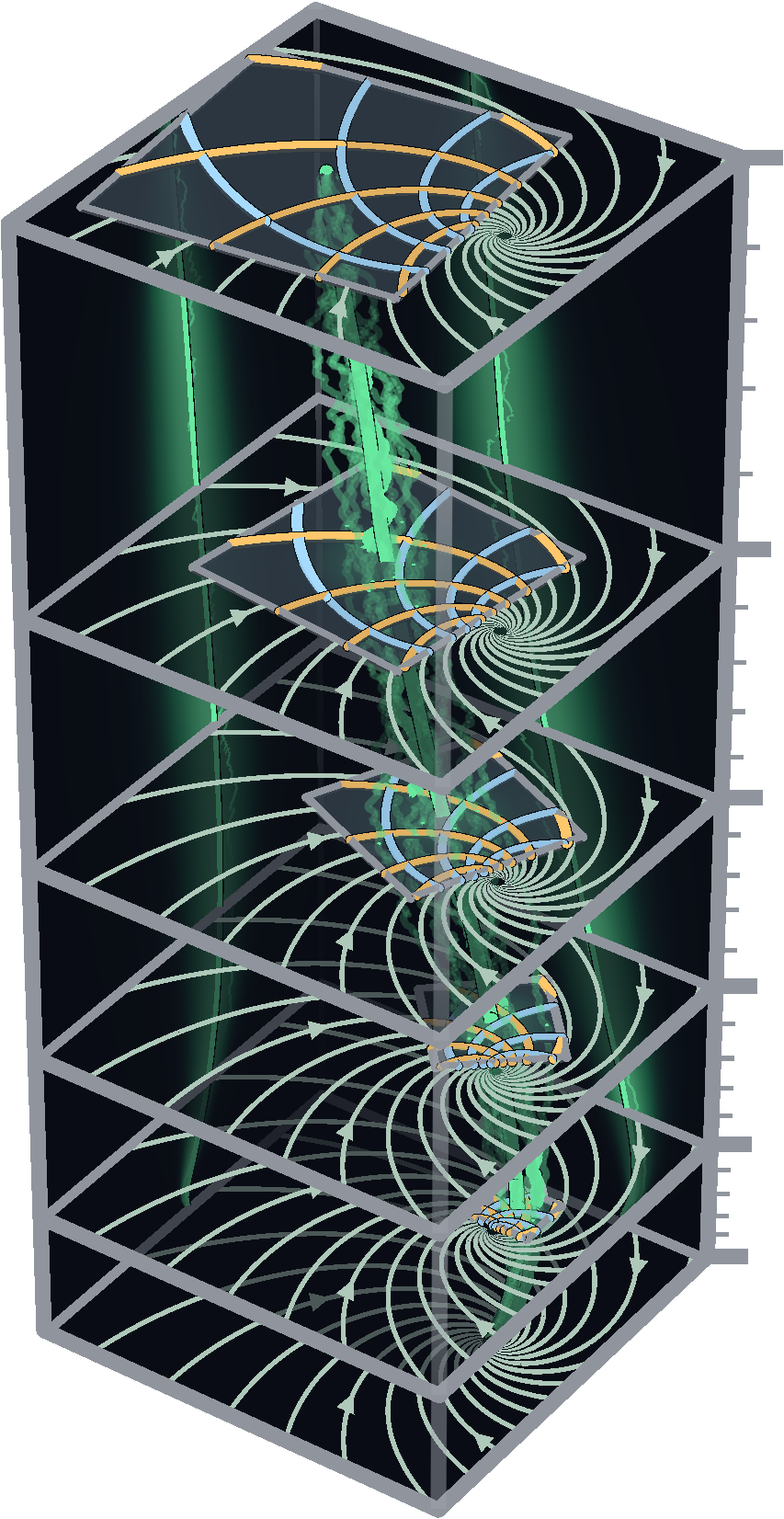}};
    \end{scope}
    \begin{scope}[
        overlay, shift={(C.north west)},
        x={($(C.north east)-(C.north west)$)},
        y={($(C.south west)-(C.north west)$)}]
      \coordinate (Tic0) at (0.99883,0.10381);
      \coordinate (Tic1) at (0.98359,0.36168);
      \coordinate (Tic2) at (0.97304,0.52512);
      \coordinate (DimAxis) at ([xshift=3.5pt]C.east);
      \coordinate (Dim0) at (DimAxis |- Tic0);
      \coordinate (Dim1) at (DimAxis |- Tic1);
      \coordinate (Dim2) at (DimAxis |- Tic2);
      \foreach \i in {0,1,2}
        \draw[black!60,line width=0.75pt]
          ([xshift=1pt]Tic\i) -- ([xshift=3pt]Dim\i);
      \draw[<->,>={Triangle[scale=0.6]},black!60,line width=0.4pt]
        (Dim0) -- node[midway,fill=white,inner sep=1.35pt,
          font=\scriptsize,text=black,scale=0.9] {$\Delta t$} (Dim1);
      \draw[<->,>={Triangle[scale=0.6]},black!60,line width=0.4pt]
        (Dim1) -- node[midway,fill=white,inner sep=1.35pt,
          font=\scriptsize,text=black,scale=0.8] {$\Delta t$} (Dim2);
    \end{scope}
  \end{tikzpicture}
  \vskip-4pt
  \caption{An overview of a complex SB and its associated conformal optimal transport geometry.} \label{fig:csb}
\end{wrapfigure}
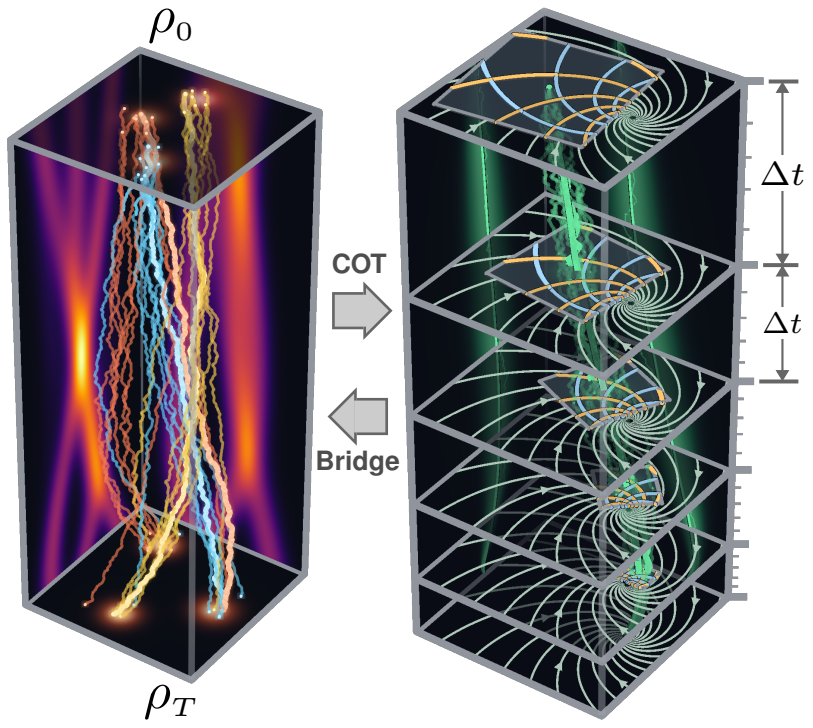

\section{Introduction}\label{sec:intro}

Diffusion bridges \citep{huang2026hogdiff,rogers2000diffusions,yue2024image} have evolved into foundational probabilistic models. Among existing formalizations, the Schr\"odinger bridge (SB) \citep{schrodinger1926quantisierung} has gained significant traction in generative modeling in machine learning \citep{dsb, deepgsb, tang2026foundations}. The framework is intrinsically linked to optimal transportation (OT), stochastic optimal control (SOC), and evolution of physical systems. These connections highlight a wide range of applications, indicating that diffusion bridges are fundamentally governed by the problem formulation and the geometric structures of the underlying spaces. However, designing coordinated priors that can effectively navigate vast optimization landscapes remains a largely open challenge. Embedding physically grounded biases into diffusion bridges provides a principled structural framework for modeling complex real-world dynamics.

The notion of extending SBs requires further elaboration, as the equivalence with dynamic OT implies solid optimization criteria in terms of kinetic energy. Recent studies \citep{deepgsb, liu2024generalized} have introduced \textit{generalized} SBs in a goal-driven manner by treating external state costs as auxiliary terms. In this way, there can be a collection of Hamiltonian functions which describe various particle motions, including scaling, translation, and rotation. This diversity raises a fundamental question:
\begingroup
\setlength{\partopsep}{0pt}
\addtolength{\topsep}{-2pt}
\begin{center}
  \textit{What principles should govern the real-world dynamics of general diffusion bridges?}
\end{center}
\endgroup
Existing SB formulations do not provide a satisfactory answer, as their restriction to linear dynamics on real-valued state spaces prevents them from characterizing rotational, oscillatory, and frequency-dependent behaviors in unified terms.  To address this challenge, we introduce an optimization framework based on a phase component endowed with a complex algebraic structure. By leveraging the nature of complex numbers, diffusion bridge models gain the freedom to explore phase trajectories that cannot be represented solely through task-dependent potential energy functions.

In this paper, we study a complex-valued formulation of the Schr\"odinger bridge problem (SBP). Our complexification of the SBP enables a succinct description of numerical and differential computations using complex operators. Furthermore, the log-Schr\"odinger potentials are unified using a rotational extension of the Hopf--Cole transform \citep{leger2021hopf}. \cref{fig:csb} presents an overview of the problem formulation; whereas OT priors yield straight paths, our method geometrically transforms SB paths into spiraling arcs. As solutions for this problem, we consider a complex-valued stochastic differential equation (CVSDE) \citep{revuz2013continuous} equipped with a McKean--Vlasov structure \citep{mckean1966class, vlasov1968vibrational}, defined by
\begin{equation}\label{eq:sse}
  \dd Z_t = i\mkern3.2mu\nabla_{\!\alpha} H\mkern-1mu\bigl(\mkern-1.2muZ_t, \bar{Z}_t, \nabla_{\!\vz}\mkern1mu\psi_t, \nabla_{\!\bar{\vz}}\mkern1mu\psi_t, \rho_t \mkern-1.2mu\bigr) \dd t + \sigma \dd W_t.
\end{equation}
Here, $H(\vz,\bar{\vz},\alpha,\beta,\rho)$ denotes a generalized Hamiltonian function tailored for conjugate variables $\vz, \bar{\vz} \in \bC^d$, paired vector fields $(\alpha, \beta) \coloneqq (\nabla_{\!\vz}\mkern1mu\psi, \nabla_{\!\bar{\vz}}\mkern1mu\psi)$ associated with the log-potential $\psi(\vz, \bar{\vz}, t)$, and a density $\rho(\vz,\bar{\vz},t)$. Notably, \eqref{eq:sse} corresponds to a formulation of the Schr\"odinger equation \citep{schrodinger1926quantisierung, wiseman2009quantum} with an associated Hamiltonian operator $\hat{H}$ on a Hilbert space:
\begin{equation}\label{eq:se}
  i\tfrac{\partial\Psi}{\partial t} = \hat{H}\Psi\hspace{3em}\textrm{and}\hspace{3em}i\tfrac{\partial\bar{\Psi}}{\partial t} = -\hat{H}^\dagger\bar{\Psi},
\end{equation}
where the factor $i\coloneqq\im$ plays the role of an orthogonal rotation on the gradient fields within the complex plane, generating phase-coherent trajectories. Under this formulation, the conjugate wavefunctions $\Psi$ and $\bar{\Psi}$ characterize a bosonic coherent state whose physics-inspired mean-field game (MFG) dynamics match the above CVSDE via exponential and logarithmic transformations.

How does this perspective relate to real-valued SB theory? We argue that our extension is a direct geometric consequence of the reference measure under the conformal invariance principle \citep{levy1965processus, lawler2008conformally, revuz2013continuous}. This introduces an extra geometric degree of freedom motivated by conformal field theory (CFT) \citep{francesco2012conformal}, with the transport cost defined along conformal geodesics. To this end, we propose the complex Schr\"odinger bridge (CSB), a novel theoretical framework designed to leverage these underlying structures in real-world physical systems with complex-valued, phase-related dynamics. We demonstrate that complex-valued channels play a role in balancing translational and rotational dynamics. We first derive the general CSB formulation, whose dynamics mirror those of real-valued diffusion bridges while incorporating analytic mapping functions in the complex plane. We then design a tractable class of CSB models tailored for image translation, termed Image-to-Image CSB (I$^2$CSB). Finally, we validate our approach in simulations of 3D circuit quantum electrodynamics (cQED), demonstrating its efficacy in modeling quantum physical systems.

This paper complements previous studies on diffusion bridges, and our CSB problem presents a novel geometric perspective on classical SBs, fundamentally expanding the capabilities of diffusion bridges for downstream tasks. Our contributions are summarized below.
\vspace{-1mm}
\begin{itemize}[leftmargin=*,noitemsep,partopsep=0pt,topsep=0pt,parsep=0pt]
  \item Building upon complex analysis and stochastic calculus, we formulate a novel generalization of the classical SB framework. Casting the problem via conformal optimal transport (COT) naturally incorporates the balance between Langevin diffusion dynamics and Hamiltonian rotational motion.
  \item We develop a comprehensive complex-valued theory and derive algorithms to solve various CSB problems. We propose SB-CVSDEs, complex-valued SDEs for general MFGs using Wirtinger calculus and conjugation algebra. This alleviates the mathematical ambiguity of solving a ``two-potential problem'': the controllable conjugate potential formally describes the reverse process.
  \item We validate our theoretical framework and algorithms across medical imaging and quantum physics. Our empirical results demonstrate that our algorithms effectively support both systems relying on implicit priors learned from static images and those governed by explicit physical priors.
\end{itemize}

%% file: text/2_prelim.tex
\section{Preliminaries}\label{sec:prelim}

\textbf{Notation.}\enspace We denote by $Z$ a $d$-dimensional complex stochastic process $(Z_t)_{0\le t \le T}$, with a map $Z: [0, T] \times \cZ\to \bC^d$ on the time interval $[0, T]$ and sample space $\cZ$. Here, $Z_t$ is a random variable representing the state at time $t$, and we write $\vz_t$ for a realization of $Z_t$. The complex Brownian motion is denoted by $W$ and is constructed from independent Brownian motions as $(\vw + i\hat{\vw})/\sqrt{2}$. Let $\Omega = \cC([0,T], \bC^d)$ denote the space of continuous paths on $[0,T]$, and let $\cP(\Omega)$ denote the space of probability measures on $\Omega$. For a path measure $\bP\in\cP(\Omega)$ describing the law of $Z$, we denote by $\bP_{\mkern-1.2mu t}$ its time marginal, describing the distribution of $Z_t$; that is, if $Z\sim\bP$, then $Z_t\sim\bP_{\mkern-1.2mu t}$. We alternatively write the density as $\rho(\cdot,t)$ to emphasize its interpretation as the limit of an empirical measure. All distributions admit densities on a measurable set $\cS\subseteq\bC^d$ with respect to a Lebesgue measure, and the KL divergence is defined accordingly and set to $+\infty$ when absolute continuity does not hold. For each SDE, local Lipschitz regularity of partial derivatives of drifts and potentials is assumed. 

\textbf{The Schr\"odinger bridge problem.}\enspace Let $\bQ_\vw$ be the reference path measure of a real Wiener process $\dd X_t = \sigma \dd \vw_t$, for a diffusion coefficient $\sigma$. Among path measures absolutely continuous with respect to $\bQ_\vw$, the SBP seeks a measure that is closest to the reference in terms of the KL divergence \citep{sbsurvey}: 
\begin{equation}\label{eq:sbp}
  \minimize_\bP\,\KL(\bP\Vert\bQ_\vw),\enspace\textrm{such that}\quad\bP\in\cP\bigl(\mkern-0.5mu\cC([0,T],\bR^d)\mkern-1.2mu\bigr),\quad\bP_{\mkern-1.2mu0}=\rho_{\scriptscriptstyle0},\,\bP_{\mkern-1.2muT}=\rho_{\scriptscriptstyle T},    
\end{equation}
where $\rho_{\scriptscriptstyle0}$ and $\rho_{\scriptscriptstyle T}$ are the marginal distributions at the beginning and end of the process, respectively. 
The problem corresponds to a dynamic formulation of entropy-regularized optimal transport (EOT) \citep{villani2009oldnew}. In EOT, the optimization yields a unique optimal coupling $\pi^\ast$ for the following objective 
\begin{equation}\label{eq:eot}
  \mathrm{EOT}(\mu,\nu)\coloneqq\! \inf_{\pi\in\Pi(\mu,\nu)}\mkern-1.2mu\iint_{\bR^d\times\bR^d}\!\mfrac{1}{2}\lVert\vx-\vy\rVert^2\, \dd\pi(\vx,\vy) + \sigma^2 \KL(\pi\Vert\mu\otimes\nu), 
\end{equation}
where $\Pi(\mu,\nu)$ represents the set of couplings and  $\mu\otimes\nu$ denotes the product of the marginals. In the Benamou--Brenier dynamic formulation \citep{brenier, bbrcd}, one finds that characterizing the optimal $\bP$ involves a pair of log-Schr\"odinger potentials, denoted in this work by $\{\chi_t\}_{t\in[0,T]}$ and $\{\eta_t\}_{t\in[0,T]}$, which together are dual objects associated with the mass-conserving density $\rho_t$ for $\rho_{\scriptscriptstyle0} =\mu$ and $\rho_{\scriptscriptstyle T} = \nu$.
 
\textbf{Geometric\;vs.\;dissipative dynamics.}\enspace Energy-based sampling methods---such as Hamiltonian Monte Carlo \citep{neal2011mcmc}---have introduced an auxiliary momentum variable to leverage the underlying symplectic geometry. Following the energy-based formulation, one can accommodate rotational particle dynamics using a broad class of stationary measures, while the corresponding Markov chains remain ergodic and time-reversible \citep{girolami2011riemann} up to momentum reversal (or conjugation). Let us consider a diffusion process with phase-space coordinates $(\vx_t, \vv_t)^\sT \in \bR^{2d}$ and a density $p(\vx, \vv)$ satisfying $\log p(\vx, \vv) \propto\! -E(\vx, \vv)$, combining Hamiltonian dynamics \citep{burylko2018coexistence} with a dissipative Langevin diffusion 
\begin{equation}\label{eq:cld_real}
  \begin{bmatrix} \dd \vx_t\\ \dd \vv_t \end{bmatrix} =
  \gamma \cdot \!\underbrace{\mfrac{\sigma^2}{4} \!\begin{pmatrix}
    -\nabla_{\vv}\mkern-1.2mu \log p\mkern1mu(\vx_t, \vv_t)\\
    \mkern14.5mu\nabla_{\vx}\mkern-1.2mu \log p\mkern1mu(\vx_t, \vv_t)
  \end{pmatrix}\!\dd t}_{\text{Hamiltonian dynamics}} + 
  \underbrace{
      \mfrac{\sigma^2}{4}\!\begin{pmatrix}
        \nabla_{\vx}\mkern-1.2mu\log p\mkern1mu(\vx_t, \vv_t)\\
        \nabla_{\vv}\mkern-1.2mu\log p\mkern1mu(\vx_t, \vv_t) 
  \end{pmatrix}\! \dd t +
  \mfrac{\,\sigma}{\sqrt{2}}\!\begin{pmatrix}
      \dd \vw_t\\
      \dd \hat{\vw}_t 
  \end{pmatrix}}_{\text{Langevin diffusion}},
\end{equation}
where $\vw_{\mkern-0.5mut}$ and $\hat{\vw}_{\mkern-0.5mut}$ are independent real Wiener processes. The real-valued coefficient $\gamma > 0$ balances oscillatory dynamics and mean-reverting diffusion, providing flexibility to account for an innate symmetry while the process reaches a stationary measure with respect to the given distribution. 

Our interest lies in a particular geometric extension (lifting), in which a pair of log-Schr\"odinger potentials induces energy-based dynamics of the SDE~\eqref{eq:cld_real}. In the special case of \cref{eq:sse}, a symmetry arises when $H$ generates unitary evolution, which prevents the formation of a bridge. It is therefore widely accepted that transitions from one steady state to another require nondeterministic mechanisms that break the underlying geometry, such as quantum jumps \citep{zoller1987quantum}. On the other hand, decoherence theory in quantum mechanics \citep{schlosshauer2007decoherence} provides a description purely in terms of wave functions, without jumps or wave function collapse. We focus on this geometric interpretation, particularly on applying conformal geometric approaches inspired by CFT \citep{francesco2012conformal, del2020decoherence, sajnok2025near} to describe stochastic Hamiltonians.

%% file: text/3_csb.tex
\section{Complex Schr\"odinger Bridges}\label{sec:csb}

This section develops a new bridge framework for the complex plane. We introduce complex operators for constructing SDEs and, drawing on a classical analogy, formulate the COT problem underlying this work. We then discuss a control-based dynamic formulation and computational considerations.

\subsection{Complex differential operators and Langevin dynamics}

We study a new geometric family of stochastic processes using complex operators. This recasts \cref{eq:cld_real} as a \textit{complexified} Langevin system \citep{kreutz2009complex, koor2023short} for a potential $\psi$ and establishes two distinct connections. The real part yields the probabilistic wave interpretation $\lVert \Psi_t \rVert^{2}\! = \rho_t$ with $\Psi = e^{\psi/\gamma}$. The imaginary part is a natural product of diffusion that identifies $\psi$ as an integrated complex Schr\"odinger potential.

Consider a complex-valued state $\vz_t \coloneqq \vx_t + i\vv_t$. We first introduce the Wirtinger gradient operators, defined with respect to $\vz$ and its complex conjugate $\bar{\vz}$. These operators permit the construction of a differential calculus for complex variables and the expression of dynamics in complex coordinates.

\begin{definition}[Multivariate Wirtinger operators] \label{def:multiwderv}
  Let $U \subseteq \bC^d$ be an open set and $f: U\to \bC$ be a smooth complex-valued function. Define multivariate Wirtinger gradient operators as
  \begin{equation}
    \begin{bmatrix} 
      \nabla_{\!\vz} f\\
      \nabla_{\!\bar\vz} f
    \end{bmatrix} \coloneqq
      \frac{1}{2}\biggl(\begin{pmatrix}
       1 & -i\,\\
       1 & +i\,
    \end{pmatrix}\otimes \vI_d\biggr) \begin{pmatrix}
      \nabla_{\!\vx} f\,\\
      \nabla_{\!\vv} f
    \end{pmatrix}\!, 
  \end{equation}
  where $i=\im$ and $\otimes$ here denotes the Kronecker product (see also \cref{def:wderiv} in Appendix~\ref{sec:cop}).
\end{definition}

\begin{wrapfigure}[16]{l}{0.36\textwidth}
  \def\figh{72.3668pt}
  \def\plth{75.5255pt}
  \def\colbarh{64.1098pt}
  \def\cpxbarh{63.4723pt}
  \vskip-18.4pt
  \centering
  \tikzset{inner sep=0, outer sep=0}
  \begin{tikzpicture}[tight background]
    \node[anchor=north west] (SETUP) at (0,0) {
      \includegraphics[height=\figh]{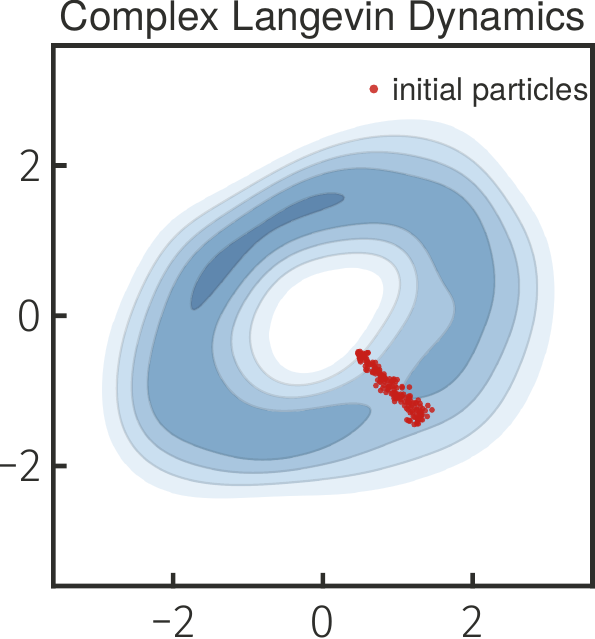}};
    \node[anchor=west] (PLT) at ([shift={(3.4893pt, -1.6797pt)}]SETUP.east) {
      \includegraphics[height=\plth]{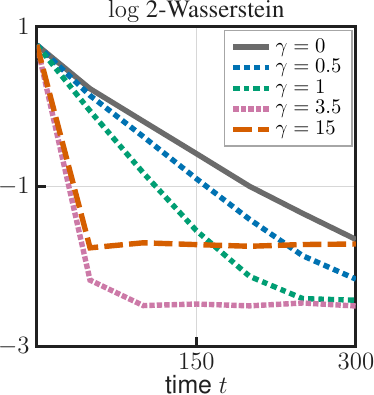}};
    \node[anchor=north west] (LOW) at ([yshift=-5.5pt]SETUP.south west) {
      \includegraphics[height=\figh]{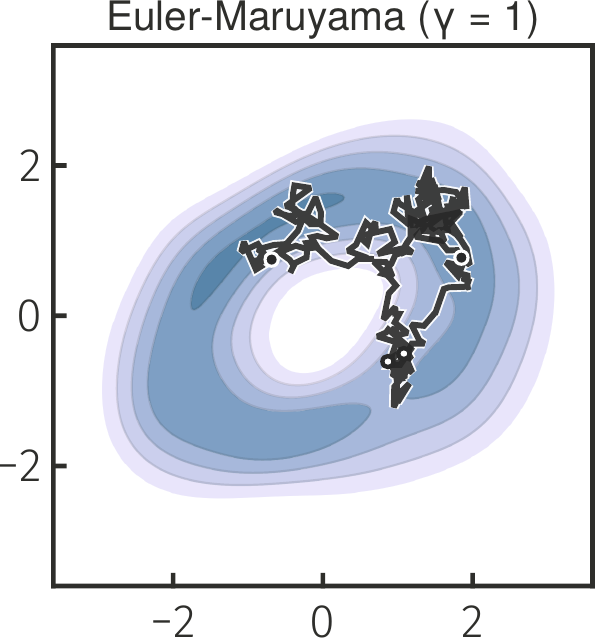}}; 
    \node[anchor=west] (HIGH) at ([xshift=4.5pt]LOW.east) {
      \includegraphics[height=\figh]{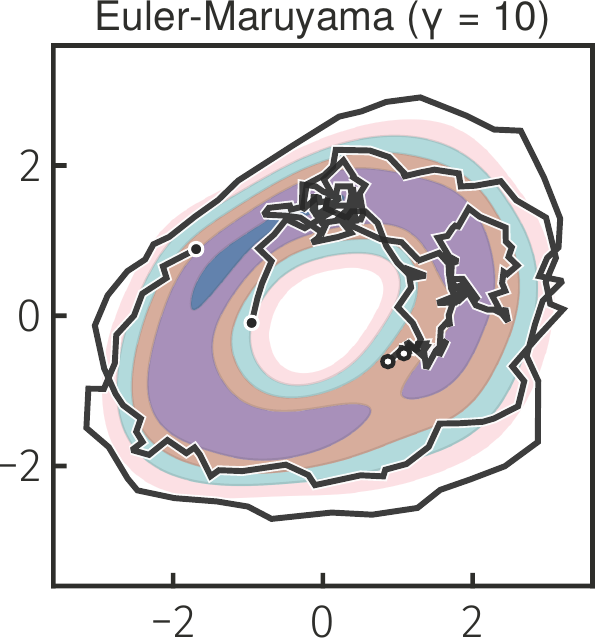}};
    \node[overlay, anchor=east] (COLBAR) at
        ([shift={(-3pt,0.3707pt)}]SETUP.west) {
      \includegraphics[height=\colbarh]{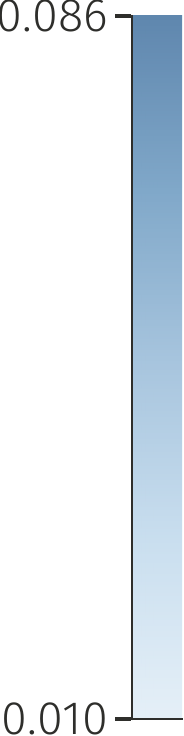}};
    \node[overlay, anchor=east] (CPXBAR) at
        ([shift={(-3pt,0.0650pt)}]LOW.west) {
      \includegraphics[height=\cpxbarh]{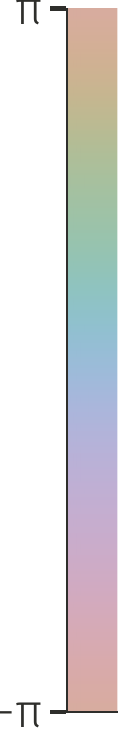}};
  \end{tikzpicture}
  \vskip-4pt
  \caption{Diffusion models approximate a distribution with different values of $\gamma$. Their stochastic trajectories exhibit varying dynamics and coverage of the density.}\label{fig:cld}
\end{wrapfigure}

Using the time-reversibility property \citep{girolami2011riemann}, we can find a pair of SDEs that describe an equilibrium state in both time directions. 
\begin{lemma}[Equilibria]\label{lem:cld}
  Consider the process $\vz_t = \vx_t + i\vv_t$ defined by \eqref{eq:cld_real} and its conjugate $\bar{\vz}_\ss$ with reverse time indexed by $s$. Consider a smooth function $\psi(\vz, \bar{\vz})$. The laws of two SDEs, denoted by the time-dependent density $\rho(\vz, \bar{\vz})$, coincide with a stationary measure for the following CVSDEs 
  \begin{subequations}\label{eq:sde0}
  \begin{empheq}[left={\empheqlbrace\mspace{3mu}}]{align}
    \dd Z_t &= i\mkern1.2mu\sigma^2\nabla_{\!\bar{\vz}}\mkern1.2mu \psi\mkern1.2mu(Z_t, \bar{Z}_t)  \dd t + \sigma \dd W_t \label{subeq:sde0f} \\[-0.1em]
    \dd \bar{Z}_\ss &= i\mkern1.2mu\sigma^2\nabla_{\!\vz}\mkern1.2mu \psi\mkern1.2mu(Z_\ss, \bar{Z}_\ss ) \dd s + \sigma \dd \bar{W}_\ss,
  \end{empheq}
  \end{subequations}
  if $\psi = \tfrac{1}{2} (\gamma - i) \log\mkern-1.2mu\rho + \mathrm{const.}$ holds for some constant in $\bC^1$. 
\end{lemma}
The lemma predicts a family of CVSDE models parameterized by $\gamma\in\bR$. \cref{fig:cld} demonstrates how this new degree of freedom in the dynamics affects convergence under different phase structures. This demonstrates that the complex formalization is innately suitable for encoding fundamental geometric priors. 

\subsection{The conformal optimal transport problem}

For an endpoint pair, the additional prior from the Hamiltonian dynamics is given by $b_\gamma(\vz, t; \vz_T) = -\frac{i\gamma}{T-t} (\vz - \vz_T)$, which introduces an endpoint-conditioned, proportional rotational drift around $\vz_T$. In this setup, we introduce a specific instance of entropy-regularized dynamical OT \citep{zhu2026geodesic, chen2016optimal} on the chosen logarithmic covering space and call this problem conformal optimal transport ($\textrm{COT}_\gamma$):
\begingroup
\addtolength{\abovedisplayskip}{-2pt}
\addtolength{\belowdisplayskip}{-5pt}
\begin{equation}\label{eq:cot}
\begin{gathered}
  \textrm{COT}_\gamma(\rho_{\scriptscriptstyle 0}, \rho_{\scriptscriptstyle T}) : \!\inf_{\pi\in\Pi(\rho_0, \rho_T)} \iint_{\bC^d\times\bC^d} \!\cC_\gamma(\vz_{\scriptscriptstyle 0}, \vz_{\scriptscriptstyle T}\mkern-1mu) \dd \pi(\vz_{\scriptscriptstyle 0}, \vz_{\scriptscriptstyle T}\mkern-0.5mu)  + \sigma^2\mkern0.5mu \KL( \pi \mkern0.5mu\Vert\mkern0.5mu \rho_{\scriptscriptstyle 0}\mkern-0.5mu \otimes \rho_{\scriptscriptstyle T}\mkern-1mu), \\[-2pt]
  \cC_\gamma(\vz_{\scriptscriptstyle 0}, \vz_{\scriptscriptstyle T}\mkern-1mu) \coloneqq\! \inf_{\substack{u(0)=\vz_0\\u(T)=\vz_T}} \int_{0}^T\mkern-1.2mu L_\gamma(u(\tau), \dot{u}(\tau), \tau)\,\dd\tau, \quad L_\gamma(\vz,\vp, t; \vz_{\scriptscriptstyle T}\mkern-1mu) = \bigl\lVert \vp - b_\gamma(\vz, t; \vz_{\scriptscriptstyle T}\mkern-1mu)\bigr\rVert^2  
\end{gathered}
\end{equation}
\endgroup
where the inner problem minimizes the integrated quadratic control cost over paths $u\in AC([0, T], \bC^d)$. Minimizing the energy relative to $b_\gamma$ yields the logarithmic spiral $u^\ast(t)=\vz_T+(\vz_0-\vz_T)(1-t/T)^{c_\gamma}$, with $\gamma$ controlling its rotation without changing the minimum control cost. If Novikov's condition holds under $\bQ_W$ on each interval $[0,t]$ with $t<T$, applying Girsanov's theorem \citep{sde} to a pushforward measure $\bP$, namely $\bP: \dd Z_t = i \sA_t \dd t + \sigma \dd W_t$, yields 
\begin{equation}\label{eq:com}
 \log \frac{\dd \bP}{\dd \bQ_W} \bigg\vert_{\cF_t} \mkern-5mu = \, 
  2\Re \bigg[\int_0^t(\bar{\vartheta}_\tau)^\sT\dd W_\tau\biggr]
 -\int_0^t \bigl\lVert \vartheta_\tau \bigr\rVert^2 \dd \tau, \qquad \vartheta_t = \frac{i \sA_t}{\sigma} 
\end{equation}
where $\Re[\cdot]$ denotes the real part and $\cF_t$ is the natural filtration up to time $t$. We can think of a pairwise mapping $(\cH_\gamma \vz)_t = \vz_T +\tfrac{h_\gamma(\log(1-t/T))}{h_0(\log(1-t/T))}(\vz_t - \vz_T)$ where $h_\gamma(\ell)\!\coloneqq e^{c_\gamma \ell}\enspace \forall \ell\in\bC$ and $c_\gamma = 1+i\gamma$. For the preimage $\bP = (\cH_{-\gamma})_\#\bP^\gamma$, we obtain $\KL(\bP^\gamma \Vert \bQ^\gamma_W) = \KL(\bP \Vert \bQ_W)$, which subsequently admits a distinct transformation using the reference transition density and a separate time scale and metric. 

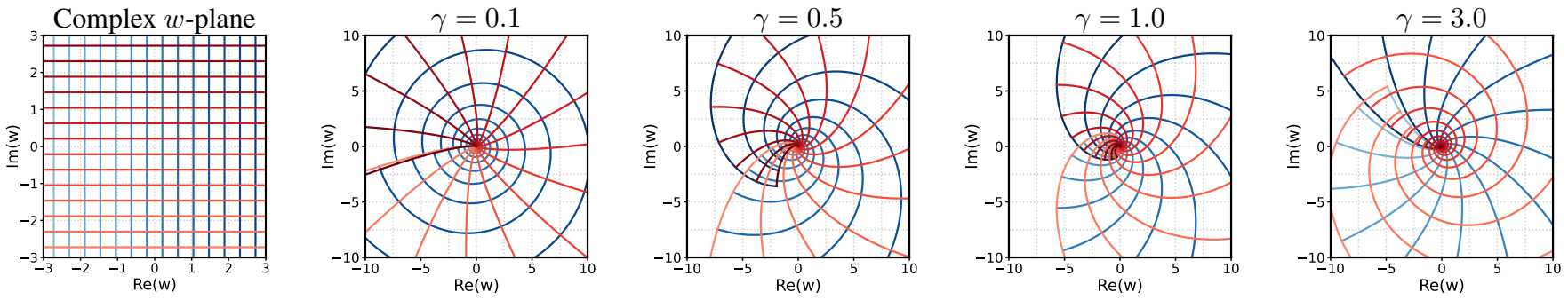
\begin{figure}
  \centering
  \def\planexx{0.12774152}\def\planeyy{1.04761420}
  \def\mapxx{0.13524856}  \def\mapyy{1.05174250}
  \tikzset{inner sep=0, outer sep=0}
  \begin{tikzpicture}[tight background]
    \begin{scope}[shift={(0.00750704, 0.00412831)}]
      \node at (0,0) {
        \includegraphics[height=55.29308741pt]{
          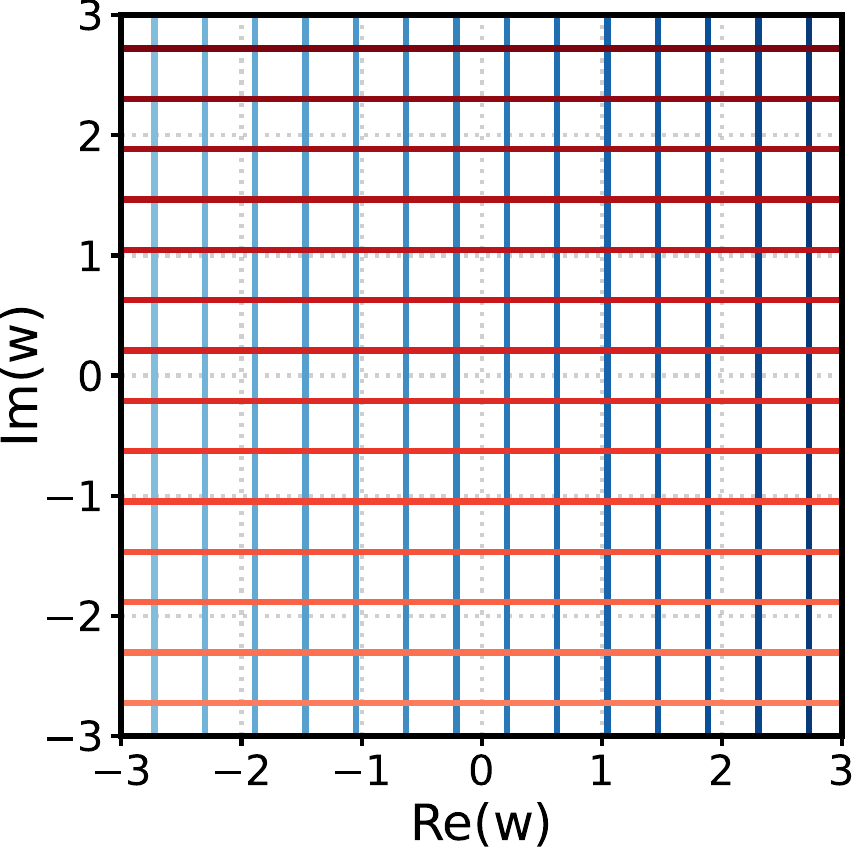}};
      \node[scale=0.6] at (\planexx, \planeyy) {Complex $w$-plane};
    \end{scope}
    \begin{scope}[xshift=2.3945620589cm]
      \node at (0,0) {
        \includegraphics[height=55.60401917pt]{
          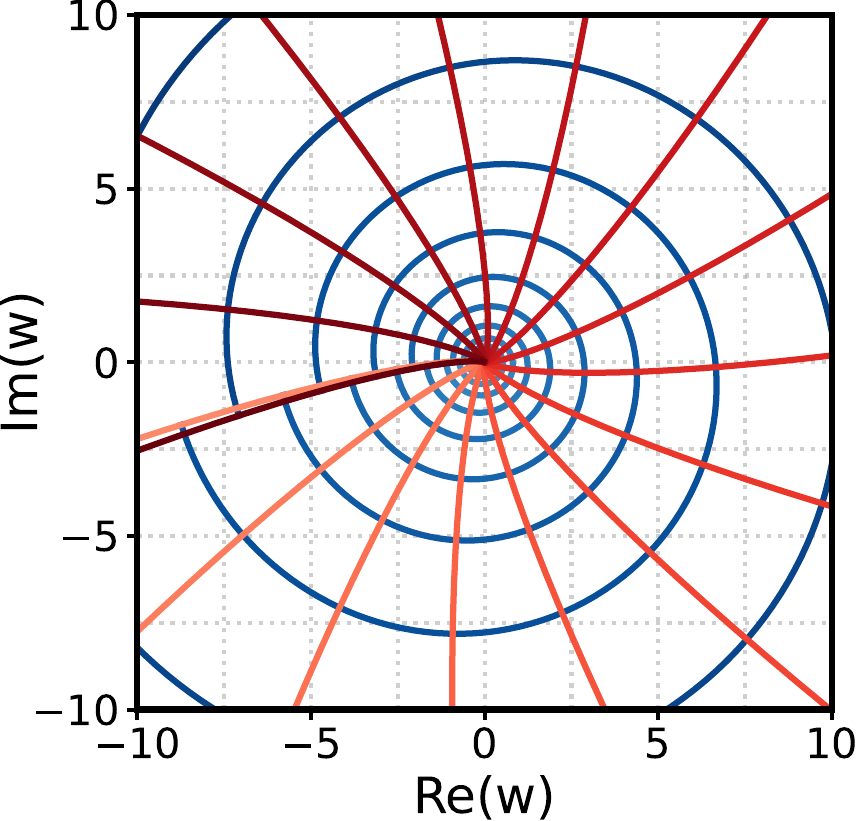}};
      \node[scale=0.6] at (\mapxx,\mapyy) {$\gamma=0.1$};
    \end{scope}
    \begin{scope}[xshift=4.7891241179cm]
      \node at (0,0) {
        \includegraphics[height=55.60401917pt]{
          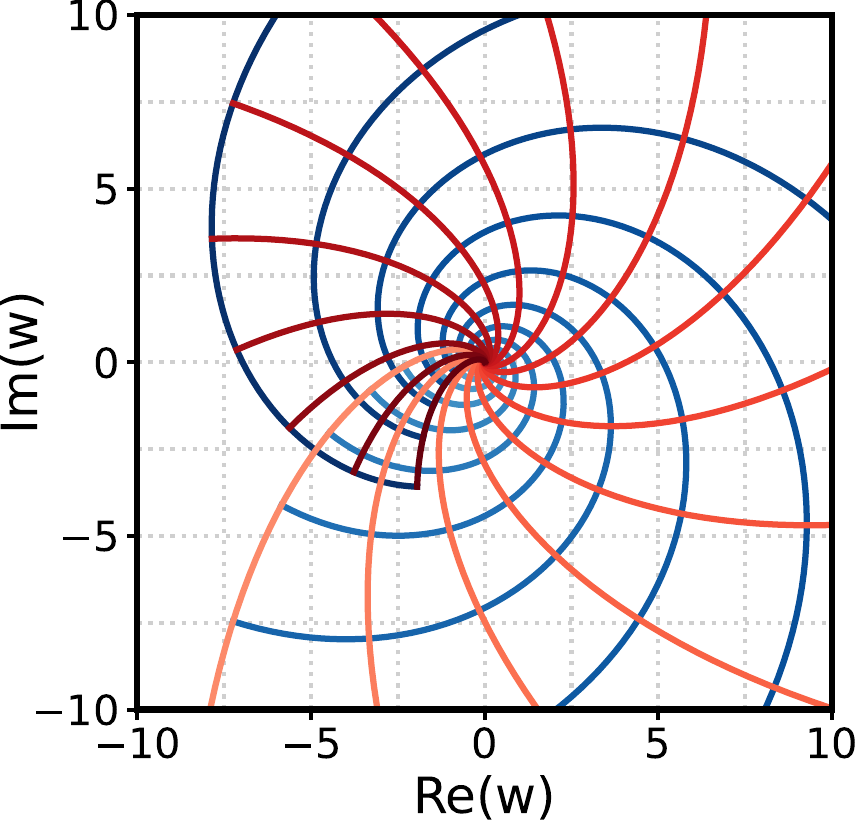}};          
      \node[scale=0.6] at (\mapxx,\mapyy) {$\gamma=0.5$};
    \end{scope}
    \begin{scope}[xshift=7.1836861768cm]
      \node at (0,0) {
        \includegraphics[height=55.60401917pt]{
          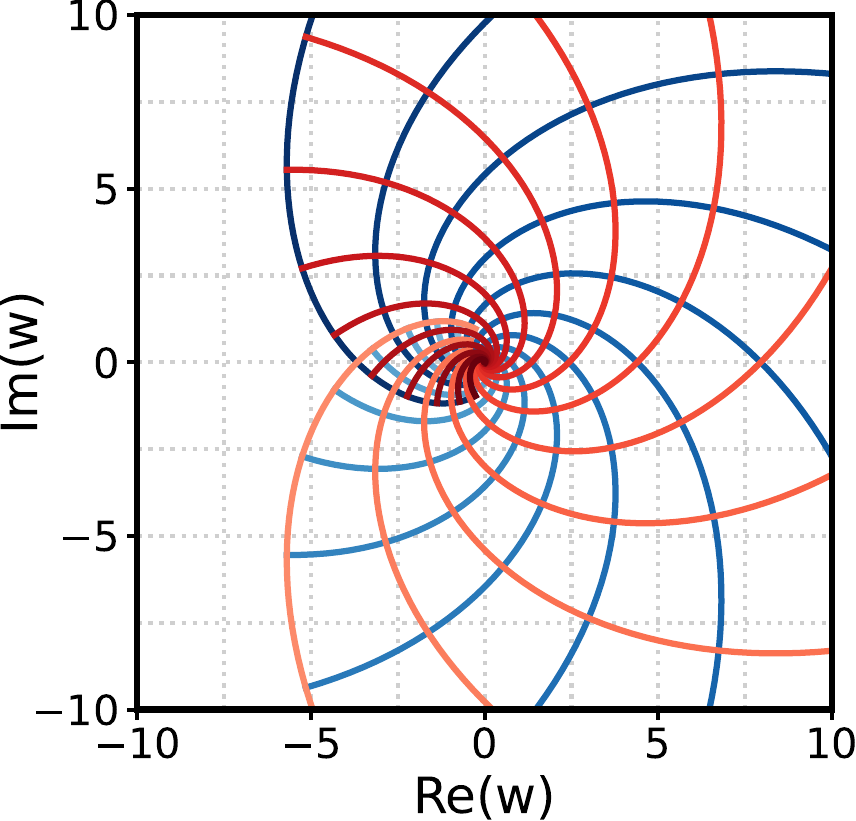}};          
      \node[scale=0.6] at (\mapxx,\mapyy) {$\gamma=1.0$};
    \end{scope}
    \begin{scope}[xshift=9.5782482358cm]
      \node at (0,0) {
        \includegraphics[height=55.60401917pt]{
          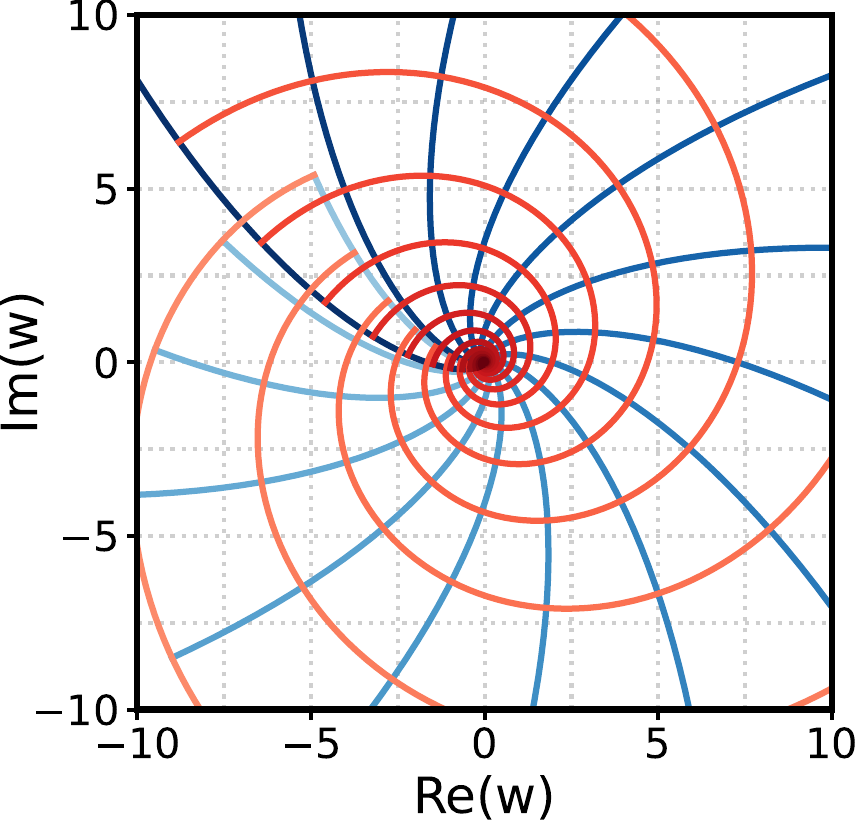}};          
      \node[scale=0.6] at (\mapxx,\mapyy) {$\gamma=3.0$};
    \end{scope}
    \path ([xshift=6.30856883pt]current bounding box.east);
  \end{tikzpicture}
  \vskip-2.76pt
  \caption{Conformal mappings under $h_\gamma(w)$ for varying coefficients. A Cartesian grid (normalized $w$-plane) is mapped biholomorphically onto a curvilinear grid, preserving right angles.} \label{fig:confmap}
\end{figure}

The rotational factor studied in \cref{eq:cot} has a distinct geometric interpretation as a coordinatewise \textit{conformal} transformation. The conformal mapping preserves angular magnitudes and orientations between curves, carrying essential geometric structure. According to L\'evy's characterization of complex Brownian motion (BM) \citep{levy1965processus, revuz2013continuous}, the pushforward $\bQ^\gamma_W = (\cH_\gamma)_\#({\bQ}_W)$ admits, in each uniquely normalized spiral coordinate under endpoint conditioning, a conformal local martingale that corresponds to a planar BM under its own quadratic-variation time change. Consequently, this allows us to formulate a KL objective:
\begingroup
\addtolength{\belowdisplayskip}{-2pt} 
\begin{equation}\label{eq:klobj} 
  \minimize_{\bP^\gamma_{0,T} = \rho_{0,T}} \mathrm{KL}(\bP^\gamma \Vert \bQ^\gamma_W), 
  \enspace\text{s.t.}\enspace\, \bQ^\gamma_W = (\cH_\gamma)_\#({\bQ}_W).
\end{equation}
\endgroup
As shown in \cref{fig:confmap}, for small $\gamma$, the mapping is predominantly characterized by scaling, whereas rotation dominates as $\gamma$ increases. As a whole, $\cH_\gamma$ is a quasiconformal homeomorphism that generates a family of boundary-constrained paths (see Appendix~\ref{sec:cmap}). Indeed, each time slice shows that the spiral connecting the endpoint pair exhibits self-similarity; this gives equivalent SB problems when the reference measure, admissible measures, and path space are transformed consistently. 

\subsection{Complex Schr\"odinger bridges in a mean field formulation} 

We identify CVSDEs that address general CSB problems. To proceed, let $f$ be an analytic function that serves as a \emph{base} drift for the diffusion bridge. By adopting an \emph{ansatz} for the generalized PDE system that retains the $\gamma$-rotational prior for \cref{eq:klobj}, we formulate the paired CVSDEs 
\begin{subequations}\label{eq:cvsb}
\begin{align} 
  \dd Z_t &= i\mkern1.2mu \bigl( f(Z_t, t) + \sigma^2 \nabla_{\!\bar\vz}\mkern1.2mu \psi\mkern1mu(Z_t, \bar{Z}_t, t) \mkern-0.5mu\bigr) \dd t + \sigma \dd W_t \label{eq:fwd}\\
  \dd \bar{Z}_\ss &= i\mkern1.2mu \bigl(\bar{f}(Z_\ss, T\shortminus s\mkern-0.5mu) + \sigma^2 \nabla_{\!\vz}\mkern1.2mu\psi\mkern1mu(Z_\ss, \bar{Z}_\ss, T\shortminus\mkern0.5mus\mkern-0.5mu) \mkern-0.5mu\bigr) \dd s + \sigma \dd \bar{W}_\ss, \label{eq:bwd}
\end{align}
\end{subequations}
in a symmetric form with respect to alternative forward and reverse time indices $t$ and $s$. The following proposition verifies the connection between an SB model and \cref{eq:se}, while $\Phi$ evolves under a twisted Hamiltonian function in heat-type PDEs on a measurable subset of $\bC^d$\!. 
\begin{proposition}[CSB-PDEs]\label{prop:csb-pde}
   Suppose that a function $H(\cdot, \cdot, \nabla_{\!\vz}\psi, \nabla_{\!\bar\vz}  \psi, \rho) = \sigma^2\langle\nabla_{\!\vz}\psi, \nabla_{\!\vz}\bar\psi \rangle + \nabla_{\!\vz} \psi^\sT f - \nabla_{\!\bar\vz} \psi^\sT \bar{f} + \frac{i\gamma\sigma^2}{2}\,\frac{\Delta\sqrt{\rho}}{\sqrt{\rho}}$ acts as a Hamiltonian in \cref{eq:mf}. Then, $\Phi = \exp(i\psi/c_\gamma)$ satisfies 
  \begingroup
  \addtolength{\abovedisplayskip}{-1.5pt}
  \addtolength{\belowdisplayskip}{-2.5pt}
  \begin{equation*}
    \begin{cases}
      \frac{\partial\Phi\mkern-1.2mu(\vz, \bar{\vz}, t)}{\partial t} = -i\nabla_{\!\vz}\Phi^\sT f + i\nabla_{\!\bar\vz} \Phi^\sT \bar{f} - \frac{\sigma^2}{4}c_\gamma\Delta\Phi + \frac{i\gamma\sigma^2}{2c_\gamma}\frac{\Delta\sqrt{\rho}}{\sqrt{\rho}}\Phi + c_\gamma^{-1}F\Phi, \\
      \frac{\partial \hat\Phi\mkern-1.2mu(\vz, \bar{\vz}, t)}{\partial t} = -i\nabla_{\!\vz}\mkern-0.5mu\boldsymbol{\cdot}(\hat{\Phi}f)+i\nabla_{\!\bar\vz}\mkern-0.5mu\boldsymbol{\cdot}(\hat{\Phi}\bar{f}) + \frac{\sigma^2}{4}c_\gamma \Delta{\hat{\Phi}} - \frac{i\gamma\sigma^2}{2c_\gamma}\frac{\Delta\sqrt{\rho}}{\sqrt{\rho}}\hat{\Phi} - c_\gamma^{-1}F\hat{\Phi},
    \end{cases}
    \mkern-12mu\text{s.t.}\enspace
    \begin{array}{@{}l@{}}
      \Phi_0\hat{\Phi}_0 = \rho_{\scriptscriptstyle0} \\[3pt] 
      \Phi_T\hat{\Phi}_T = \rho_{\scriptscriptstyle T}  
    \end{array}
  \end{equation*}
  \endgroup
  where $\Delta$ denotes the Laplacian operator, $\hat{\Phi}=\rho/\Phi$, and $F(\cdot, \cdot, \rho(\cdot, \cdot, t))$ denotes a real-valued function of state-dependent penalties and distributional interaction costs. 
\end{proposition} 
The complex Hopf–-Cole transform $(\Phi, \hat{\Phi})$ preserves mass under dynamics governed by the Hamiltonian and a heat term $c_\gamma\Delta\psi$. This allow us to derive forward and backward SDEs (FBSDEs) with mean field costs akin to the SB-FBSDEs \citep{likesb}, yet generalized to diffusion models of the COT problem.
\begin{theorem}[General SB-CVSDEs]\label{thm:sb-cvsde} 
  Suppose that $Y_t \coloneqq \chi(\vz_t, \bar{\vz}_t, t)$ and $\widehat{Y}_t \coloneqq -\eta(\vz_t, \bar{\vz}_t, t)$, with analytic $f$ satisfying $\nabla_{\!\vz}\boldsymbol{\cdot}f_t = \nabla_{\!\bar\vz}\boldsymbol{\cdot}\bar{f}_t$. For $\Phi=e^{i\psi/c_\gamma}$ and $\hat{\Phi}=e^{i\eta/c_\gamma}$, the CSB-PDEs give forward and backward SDEs with respect to \cref{eq:fwd}, $t\in[0,T)$:
  \begingroup
  \addtolength{\abovedisplayskip}{-1.5pt}
  \addtolength{\belowdisplayskip}{-1.5pt}
  \begin{subequations}\label{eq:cvsde1}
  \begin{empheq}[left={\begin{gathered}\textrm{SB}\\[-0.4em]{\textrm{\small CVSDEs}}\\[-0.4em]\textrm{wrt \eqref{eq:fwd}}\end{gathered}\mspace{-6mu}:
  \empheqlbrace}]{align}
    \dd Z_t &= i\mkern.5mu\bigl(f_t+\sigma^2\nabla_{\!\bar\vz}\mkern1.2mu\psi_t(Z_t, \bar{Z}_t, t)\mkern-0.5mu\bigr) \dd t + \sigma \dd W_{t}, \label{eq:cvsde1a}\\ 
    \dd Y_t &= -i\mkern.5mu\bigl(\mkern-1.2mu \lVert v_t \mkern-0.5mu\rVert^2 + F_t + \gamma\sigma\nabla_{\!\vz}\!\boldsymbol{\cdot}\mkern-1.2mu\check{v}_t - \tfrac{i\gamma}{2}\lVert \hat{v}_t\!+\!\check{v}_t\mkern-0.5mu\rVert^2 \mkern-0.5mu\bigr) \dd t + v^\ssT_t \dd W_t + \hat{v}^\ssT_t \dd \bar{W}_t,\label{eq:cvsde1b}\\
    \dd\widehat{Y}_t &= \begin{aligned}[t]
      -i\mkern.5mu &\bigl(\lVert v_t\mkern-0.5mu\rVert^2 + F_t - \gamma\sigma\nabla_{\!\vz}\!\boldsymbol{\cdot}\mkern-1.2mu\hat{v}_t - (1\!+\!\tfrac{i\gamma}{2})\lVert \hat{v}_t\!+\!\check{v}_t\mkern-0.5mu\rVert^2\mkern-0.5mu\bigr)  \dd t - {}\\[-1pt]
      &\bigl(\sigma\nabla_{\!\vz}\!\boldsymbol{\cdot}\mkern-1.2mu\check{v}_t + \sigma\nabla_{\!\bar\vz}\!\boldsymbol{\cdot}\mkern-1.2mu\bar{v}_t \mkern-0.5mu\bigr) \dd t - \bar{v}^\ssT_t\dd W_t - \check{v}^\ssT_t\dd \bar{W}_t,
    \end{aligned} \label{eq:cvsde1c}
  \end{empheq}
\end{subequations} 
  \endgroup
  where $\psi=\chi+\gamma\log\mkern-1.2mu\rho$, $\chi+\eta=-i\log\rho$, and
$(v_t, \hat{v}_t, \check{v}_t, \bar{v}_t)$ denote the Wirtinger gradients 
  \[
    v_t \coloneqq \sigma\nabla_{\!\vz}\mkern1mu\chi(\vz_t, \bar{\vz}_t, t),\enspace \hat{v}_t \coloneqq \sigma\nabla_{\!\bar\vz}\mkern1mu \chi(\vz_t, \bar{\vz}_t, t), \enspace  
    \check{v}_t \coloneqq \sigma\nabla_{\!\bar\vz}\mkern1mu\eta(\vz_t, \bar{\vz}_t, t), \enspace
    \bar{v}_t \coloneqq \sigma\nabla_{\!\vz}\mkern1mu\eta(\vz_t, \bar{\vz}_t, t). 
  \]
  Conversely, the reversed time evolution for $s=T-t$ with respect to \cref{eq:bwd} follows symmetrically (Appendix~\ref{subsec:sb-cvsde_proof}). The law of $Z_t$ matches that of $\bar{Z}_\ss$ under complex conjugation. 
\end{theorem}

\cref{thm:sb-cvsde} characterizes the general dynamics, where the potential induces a drift that follows the transformed reference along its gradient. For actual learning with nonlinear neural networks, one may employ an alternating optimization scheme based on changes of measure in opposite directions, or iterative proportional fitting (IPF) \citep{iproj_ipf, sscs}. In its imaginary channel, a derived algorithm updates parameters using bidirectional errors, with the optimum achieved when both fitted path laws agree.  
\begin{theorem}[Iterative proportional fitting]\label{thm:ipf}
  Consider the CSB problem without external costs. For parameters $(\theta,\phi)$, let the IPF estimates be defined using backward propagated changes of measure 
  \begin{align*}
    \cE^{\scriptscriptstyle+}_{T-\tau}(\phi; \theta) ={} &i\!\int_0^{\tau}\mkern-3mu\bE\Bigl[\lVert \check{v}^\phi_\ss \rVert^2 - \gamma\sigma\nabla_{\!\vz}\!\boldsymbol{\cdot}\mkern-1.2mu\check{v}^\phi_\ss - \bigl(1\!+\!\tfrac{i\gamma}{2}\bigr)\lVert v^\theta_\ss\!+\!\bar{v}^\phi_\ss\rVert^2 
    -i\sigma\bigl(\nabla_{\!\vz}\!\boldsymbol{\cdot}\mkern-1.2mu\hat{v}^\theta_\ss + \nabla_{\!\bar\vz}\!\boldsymbol{\cdot}\mkern-1.2muv^\theta_\ss\bigr) \Bigr]\!\dd s,\\[-1pt]
    \cE^{\scriptscriptstyle-}_{\tau}(\theta; \phi) = -&i\!\int_0^{\tau}\mkern-3mu\bE\Bigl[\lVert v^\theta_t \rVert^{2} - \gamma\sigma\nabla_{\!\vz}\!\boldsymbol{\cdot}\mkern-1.2mu\hat{v}^\theta_t - \bigl(1\!+\!\tfrac{i\gamma}{2}\bigr)\lVert \hat{v}^\theta_t\!+\!\check{v}^\phi_t\rVert^2  
    - i\sigma\bigl(\nabla_{\!\vz}\!\boldsymbol{\cdot}\mkern-1.2mu\check{v}^\phi_t + \nabla_{\!\bar\vz}\boldsymbol{\cdot}\bar{v}^\phi_t\bigr) \Bigr]  \dd t,
  \end{align*}
  where the signs $+$ and $-$ denote integration with respect to $s$ and $t$, respectively. The signed imaginary parts of the accumulated errors agree with the corresponding KL functionals up to additive constants,
  \[
    \mathop{\mathrm{KL}}(\bQ_{\phi} \Vert\, \bP_{\mkern-1.2mu\theta}) = -\Im\bigl[\cE^{\scriptscriptstyle +}_0(\phi; \theta)\bigr] + C_\phi \qquad\textrm{and}\qquad \mathop{\mathrm{KL}}(\bP_{\mkern-1.2mu\theta}\Vert\, \bQ_{\phi}) = \Im \bigl[\cE^{\scriptscriptstyle-}_T(\theta; \phi)\bigr] + C_\theta,
  \]
  where $\bP_\theta$ and $\bQ_{\mkern-1.2mu\phi}$ are the path measures obtained from CVSDEs~\eqref{eq:fwd}~and~\eqref{eq:bwd}, respectively.
  Here, $C_\phi$ is independent of $\theta$, and $C_\theta$ is independent of $\phi$, derived in Appendix~\ref{subsec:ipf_new}.
\end{theorem}
The IPF theorem suggests that the proposed family of complex-valued bridges can be trained.

%% file: text/4_algo.tex
\section{Algorithmic Considerations}\label{sec:algo}

We derive two primary training approaches from our theoretical framework. The first is a general CSB that utilizes IPF updates derived from SB-CVSDEs, whereas the second is a conditional CSB derived from COT. Both derived algorithms are distinguished by their capabilities and scalability.

\begingroup
\captionsetup{hypcap=false}
\begin{figure}[!ht]
\vskip-12pt 
\noindent
\begin{minipage}[t]{0.585\textwidth} 
  \begin{algorithm}[H]
    \caption{General CSB: biproportional projection.} \label{alg:csb} 
    \renewcommand{\algorithmicindent}{1.15em}
    \begin{algorithmic}[1] 
      \State \textbf{Input:} networks $(Y_{\mkern-1.2mu\theta}, \widehat{Y}_{\!\phi}, \hat{v}_\theta, \bar{v}_{\phi})$ and $\lambda_\textrm{IPF}, \lambda_\textrm{TD}\in\bR^+$\!.
      \While{$(\theta, \phi)$ have not converged}
        \For{$k = 1$ \textbf{to} $K$} \Comment{{\small Forward KL projection}}                \State $\cL^{\scriptscriptstyle+}(\phi;\phi^\prime, \theta^\prime) \gets 
            \lambda_\textrm{IPF}\cL^{\scriptscriptstyle+}_\textrm{IPF}\! + 
            \lambda_\textrm{TD}\cL^{\scriptscriptstyle+}_\textrm{TD}\!  + 
            \cL^{\scriptscriptstyle+}_0\mkern-1.2mu  + 
            \cL^{\scriptscriptstyle+}_\textrm{Drift}$
          \State Update $\phi$ with gradient $\hat{g}^{\scriptscriptstyle+}_\phi\!\approx\mkern-1mu \bE[\nabla_\phi\cL^{\scriptscriptstyle+}_\phi]$.
          \State $\phi^\prime \gets \mathtt{EMA}(\phi, \phi^\prime)$ 
        \EndFor
        \For{$k = 1$ \textbf{to} $K$} \Comment{{\small Reverse projection}}
          \State $\cL^{\scriptscriptstyle-}(\theta;\theta^\prime, \phi^\prime) \gets 
            \lambda_\textrm{IPF}\cL^{\scriptscriptstyle-}_\textrm{IPF}\! + 
            \lambda_\textrm{TD}\cL^{\scriptscriptstyle-}_\textrm{TD}\! + 
            \cL^{\scriptscriptstyle-}_T\mkern-1.2mu + 
            \cL^{\scriptscriptstyle-}_\textrm{Drift}$
          \State Update $\theta$ with gradient $\hat{g}^{\scriptscriptstyle-}_\theta\!\approx\mkern-1mu \nabla_\theta\cL^{\scriptscriptstyle-}_\theta$.
          \State $\theta^\prime \gets \mathtt{EMA}(\theta, \theta^\prime)$ 
        \EndFor
      \EndWhile
      \State \Return trained potentials and drifts $(Y_{\mkern-1.2mu\theta}, \widehat{Y}_{\!\phi}, \hat{v}_\theta, \bar{v}_{\phi})$.
    \end{algorithmic}
  \end{algorithm}
\end{minipage}
\hfill
\begin{minipage}[t]{0.398\textwidth} 
  \begingroup
  \makeatletter
  \xpatchcmd{\algorithmic}
  {\ALG@tlm\z@}
  {\ALG@tlm\z@\leftmargin0pt} 
  {}{}
  \makeatother
  \begin{algorithm}[H]
    \caption{Conditional CSB.}\label{alg:i2csb}
    \renewcommand{\algorithmicindent}{1em}
    \begin{algorithmic}
      \State \textbf{Input:} a model $\epsilon_\theta$, $p_{\scriptscriptstyle 0}(\cdot)$, and $p_{\scriptscriptstyle T}(\cdot\vert \vx_0)$.
      \Repeat
        \State $\vz_0\gets \vx_0 + i\vv_0$,\enspace  $\vz_{\scriptscriptstyle T}\gets \vx_{\scriptscriptstyle T} + i\vv_{\mkern-1.2mu \scriptscriptstyle T}$
        \State $\vz_t\sim q(\,\cdot\,\vert\, \vz_{\scriptscriptstyle0}, \vz_{\scriptscriptstyle T})$ according to \eqref{eq:posterior}.
        \State Update $\epsilon_\theta$ using the gradient of \eqref{eq:simple_loss}.
      \Until {convergence}
    \end{algorithmic}
  \end{algorithm}
  \endgroup
  \begingroup
    \vskip-4pt
    \centering
    \tikzset{inner sep=0, outer sep=0}
    \definecolor{textcol}{HTML}{292B31}
    \begin{tikzpicture}[tight background] 
      \node (Fig) at (0,0) {\includegraphics[width=125pt]{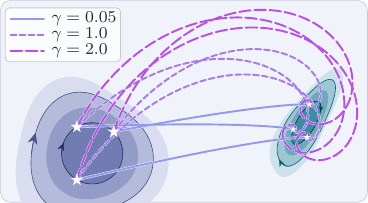}};
      \begin{scope}[
          shift={(Fig.south west)},
          x={($(Fig.south east)-(Fig.south west)$)},
          y={($(Fig.north west)-(Fig.south west)$)},
          text=textcol]
        \node[scale=0.7] (Rho0) at (0.475,0.12) {$\rho_{\scriptscriptstyle 0}$};
        \node[scale=0.7] (RhoT) at (0.87,0.17) {$\rho_{\scriptscriptstyle T}$};
        \node[scale=0.8] (Cd) at (0.945,0.92) {$\bC^d$};
      \end{scope}
    \end{tikzpicture}
    \vskip-2pt
    \captionof{figure}{Pairwise posterior flows.}
    \label{fig:post}
  \endgroup
\end{minipage}
\vskip-12pt 
\end{figure}
\endgroup

\subsection{Training general SB-CVSDEs via neural networks}

\cref{alg:csb} presents a straightforward realization of the general theory.
To solve the SB-CVSDEs, we parameterize independent potentials $(Y^\theta_t, \widehat{Y}^\phi_\ss)$ \citep{deepgsb}. As a direct consequence of \cref{thm:ipf}, we propose IPF losses, namely $\cL^{\scriptscriptstyle+}_\textrm{IPF}(\phi)$ and $\cL^{\scriptscriptstyle-}_\textrm{IPF}(\theta)$, to optimize likelihoods of the path measures with respect to KL projections. As boundary conditions, we apply score losses for marginals using the log-density estimate 
$i(Y^\theta - \widehat{Y}^\phi)$: 
$\cL^{\scriptscriptstyle+}_0(\phi) = \cL_\textrm{Score}(\phi; \rho_{\scriptscriptstyle0})$ and $\cL^{\scriptscriptstyle-}_T(\theta) = \cL_\textrm{Score}(\theta; \rho_{\scriptscriptstyle T})$.
For each time direction, the algorithm calculates the temporal differences (TDs) for small intervals $(\delta t, \delta s)$ as
\begin{subequations}
\begin{align} 
  \textrm{TD}^{+}_\ss(\phi;\theta^\prime) &\approx b(Y^{\theta^\prime}_\ss) \delta s +  (v^{\theta^\prime}_\ss)^\ssT \delta W_\ss + (\hat{v}^{\theta^\prime}_\ss)^\ssT \delta \bar{W}_\ss,\\ 
  \textrm{TD}^{-}_t(\theta;\phi^\prime) &\approx b(\widehat{Y}^{\phi^\prime}_t) \delta t - (\bar{v}^{\phi^\prime}_t)^\ssT \delta W_t - (\check{v}^{\phi^\prime}_t)^\ssT \delta \bar{W}_t,
\end{align}    
\end{subequations}
where $b(Y^{\theta^\prime}_\ss)$ and $b(\widehat{Y}^{\phi^\prime}_t)$ denote the target It\^o drift coefficients in \cref{appeq:cvsde2c} and \cref{eq:cvsde1c}, respectively, and $\delta W_t \sim \cC\cN(0, \delta t\,\vI_d)$ and $\delta W_\ss \sim \cC\cN(0, \delta s\,\vI_d)$ denote the standard approximation of complex Wiener differentials, with barred increments given by complex conjugation. Suppose that $R^+_\ss$ and $R^-_t$ are single-step or multistep return estimates \citep{rl} of $Y_\theta$ and $\widehat{Y}_\phi$ constructed from $(\textrm{TD}^{+}_\ss, \textrm{TD}^{-}_t)$, respectively, including the ground-truth boundary or bootstrap values. We propose regression losses   
\[ 
  \cL^{\scriptscriptstyle+}_\textrm{TD}(\phi; \phi^\prime, \theta^\prime) = \bE \bigl\lVert \widehat{Y}_\phi(\vz_\ss, \bar{\vz}_\ss, T-s) - \bar{G}^+_\ss \bigr\rVert,\qquad 
  \cL^{\scriptscriptstyle-}_\textrm{TD}(\theta; \theta^\prime, \phi^\prime) = \bE \bigl\lVert Y_\theta(\vz_t, \bar{\vz}_t, t) -G^-_t \bigr\rVert. 
\]  
Using our Hopf--Cole relation $\widehat{Y}=Y+i\log\rho$ in the corresponding parameterization, we set $\bar{G}^+_\ss=R^+_\ss+i\log\rho_\ss$ and $G^-_t=R^-_t-i\log\rho_t$, with the density evaluated at the return's starting state and time. Jointly minimizing the IPF, TD, and marginal score losses yields a unique solution to SB-CVSDEs, as established in \cref{thm:sb-cvsde}. We adopt exponential moving averages (EMAs) for the potential targets for additional stability. Additionally, we minimize the drift-matching losses $\cL^{\scriptscriptstyle\pm}_\mathrm{Drift}$ to enable fast SDE sampling and efficient iterative updates in practice \citep{deepgsb}.

\subsection{Analytic posterior for high-dimensional image translation and matching} 
\cref{alg:i2csb} demonstrates a conditional, image-to-image CSB (I$^2$CSB) that can be solved analytically from known boundary pairs using conformal geodesics.
For a fixed endpoint pair $(\vz_0, \vz_{\scriptscriptstyle T})$ and the posterior density \citep{i2sb} $q(\bar{\vz}_t\mid\bar{\vz}_T,\bar{\vz}_0)$, the reverse-drift target of our CSB at $0<t<T$ is given by
\begingroup
\addtolength{\abovedisplayskip}{5.5pt}
\addtolength{\belowdisplayskip}{5.5pt}
\begin{equation}
  i\hat{\sigma}^2_\ss\nabla_{\!\vz}\psi_\theta(\vz_t, \bar{\vz}_t, t\vert \vz_{\scriptscriptstyle0}) = -\bar{c}_\gamma \hat{\sigma}_\ss^2
  \nabla_{\!\vz_t}\log p_{\rm ref}(\vz_T,T\mid\vz_t,t)
+ \hat{\sigma}_\ss^2
  \nabla_{\!\vz_t}\log q(\bar{\vz}_t\mid\bar{\vz}_T,\bar{\vz}_0),
\end{equation}
\endgroup
where $s = T-t$ and $p_{\rm ref}$ denotes the transition density of the drift-free reference Brownian motion. Without diffusion, the trajectory follows the logarithmic spiral predicted from $h_\gamma$. We utilize the following proposition for computing an analytic posterior using the conjugate conformal geodesic.
\begin{proposition}[Analytic posterior]\label{prop:posterior}
  Let $\gamma \in \bR$, and let $\sigma_t > 0$ be scheduled continuously on $[0, T]$. Given a complex-valued boundary pair $(\bar{\vz}_0,\bar{\vz}_T)\in\!\bC^d\times\!\bC^d$ and a time-dependent diffusion coefficient, the $\textrm{COT}_\gamma$ posterior admits an analytic form based on a conjugate conformal geodesic:
  \begingroup\addtolength{\abovedisplayskip}{-1.2pt}
  \begin{equation}\label{eq:posterior}
    q(\bar{\vz}_\ss \vert \bar{\vz}_T, \bar{\vz}_0) = \cC\cN(\bar{\vz}_\ss \vert \bar{\mu}_\ss, \Sigma_\ss), \quad \bar{\mu}_\ss = \bar{\vz}_{\mkern-1mu\scriptscriptstyle T} + (\bar{\vz}_{\scriptscriptstyle0}\! - \bar{\vz}_{\mkern-1mu\scriptscriptstyle T}) \biggl(\!\mfrac{\hat{\omega}^2_\ss}{ \hat{\omega}^2_\ss+ \omega^2_\ss}\!\biggr)^{\mkern-7mu \bar{c}_\gamma}\mkern-5mu,\enspace\Sigma_\ss = \biggl(\!\mfrac{\omega^2_\ss \hat{\omega}^2_\ss}{\omega^2_\ss + \hat{\omega}^2_\ss }\!\biggr) \vI_d,
  \end{equation}
  \endgroup
  where $\hat{\omega}^2_\ss = \int^s_0 \hat{\sigma}^2_\tau \dd \tau$, $\omega^2_\ss = \int^T_\ss \hat{\sigma}^2_\tau \dd \tau$ with $\hat{\sigma}^2_\tau = \sigma^2_{T-\tau}$, and $\bar{c}_\gamma = 1-i\gamma$.
\end{proposition}
\cref{fig:post} shows that, with the rotational drift controlled by $\gamma$, the diffusion model will exhibit spiraling motion in the plane, for example by accelerating early and then rotating around the true samples. This is oftentimes useful in practice since we can reuse most of the score-based learning structure. For our approach, we consider a loss function 
\begingroup\addtolength{\abovedisplayskip}{-0.5pt}
\begin{equation}\label{eq:simple_loss}
  \cL(\theta) = \bE\Bigl\lVert\mathtt{cpx}\bigl(\epsilon_\theta(\vx_t, \vv_t, T\shortminus\mkern0.5mus\mkern-0.5mu)\bigr) - \mathtt{conj}\Bigl(\mfrac{\bar{\vz}_s - \bar{\vz}_0}{\hat{\omega}_s}\Bigr)\Bigr\rVert^2,
\end{equation}
\endgroup
where $\mathtt{cpx(\cdot)}$ denotes the basic operation of merging each corresponding pair of channels $\vx_{j}$ and $\vv_j$ into a single complex channel $\vz_j$ and $\mathtt{conj(\cdot)}$ denotes complex conjugation within each channel. With the model trained to predict the exact expected value, we perform DDPM-style generation \citep{ddpm, i2sb} with $\cC\cN(\bar{\vz}_n \mid \bar{\mu}_{n \mid n + 1}, \Sigma_{n \mid n + 1})$ for $1 \le n \le N - 2$ on a proper discretization into $N$ steps, which often improves robustness on image tasks. We defer the implementation details to \cref{sec:detail}.

%% file: text/5_expr.tex
\section{Experimental Results}\label{sec:expr}

We delineate our objectives as follows. \one{} We aimed to validate our theory across a wide range of domains in terms of generality, scalability, and the incorporation of explicit priors regarding complex values. \two{} We sought to analyze the derived algorithms and thoroughly understand their strengths and weaknesses compared with those of conventional real-valued models. \three{} We aimed to apply our method to data from realistic complex-valued datasets to observe the effectiveness of complex-valued data processing. To achieve these goals, we evaluated our algorithm across diverse problem settings, including classical control, image translation, and applications with complex-valued data.

\begin{figure}
  \centering
  \providecommand{\ctrlMetricsFile}{fig/ctrl/metrics_stacked_box.pdf}
  \tikzset{inner sep=0, outer sep=0}
  \begin{tikzpicture}[tight background]
    \DeclareFontFamily{U}{ctrlgreek}{}
    \DeclareFontShape{U}{ctrlgreek}{m}{n}{<-> gsmn1000}{}
    \providecommand{\ctrlgamma}{{\usefont{U}{ctrlgreek}{m}{n}g}}
    \def\aw{27.1bp}
    \def\bw{35.32bp}
    \def\sc{.3456}
    \def\hh{1.32bp}
    \foreach \t/\title/\i in {0.0/{Moons (t = 0)}/0,0.5/{t = 0.5 T}/1,1.0/{Circles (t = T)}/2} {
      \foreach \g/\j in {0.1/0,0.5/1,3/2} {
        \node (M\i\j) at (0.99836*\i,-0.99836*\j) {
          \includegraphics[width=\aw]{fig/ctrl/moons/trajectories/gamma\g_t\t.pdf}};
      }
      \node[scale=\sc,anchor=base] at ([yshift=\hh]M\i0.north) {\textsf{\title}};
    }
    \foreach \g/\j in {0.1/0,0.5/1,3/2} {
      \node[scale=\sc,rotate=90,anchor=south] at ([xshift=-.72bp]M0\j.west) {\textsf{\ctrlgamma{} = \g}};
    }
    \foreach \offset/\value in {-0.31485/{$-2$},0/0,0.31485/2} {
      \foreach \i in {0,1,2} {
        \node[scale=.24,text=black,anchor=north] at ([xshift=\offset cm,yshift=-.45bp]M\i2.south) {\textsf{\value}};
      }
    }
    \foreach \method/\m/\dx/\name in {deepgsb2/0/0/{DeepGSB},ours/1/3.92853/{CSB}} {
      \foreach \window/\i/\x/\interval in {0-25/0/3.63267/{t = 0 $\sim$ 0.25T},25-75/1/4.92101/{t = 0.25T $\sim$ 0.75T},75-100/2/6.20936/{t = 0.75T $\sim$ T}} {
        \node[anchor=north,xshift=-1.63bp] (S\m\i) at (\x+\dx,0 |- M00.north) {
          \includegraphics[width=\bw]{fig/ctrl/stunnel/\method/trajectory_forward_\window pct.pdf}};
        \node[anchor=north] (G\m\i) at ([yshift=-1.2bp]S\m\i.south) {
          \includegraphics[width=\bw]{fig/ctrl/gmm/\method/trajectory_forward_\window pct.pdf}};
        \node[anchor=north] (V\m\i) at ([yshift=-1.2bp]G\m\i.south) {
          \includegraphics[width=\bw]{fig/ctrl/vneck/\method/trajectory_forward_\window pct.pdf}};
        \ifnum\i=0
          \node[scale=\sc,anchor=base] at ([yshift=\hh]S\m\i.north) {\textsf{\name}};
        \else
          \node[scale=\sc,anchor=base] at ([yshift=\hh]S\m\i.north) {\textsf{\interval}};
        \fi
      }
    }
    \node[anchor=north west] (Ccost) at (11.08763,0.62145) {\includegraphics{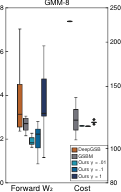}};
    \node[scale=1.28,anchor=north east] at ([xshift=-1.2bp,yshift=3.2bp]M00.north west) {\textsf{a}};
    \node[scale=1.28,anchor=north east] at ([xshift=-1.2bp,yshift=3.2bp]S00.north west) {\textsf{b}};
    \node[scale=1.28,anchor=north east] at ([xshift=-1.2bp]Ccost.north west) {\textsf{c}};
    \path ([xshift=1bp,yshift=1bp]current bounding box.north east);
    \path ([yshift=-.6bp]current bounding box.south west);
  \end{tikzpicture}
  \vskip-3.5pt
  \caption{Results on two-dimensional control problems. (a) Visualization of the potential function and Langevin dynamics for 3 seconds in the Moons-Circles bridge. (b) Learned CSB dynamics for S-Tunnel, GMM-8, and V-Neck. (c) Statistics on marginal fitting and total costs (5 different seeds).\vspace{-2mm}} \label{fig:ctrl}
\end{figure}

\subsection{General CSB for two-dimensional control tasks.}

We trained MLPs with the general CSB algorithm for four two-dimensional (2D) control benchmarks involving non-Gaussian initial marginals or a subtask of obstacle navigation. The mean-field particles were sampled from task-specific distributions and assigned random momentum drawn from a scaled normal distribution. We trained full potential and drift neural networks according to \cref{alg:csb}.

\begin{wrapfigure}[9]{l}{188.2bp}
  \centering
  \begin{tikzpicture}[tight background]
    \node[anchor=west,inner sep=0, outer sep=0] at (0,0)
      {\includegraphics[width=62bp]{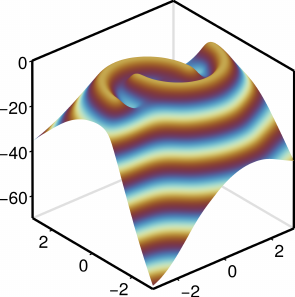}};
    \node[anchor=west,inner sep=0, outer sep=0] at (2.26,0)
      {\includegraphics[width=62bp]{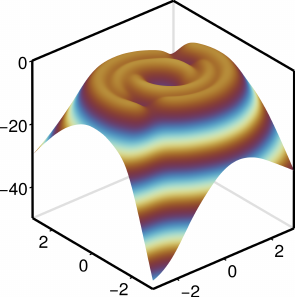}};
    \node[anchor=west,inner sep=0, outer sep=0] at (4.52,0)
      {\includegraphics[width=62bp]{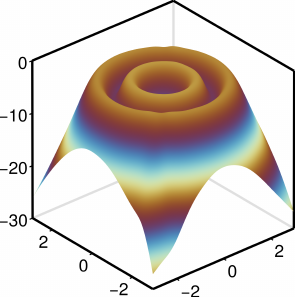}};
    \node[overlay,anchor=east,inner sep=0, outer sep=0] at (-.10333,0)
      {\includegraphics{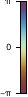}};
    \node[anchor=north west,draw=black,fill=white,
      rounded corners=0.3mm,line width=.4pt,inner xsep=2.5pt,inner ysep=1.4pt,minimum height=2.2mm]
      at (.10,1.23) {\scalebox{.65}{\textsf{Moons (t = 0)}}};
    \node[anchor=north west,draw=black,fill=white,
      rounded corners=0.3mm,line width=.4pt,inner xsep=2.5pt,inner ysep=1.4pt,minimum height=2.2mm]
      at (2.36,1.23) {\scalebox{.65}{\textsf{t = 0.5 T}}};
    \node[anchor=north west,draw=black,fill=white,
      rounded corners=0.3mm,line width=.4pt,inner xsep=2.5pt,inner ysep=1.4pt,minimum height=2.2mm]
      at (4.62,1.23) {\scalebox{.65}{\textsf{Circles (t = T)}}};
  \end{tikzpicture}
  \caption{Learned complex-valued log-potential functions at three distinct time points ($\gamma=0.5$).}\label{fig:pot3d}
\end{wrapfigure}

In \cref{fig:ctrl}~(a), our method successfully learned the Moons-Circles bridge, where color stripes depict energy levels across the phase structure. \cref{fig:pot3d} further displays the amplitude of $\frac{1}{2}c_\gamma \log\rho_t$ on 3D plots at zero momentum. For time indices $t\in\mkern-1.2mu\{0, 0.5 T\mkern-1.2mu, T\}$, we ran Langevin SDEs at each checkpoint. As $\gamma$ increases, the trajectories follow the internal energy landscape rather than arbitrary paths, demonstrating the correctness of our claim.

\cref{fig:ctrl}~(b)~and~(c) show qualitative and quantitative results on the S-Tunnel, GMM-8, and V-Neck benchmarks, where mean-field particles must avoid fixed obstacles while maintaining a diffusion bridge. We compared against two baselines: DeepGSB on the equivalent state space ($\bR^{2d}$), and the standard GSBM algorithm ($\bR^d$) using trainable splines for SB matching. Across all tested values of $\gamma$, CSB achieved lower mean endpoint $W_2$ than DeepGSB on all three benchmarks. CSB was competitive with GSBM on S-Tunnel and achieved the lowest mean endpoint 2-Wasserstein distance on GMM-8, where both baselines had their largest endpoint errors. CSB also achieved the lowest mean total trajectory cost in all three 2D control environments.

\begin{figure}[!ht]
  \vskip-2pt 
  \centering
  \begin{tikzpicture}[x=1bp,y=1bp,tight background,
      every node/.style={anchor=north west,inner sep=0pt,outer sep=0pt}]
    \pgfmathsetmacro{\groupgap}{(\linewidth/1bp-372.384)/2.5}
    \pgfmathsetmacro{\cifarx}{73.5+\groupgap}
    \pgfmathsetmacro{\imagenetx}{\cifarx+149.1+.5*\groupgap}
    \pgfmathsetmacro{\fidaxisx}{\imagenetx+73.5+\groupgap}
    \path[use as bounding box] (0,6) rectangle (\linewidth,-79.5);
    \node[scale=.5,anchor=base] at (36.75,2.1) {\textsf{MNIST}};
    \node[scale=.5,anchor=base] at ({\cifarx+36.75},2.1) {\textsf{CIFAR-10 (corrupted)}};
    \node[scale=.5,anchor=base] at ({\cifarx+112.35},2.1) {\textsf{CIFAR-10 (recovery)}};
    \node[scale=.5,anchor=base] at ({\imagenetx+36.75},2.1) {\textsf{ImageNet256}};
    \node at (0,0) {\includegraphics[width=73.5bp]{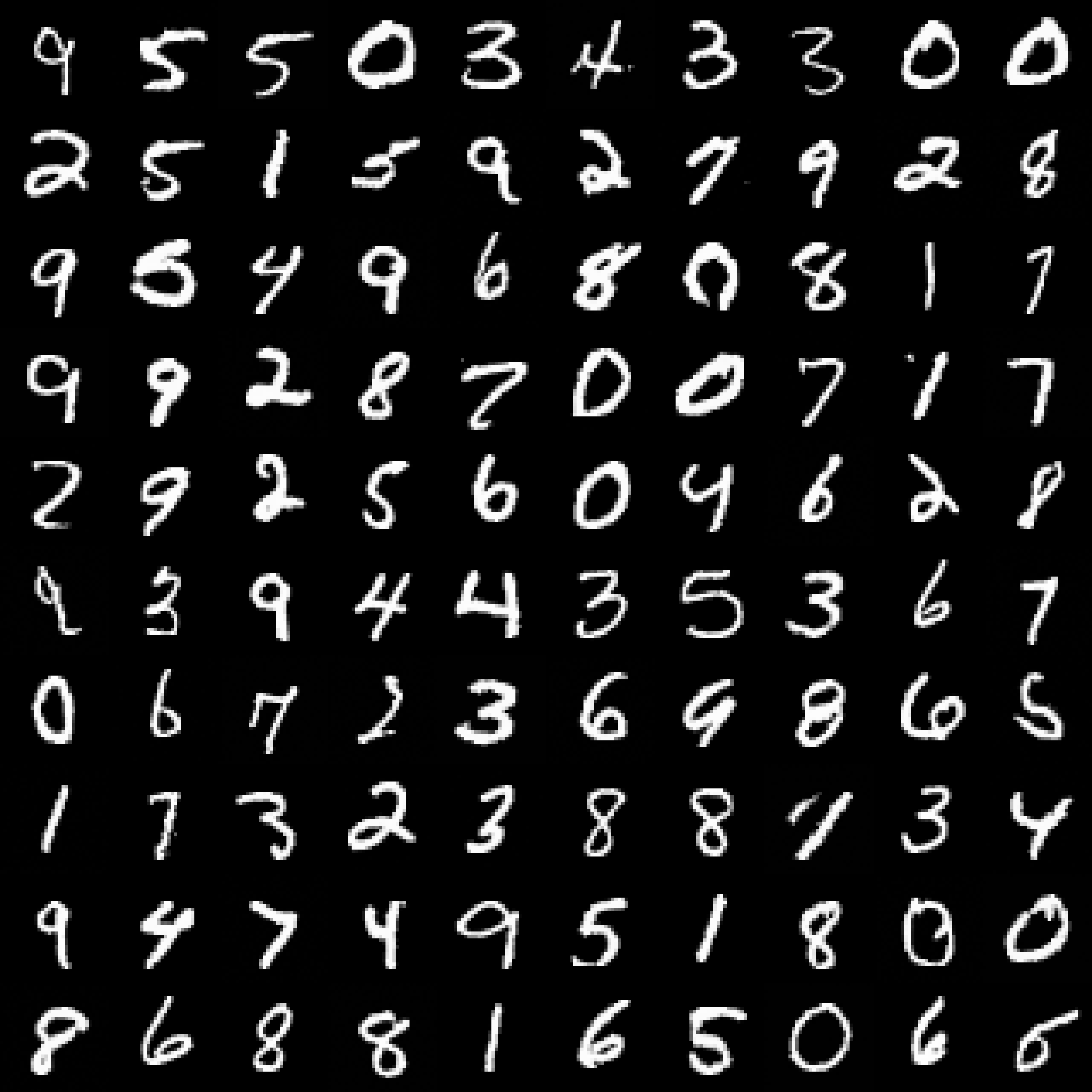}};
    \node at (\cifarx,0) {\includegraphics[width=73.5bp]{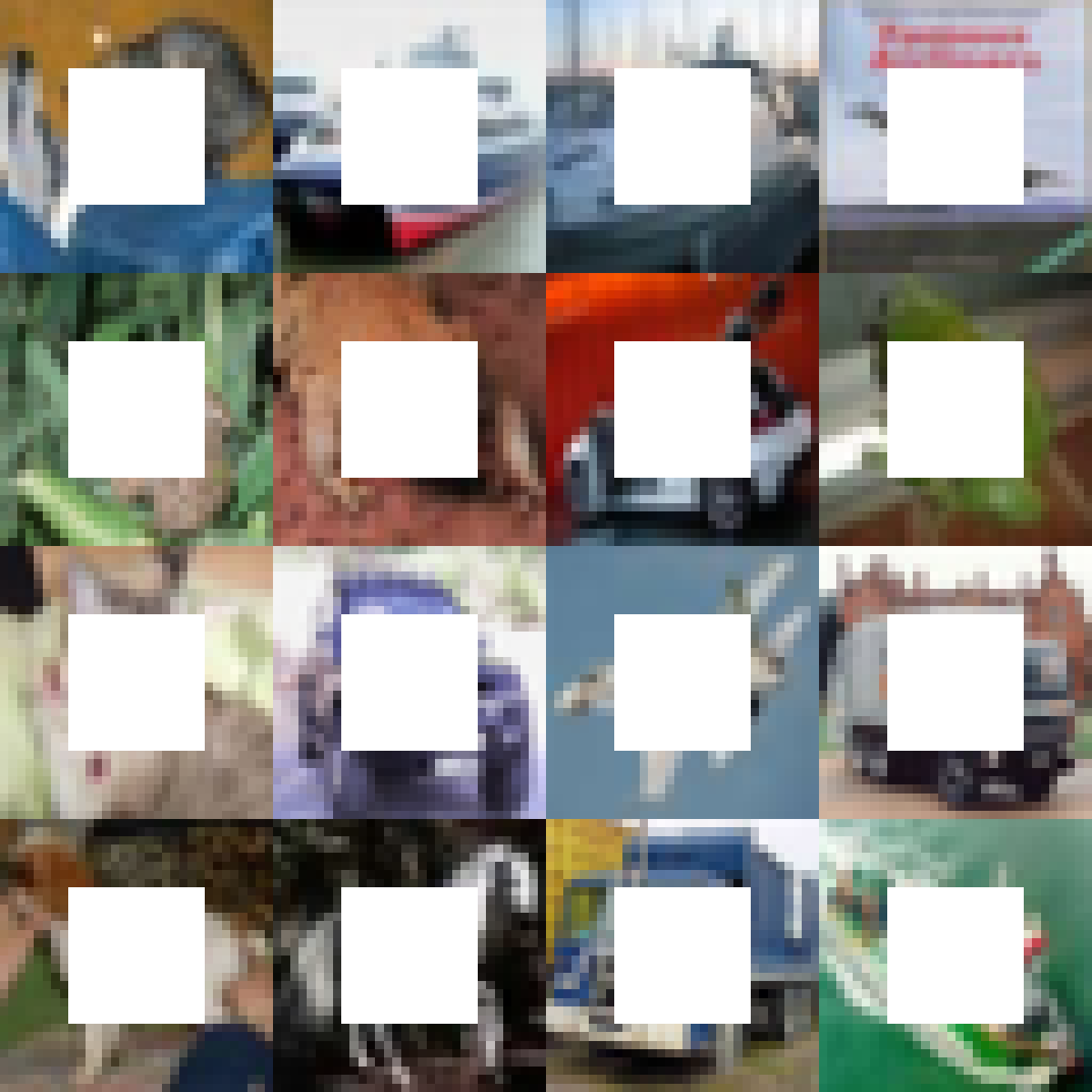}};
    \node at ({\cifarx+75.6},0) {\includegraphics[width=73.5bp]{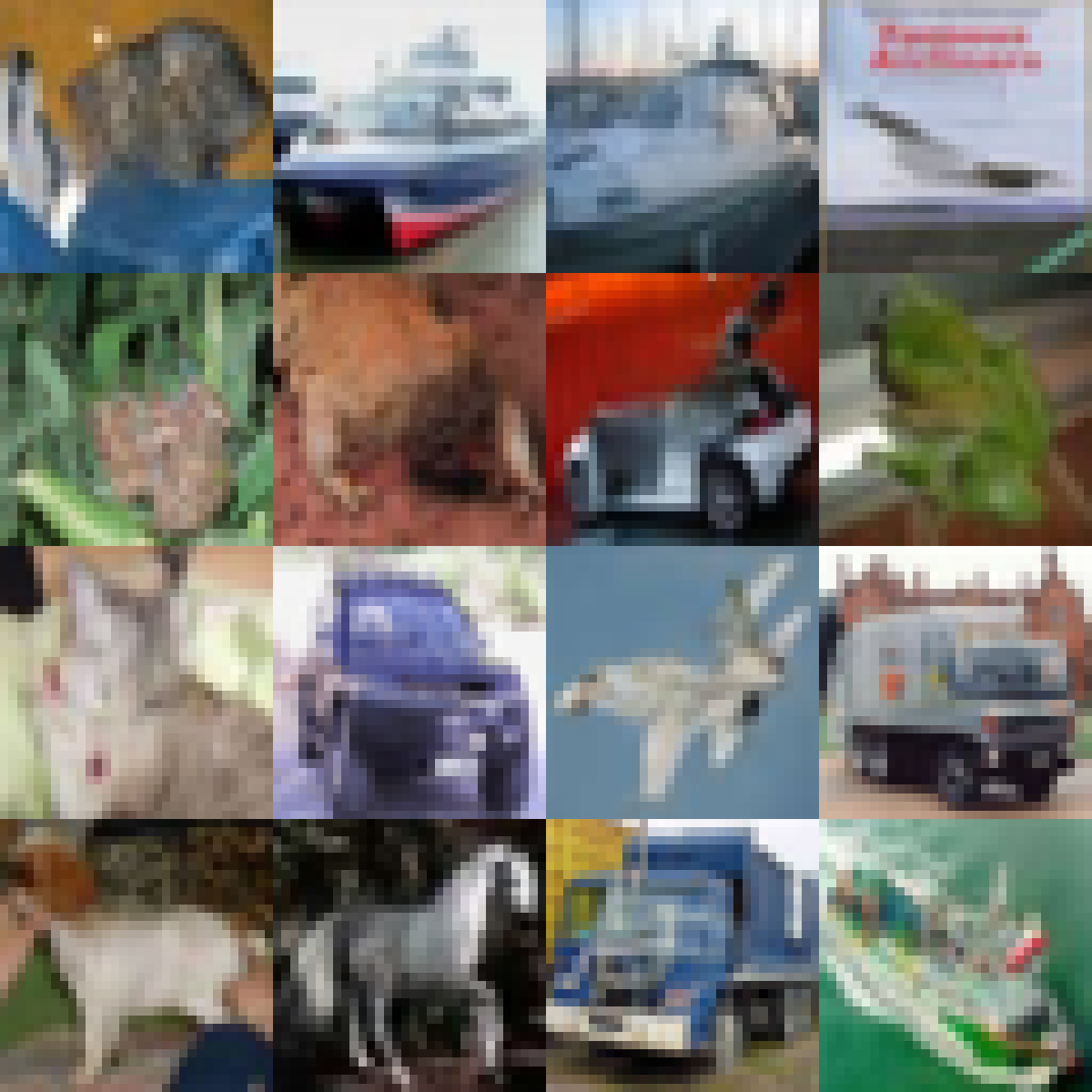}};
    \node at (\imagenetx,0) {\includegraphics[width=35.7bp]{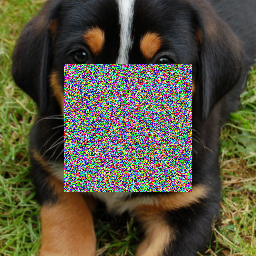}};
    \node at ({\imagenetx+37.8},0) {\includegraphics[width=35.7bp]{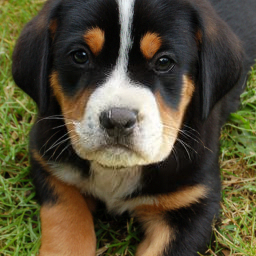}};
    \node at (\imagenetx,-37.8) {\includegraphics[width=35.7bp]{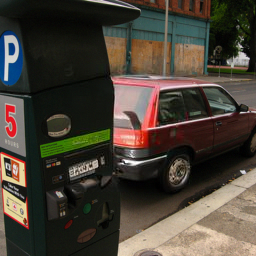}};
    \node at ({\imagenetx+37.8},-37.8) {\includegraphics[width=35.7bp]{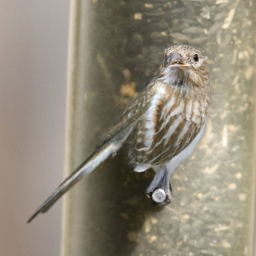}};
    \node at ({\fidaxisx-6.164},5.582) {\includegraphics{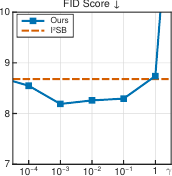}};
  \end{tikzpicture}
  \vskip-3.5pt
  \caption{Image-to-image translation results on RGB datasets. Left: unpaired MNIST generation. Middle: CIFAR-10 and ImageNet256 inpainting (25\%). Right: CIFAR-10 inpainting FID scores.}
  \label{fig:rgb}
  \vskip-12pt 
\end{figure}

\begingroup
\captionsetup{hypcap=false}
\begin{figure}[!t]
  \centering
  \begin{minipage}[t]{\dimexpr\linewidth-6.90cm-12bp\relax}
    \vspace{0pt}
    \centering
    \captionof{table}{Magnitude FID and SSIM (structural similarity index measure) metrics on the paired $64\times64$ synthetic brain MRI validation dataset.}
    \label{tab:mri_summary}
    \begin{tabular}{lrr}
      \toprule
      Method & FID $\downarrow$ & SSIM $\uparrow$ \\
      \midrule
      DSBM \citep{dsbm} & $23.1730$ & $0.6126$ \\
      SB Flow \citep{de2024schrodinger} & $21.4061$ & $0.6312$ \\
      I$^2$SB \citep{i2sb} & $15.2626$ & $0.6597$ \\
      Ours (best) & $\mathbf{12.0135}$ & $\mathbf{0.7139}$ \\
      \bottomrule
    \end{tabular}
  \end{minipage}%
  \hspace{12bp}%
  \begin{minipage}[t]{6.90cm}
    \vspace{0pt}
  \centering
  \begin{tikzpicture}[x=1cm,y=1cm,tight background,
      every node/.style={inner sep=0cm,outer sep=0cm}]
    \path (0,0) -- (6.90,0);
    \foreach \method/\x/\title in {
        source/0.24/{Source},
        dsbm/1.58/{DSBM},
        i2sb/2.92/{I$^2$SB},
        {csb/gamma_0.1}/4.26/{Ours},
        target/5.60/{Target}} {
      \node[scale=.5,anchor=base] at (\x+0.65,0.07) {\textsf{\title}};
      \node[anchor=north west] at (\x,0)
        {\includegraphics[width=1.30cm]{fig/mri/\method/sample_1739_magnitude.png}};
      \node[anchor=north west] at (\x,-1.34)
        {\includegraphics[width=1.30cm]{fig/mri/\method/sample_1739_phase.png}};
    }
    \node[scale=.55,rotate=90,anchor=south] at (0.19,-0.65) {$|z|$};
    \node[scale=.55,rotate=90,anchor=south] at (0.19,-1.99) {$\arg z$};
    \begin{scope}[every node/.append style={overlay}]
      \node[anchor=north west] at (7.00,-0.08)
        {\includegraphics[width=0.09cm,height=1.14cm]{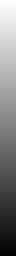}};
      \node[scale=.43,anchor=west] at (7.15,-0.08) {$1$};
      \node[scale=.43,anchor=west] at (7.15,-1.22) {$0$};
      \node[anchor=north west] at (7.00,-1.42)
        {\includegraphics[width=0.09cm,height=1.14cm]{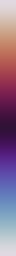}};
      \node[scale=.43,anchor=west] at (7.15,-1.42) {$\pi$};
      \node[scale=.43,anchor=west] at (7.15,-2.56) {$-\pi$};
    \end{scope}
  \end{tikzpicture}
  \caption{Complex-valued MRI generation: magnitude (top) and phase (bottom) representations.} 
  \label{fig:mri}
  \end{minipage}
  \par
  \vspace{10.8bp} 
  \providecommand{\qtPhaseIndex}{a}
  \providecommand{\qtSequenceIndex}{b}
  \providecommand{\qtGateIndex}{c}
  \tikzset{inner sep=0, outer sep=0}
  \def\bb{1.01}
  \def\xx{2.1}
  \def\yy{1.06}
  \def\ll{1.08}
  \def\llyy{2.14} 
  \def\llsc{0.5}
  \def\ow{56.5pt}
  \def\tw{76pt} 
  \def\qw{51.3pt}
  \begin{tikzpicture}[tight background]
    \begin{scope}[shift={(-3.6, 0)}] 
      \node[scale=\llsc] at (0, \llyy) {\textsf{Phase spaces}};
      \begin{scope}[shift={(0.01,0)}]
        \node at (-\bb, \bb) {
          \includegraphics[width=\ow]{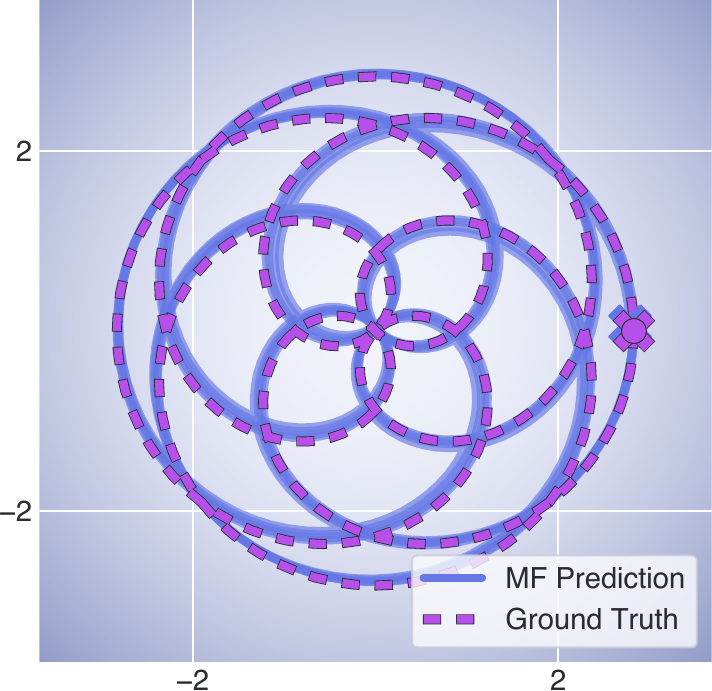}};
        \node at (\bb, \bb) {
          \includegraphics[width=\ow]{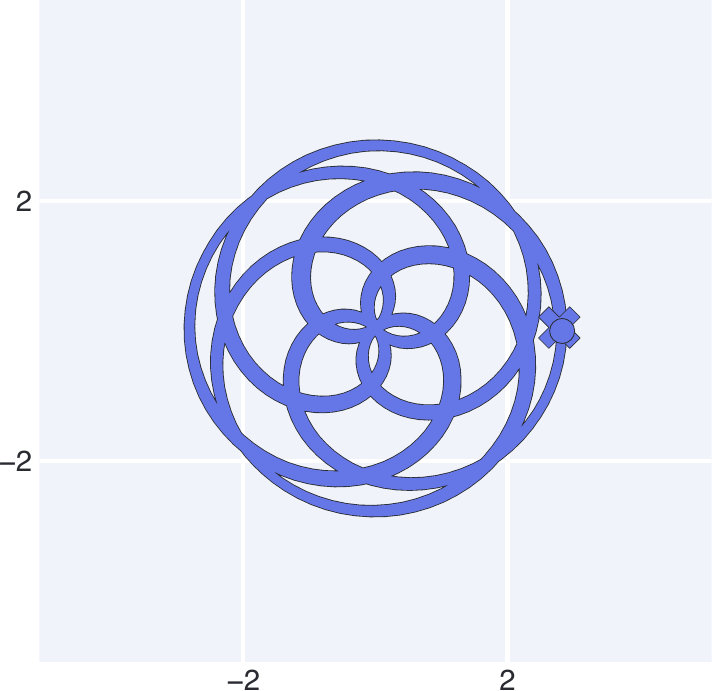}};
        \node at (-\bb, -\bb) {
          \includegraphics[width=\ow]{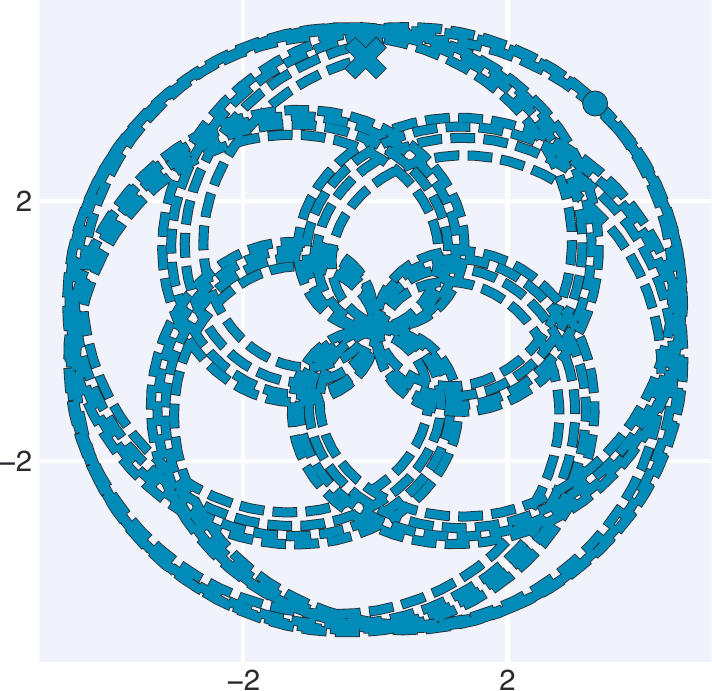}};
        \node at (\bb, -\bb) {
          \includegraphics[width=\ow]{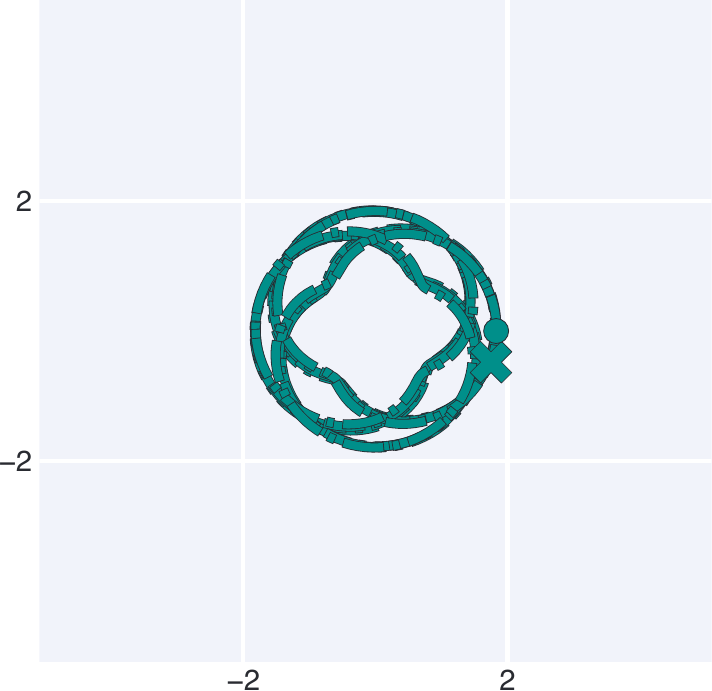}};
      \end{scope}
    \end{scope}
    \begin{scope}
      \node[scale=\llsc] (Seq) at (0, \llyy) {\textsf{Sequence}};
      \begin{scope}[xshift=0cm, yshift=0.038cm] 
        \node[anchor=north] (X1) at ([yshift=-1pt]Seq.south) {
          \includegraphics[width=\tw]{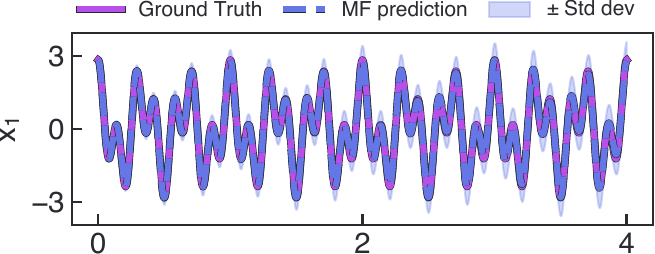}};
        \node[anchor=north] (X2) at ([yshift=-1pt]X1.south) {
          \includegraphics[width=\tw]{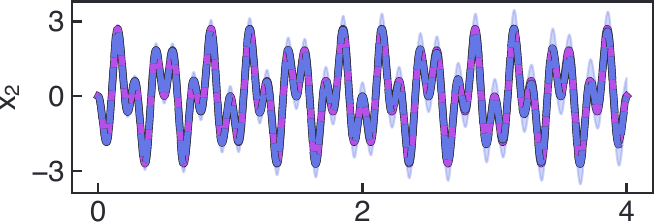}};
        \node[anchor=north] (V1) at ([yshift=-1pt]X2.south) {
          \includegraphics[width=\tw]{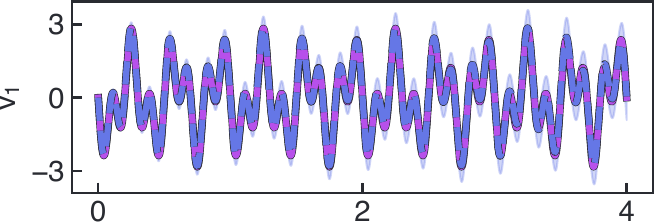}};
        \node[anchor=north] (V2) at ([yshift=-1pt]V1.south) {
          \includegraphics[width=\tw]{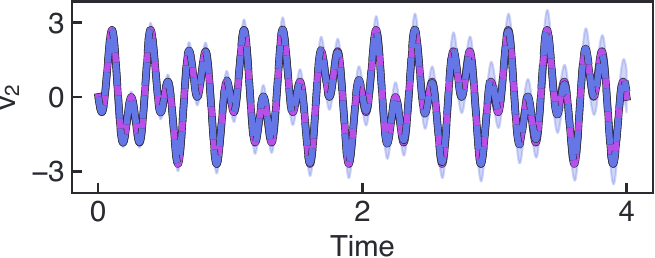}};
      \end{scope}      
    \end{scope}
    \begin{scope}[shift={(4.86, 0)}]
      \begin{scope}[shift={(-\xx, \yy)}]
        \node at (0,0) {\includegraphics[width=\qw]{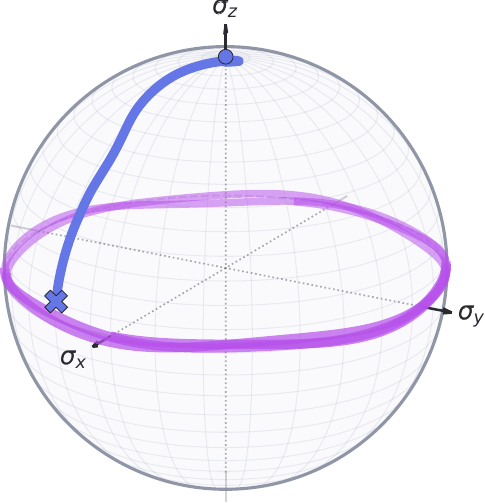}};        
        \node[scale=\llsc] at (-0.09,\ll) {\textsf{Ours {\small(Hadamard)}}};
      \end{scope} 
      \begin{scope}[shift={(0, \yy)}]
        \node at (0,0) {\includegraphics[width=\qw]{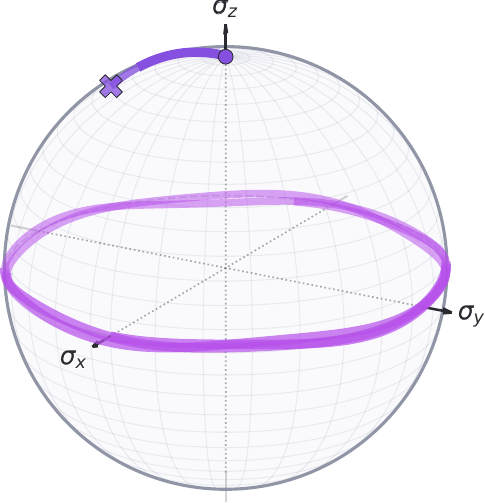}};
        \node[scale=\llsc] at (-0.09,\ll) {\textsf{ODE}};
      \end{scope}
      \begin{scope}[shift={(-\xx, -\yy)}]
        \node at (0,0) {\includegraphics[width=\qw]{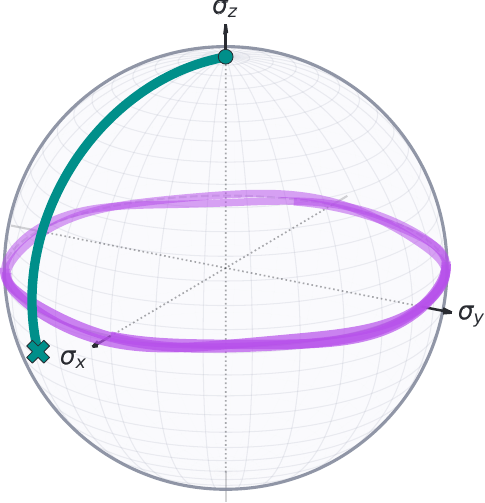}};
        \node[scale=\llsc] at (-0.09,\ll) {\textsf{SB-CFM}};
      \end{scope}       
      \begin{scope}[shift={(0, -\yy)}]
        \node at (0,0) {\includegraphics[width=\qw]{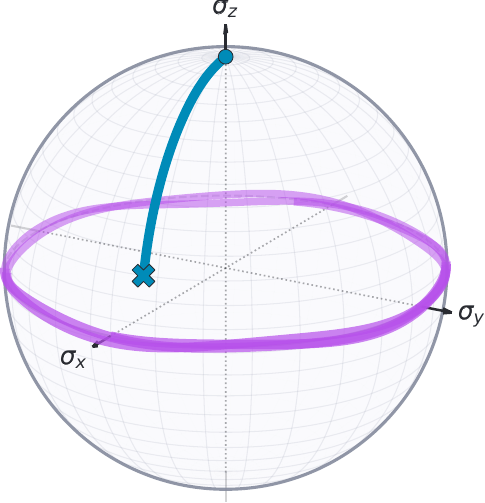}};
        \node[scale=\llsc] at (-0.09,\ll) {\textsf{OT-CFM}};
      \end{scope} 
      \begin{scope}[shift={(\xx, \yy)}]
        \node at (0,0) {\includegraphics[width=\qw]{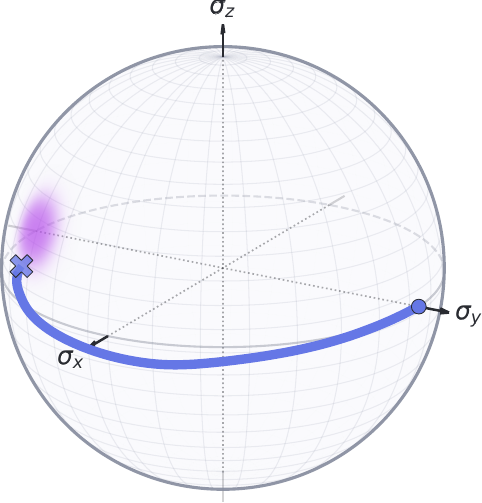}};
        \node[scale=\llsc] at (-0.09,\ll) {\textsf{Ours {\small(Pauli-X)}}};
      \end{scope}
      \begin{scope}[shift={(\xx, -\yy)}]
        \node at (0,0) {\includegraphics[width=\qw]{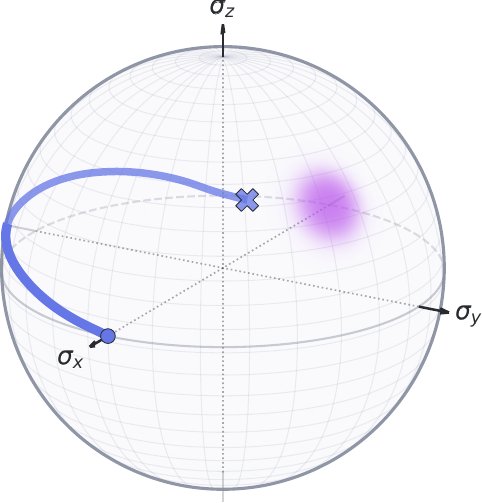}};
        \node[scale=\llsc] at (-0.09,\ll) {\textsf{Ours {\small(Pauli-Y)}}};
      \end{scope}      
    \end{scope}
    \begin{scope}
      \node[scale=1.28] at (-5.65, 2.1) {\textsf{\qtPhaseIndex}};
      \node[scale=1.28] at (-1.24, 2.075) {\textsf{\qtSequenceIndex}};
      \node[scale=1.28] at (1.75,  2.1) {\textsf{\qtGateIndex}};
    \end{scope}
    \path ([xshift=5pt]current bounding box.east);
  \end{tikzpicture}
  \caption{Complex SB with a synthetic cQED dataset. (a) Physical simulation and prediction for a harmonic oscillator with fluctuation noise. (b) For a long time horizon, our SB representation was fitted to the optimal state. (c) 3D qubit logic gate simulation (averaging over 100 particles).} \label{fig:qt}
\end{figure}
\endgroup
        
\subsection{Image-to-image translation}

\textbf{Paired and unpaired image generation.}\enspace Using I$^2$CSB derived from \cref{alg:i2csb}, we conducted unpaired and paired image-to-image translation initialized with Gaussian noise and corrupted inputs, respectively, alongside a Gaussian momentum prior. \cref{fig:rgb} demonstrates that our analytic posterior maintains superior stability across a wide hyperparameter window. Consistent with the trend observed for general CSB, the FID scores on CIFAR-10 also exhibited U-shaped dependence on the hyperparameter $\gamma$ (see \cref{tab:mnifid} in the appendix for the MNIST case). Because FID reflects domain discrepancy on the latent space under Gaussian assumptions, this empirical minimum confirms that per-step tuning of $\gamma$ enables more faithful diffusion trajectories covering target modalities.

\textbf{ImageNet256.}\enspace We extended our method to the ImageNet256 inpainting tasks, featuring higher dimensionality and various corruption methods. We fine-tuned the pretrained \textit{FreeForm1020}, \textit{FreeForm2030}, and \textit{Center} checkpoints from \citep{i2sb}. We replaced the training posterior with one based on the conformal geodesic. After 45K training iterations for each $\gamma\in\{10^{-4}, 10^{-3}, \dots, 10^1\}$, the I$^2$CSB algorithm was able to recover the performance of I$^2$SB in terms of the FID-10K score \citep{saharia2022palette}. This verifies that our method is applicable to diffusion bridge workflows involving RGB images.

\textbf{MRI.}\enspace We evaluated our method on simulated brain MRI slices at $64\!\times\!64$ resolution, with 24\,mm source--target separation. Our evaluation in \cref{tab:mri_summary} confirmed that our method achieved high-fidelity translation. \cref{fig:mri} compares translation results across DSBM \citep{dsbm}, I$^2$SB, and I$^2$CSB, illustrating qualitative improvements in both magnitude and phase reconstruction. Quantitative performance gains in FID and magnitude structural similarity index measure (SSIM) are summarized in \cref{tab:mri_summary}, where Ours (best) uses $\gamma=0.1$; full results are reported in \cref{tab:r5} of Appendix~\ref{sec:add_expr}. This confirms that our method holds a distinct advantage in handling explicit complex-valued data.

\subsection{Circuit quantum electrodynamics simulation}

\textbf{Closed systems.}\enspace We used a quantum physics simulator \citep{quantumtoolbox} to acquire trajectories of a harmonic oscillator in a system with $d=2$ modes. In this simulation, the system dynamics are governed by a Hamiltonian featuring local cavity detuning and a hopping interaction between the two modes:
\begingroup
\addtolength{\abovedisplayskip}{-1.8pt} 
\addtolength{\belowdisplayskip}{-1.8pt}
\begin{equation}\label{eq:osc}
  \hat{H}_\mathrm{osc} = D(\hat{a}^\dagger_1 \hat{a}_1) + D(\hat{a}^\dagger_2 \hat{a}_2) + J (\hat{a}^\dagger_1 \hat{a}_2 + \hat{a}_1 \hat{a}_2^\dagger),
\end{equation}
\endgroup
where $\hat{a}_j$ and $\hat{a}_j^\dagger$ are the bosonic annihilation and creation operators for each mode $j\in\{1,2\}$; $D$ denotes a detuning parameter; $J$ denotes inter-mode coupling strength. At each step, we extracted expectation values for the position and momentum quadratures as an observation with added fluctuation noise, using the quadrature definitions $\hat{x}_j = (\hat{a}_j + \hat{a}_j^\dagger) / 2$ and $\hat{p}_j = (\hat{a}_j - \hat{a}_j^\dagger) / (2i)$.

\cref{fig:qt}~(a) and \cref{fig:qt}~(b) show that particles' averaged trajectories (blue) given by the Wirtinger gradient of a potential neural network with 4 hidden layers successfully tracked the target trajectory (violet) for a long time period. The teal and green curves in \cref{fig:qt}~(a) show generalization to unseen high- and low-energy trajectories with different starting points; the model successfully generated harmonic motion. These results verify our hypothesis regarding a bosonic coherent state.

\textbf{3D qubit gate transformation.}\enspace We designed a physics-informed OT setup in this simulation problem domain. We expanded \cref{eq:osc} to approximate the dynamics of a hybrid tripartite cQED system. We designed a three-mode harmonic cavity coupled to a three-mode Josephson junction (JJ), which in turn interacted with a two-level qubit \citep{nakamura1999coherent}. The system evolution is governed by a total Hamiltonian formulated in the rotating frame:
  $\hat{H} = \hat{H}_\mathrm{osc} + \hat{H}_\mathrm{jj} + \hat{H}_\mathrm{qubit} +  \hat{H}_\mathrm{int}$.
We structured the simulations as source-to-target mapping tasks. Each algorithm was provided with learnable physics heads, \ie{} learnable differentials for the JJ operator and ground-truth qubit projection operators.

\begin{wraptable}[16]{r}{0.48\textwidth}
  \centering
  \caption{Qubit OT validation at higher, unseen energies for Hadamard and Pauli gates (5 runs).}\label{tab:highwave}
  \begin{tikzpicture}[tight background]
    \node[scale=0.7] at (0,0) {\begin{tabular}{c|c|c|c}
    \toprule
    \textbf{Hadamard} & $L_2$ ($\downarrow$) & EMD ($\downarrow$) & SWD ($\downarrow$) \\
    \midrule
    ODE (RK4) & $1.885_{\pm 0.20}$ & $3.702_{\pm 0.61}$ & $1.306_{\pm 0.25}$ \\
    OT-CFM & $1.910_{\pm 0.06}$ & $3.653_{\pm 0.18}$ & $1.094_{\pm 0.08}$ \\ 
    SB-CFM & $1.888_{\pm 0.063}$ & $3.621_{\pm 0.24}$ & $1.164_{\pm 0.08}$ \\
    Ours & $\mathbf{0.998_{\pm 0.61}}$ & $\mathbf{1.378_{\pm 1.18}}$ & $\mathbf{0.601_{\pm 0.33}}$ \\  
    \end{tabular}};
    \node[scale=0.7] at (0,-1.6) {\begin{tabular}{c|c|c|c}
    \toprule
    \textbf{Pauli-X} & $L_2$ ($\downarrow$) & EMD ($\downarrow$) & SWD ($\downarrow$) \\
    \midrule
    ODE (RK4) & $1.250_{\pm 0.45}$ & $1.850_{\pm 1.41}$ & $0.753_{\pm 0.28}$ \\
    OT-CFM & $\mathbf{0.557_{\pm 0.54}}$ & $0.610_{\pm 0.90}$ & $\mathbf{0.359_{\pm 0.318}}$ \\ 
    SB-CFM & $1.544_{\pm 0.107}$ & $2.404_{\pm 0.36}$ & $0.710_{\pm 0.40}$ \\
    Ours & $0.584_{\pm 0.25}$ & $\mathbf{0.538_{\pm 0.43}}$ & $0.393_{\pm 0.167}$ \\  
    \end{tabular}};
    \node[scale=0.7] at (0,-3.2) {\begin{tabular}{c|c|c|c}
    \toprule
    \textbf{Pauli-Y} & $L_2$ ($\downarrow$) & EMD ($\downarrow$) & SWD ($\downarrow$) \\
    \midrule
    ODE (RK4) & $1.232_{\pm 0.66}$ & $1.863_{\pm 1.34}$ & $0.738_{\pm 0.38}$ \\
    OT-CFM & $1.039_{\pm 0.58}$ & $1.315_{\pm 0.11}$ & $0.709_{\pm 0.52}$ \\ 
    SB-CFM & $1.213_{\pm 0.61}$ & $1.641_{\pm 1.08}$ & $0.671_{\pm 0.40}$ \\
    Ours & $\mathbf{0.741_{\pm 0.25}}$ & $\mathbf{0.838_{\pm 0.35}}$ & $\mathbf{0.601_{\pm 0.33}}$ \\  
    \bottomrule
    \end{tabular}};
  \end{tikzpicture}
\end{wraptable}

For benchmarks, we considered a neural ODE \citep{node} model with RK4 integration and conditional flow matching (CFM) \citep{tong2023improving} methods based on SB and OT solutions. When training the CFM variants and our model, we configured the methods to use mini-batch OT with cost matrices based on geodesic priors \citep{tong2023improving}. In particular, we utilized the conformal geodesic calculation to approximate ground-truth pairs. \cref{fig:qt}~(c) shows that, for an unseen energy level (doubling of the expected photon number from $1.0$ to $2.0$), our model successfully transformed one qubit state into another, mimicking a gate operation through the transport of particles in the oscillator. As shown in \cref{tab:highwave}, our approach outperformed the benchmarks in seven of the nine settings for cQED at unseen energies.

%% file: text/6_work.tex
\vspace{-1mm}
\section{Related Work}\label{sec:work}
\vspace{-1mm}

Diffusion bridges \citep{zhou2023denoising, dsb} have gained attention in recent years, with connections to score and flow matching \citep{i2sb, dsbm}. SBs were subsequently applied to jump measures~\cite{zlotchevski2024schr,de2026schr,zlotchevski2025schr}, second-order dynamics~\cite{tang2025entangled}, and structured data~\cite{howard2026schrdinger}, as well as extended to Riemannian manifolds~\cite{thornton2022riemannian}. Concurrently, \citet{noble2026twisted} studied reference processes modified by Feynman--Kac potentials, while \citet{mahmood2026geometric} applied SBs to rotating $\mathrm{SO}(2)$ geometries to control angular distributions. In contrast, we explore complex-valued dynamics through complex stochastic calculus and conformal geometry \citep{francesco2012conformal,del2020decoherence}, providing a new analytic posterior for downstream tasks.

Our work also connects to complex optimal transport \citep{cot}, a foundational problem in complex geometry. Through generalized dual structures \citep{introeot, han2025variational}, diffusion formulations attain a high level of abstraction and have been applied to topological structures \citep{yang2025topological}. Studies on complex-valued neural networks \citep{cvcnn, cvnn, hammad2024comprehensive} have considered both holomorphic and non-holomorphic network architectures and their backpropagation. Our Schr\"odinger potentials are complex-valued functions of $\vz$ and $\bar{\vz}$ that serve as dual parameters of distributions and need not be holomorphic.

The construction of complex-valued fractional Brownian motion \citep{jumarie1998new} and recent fractional diffusion bridge models \citep{nobisfractional} suggest extensions to complex-valued bridges with non-Markovian dynamics, which may be useful for modeling chaotic systems. Quantum Schr\"odinger bridges (QSBs) \citep{movilla2025quantum} connect density matrices using Kraus maps in a noncommutative operator setting. Furthermore, numerical methods for the Lindblad equation explicitly preserve complete positivity and trace conservation \citep{appelo2025kraus}. These works address density matrix evolution within quantum field theory, whereas our cQED experiments investigate diffusion bridge trajectories for bosonic states.

%% file: text/7_concl.tex
\vspace{-1mm}
\section{Conclusion}\label{sec:concl}
\vspace{-1mm}

In this paper, we presented complex Schr\"odinger bridges, a novel diffusion bridge formulation for solving the SBP in complex-valued domains. The proposed framework successfully modeled stochastic Hamiltonian physics and medical imaging, and also was able to perform existing diffusion-based generative workflows with improved stability. Our findings show that navigating complex spaces effectively requires structured geometric priors. We anticipate that our theoretical and empirical results will contribute to the broader field of AI for science (AI4Science).
 
\textbf{Limitations and future work.}\enspace 
Our work has the potential to broaden algorithmic capabilities for solving various OT problems. Nevertheless, the use of complex-valued arguments might not always be the optimal choice, especially if the domain lacks a general energy-based structure. Additionally, complex-valued neural architectures based on non-holomorphic functions remain underexplored, and incorporating them might have further consolidated our findings. Conformality itself does not guarantee the existence of bridges in the complex plane. The underlying complex structure becomes rigid as the dimension grows; thus, the mappings for OT must eventually be quasiconformal. In future work, we will extend this line of research to OT on general manifolds, which provide the geometric formulation of complex structures \citep{berman2014complex}, as well as to the matrix OT problem \citep{chen2017matrix}.

%% file: text/A_cop.tex
\vspace{-3.5mm}
\section{Complex Differential Operators}\label{sec:cop} 
\vspace{-2mm}
The complex number system \citep{conway2012functions, ash2014complex} consists of elements of the form $a+ib$, where $a,b\in\bR$ and $i$ denotes the imaginary unit. This field exhibits profound mathematical properties that distinguish it from the real numbers. In this section, we establish the foundational mathematical background required to formulate complex-valued differential equations and develop the Hopf--Cole transformation.

\vspace{-1mm}
\subsection{Background on complex differentiation}
\vspace{-1mm}

The concept of differentiation in the complex plane requires more stringent analytical conditions than its real-valued counterpart; that is, the existence of a complex derivative in a neighborhood is a remarkably strong condition. We define differentiability and holomorphic functions as follows.
\begin{definition}[Complex differentiability]
  Let $U \subset \bC$ be an open set. A function $f: U \to \bC$ is said to be (complex) differentiable at a point $z_0 \in U$ if the limit 
  \begingroup
  \addtolength{\abovedisplayskip}{-1pt}
  \addtolength{\belowdisplayskip}{-2pt}
  \begin{equation}
    f'(z_0) = \lim_{z \to z_0}\!\frac{f(z) - f(z_0)}{z - z_0}
  \end{equation}
  \endgroup
  exists regardless of the path along which $z$ approaches $z_0$. This limit is denoted by $f^\prime(z_0) = \left.\frac{\dd f}{\dd z}\right|_{z=z_0}$ and is called the complex derivative of $f$ at $z_0$.
\end{definition}
\begin{definition}[Holomorphic functions]
  Let $\cS \subseteq U$ be a nonempty open set. A function $f$ is called holomorphic (or analytic) on $\cS$ if $f$ is complex differentiable at every point $z_0 \in \cS$.
\end{definition}
Holomorphic functions are central objects in complex analysis \citep{rca} due to their analytic properties, such as a Taylor series representation around every point where the function is locally differentiable. Furthermore, complex differentiability is characterized by a fundamental system of partial differential equations known as the Cauchy--Riemann equations. 
\begin{lemma}[Cauchy--Riemann equations]
  Let $f: \bC \to \bC$ be defined as $f(z) = u(x,y) + i v(x,y)$, where $z = x+iy$ and $u, v \in \bR$. A necessary condition for $f$ to be analytic in a region $\cS$ is that $u$ and $v$ satisfy the Cauchy--Riemann equations \citep{krantz1999handbook, brown2009complex} throughout $\cS$:
  \begin{equation} \label{eq:cr}
    \frac{\partial u}{\partial x} = \frac{\partial v}{\partial y} \hspace{3em}\text{and}\hspace{3em} 
    \frac{\partial u}{\partial y} = -\frac{\partial v}{\partial x}.      
  \end{equation}
  Moreover, if these partial derivatives are continuous in $\cS$, then the Cauchy--Riemann equations serve as a sufficient condition for $f(z)$ to be analytic in $\cS$. 
\end{lemma}
In physics, \cref{eq:cr} is known as Hamilton's equation \citep{moc}, if we consider $z$ as an element of a $1$-dimensional combined position-momentum space. Lastly, we discuss integration of holomorphic functions.
\begin{definition}[Cauchy integral theorem]
  Let $f$ be a holomorphic function on a simply connected open region $\vD$. Then, for any piecewise smooth, closed contour $C$ contained entirely within $\vD$, the contour integral of $f(z)$ is zero:
  \begin{equation}
    \oint_C f(z) \dd z = 0.      
  \end{equation}
\end{definition}
A direct consequence of this theorem is that the value of an analytic function at any interior point of a region is completely determined by its values on the boundary.
\begin{lemma}[Cauchy integral formula]
  Let $f$ be holomorphic on an open set containing a closed contour $C$ and its interior. For an arbitrary $z_0$ inside $C$, the function evaluates to 
  \begin{equation}
    f(z_0) = \frac{1}{2\pi i} \oint_C \frac{f(z)}{z - z_0} \dd z,      
  \end{equation}
  where the contour of integration is traversed in the positive (counterclockwise) direction.
\end{lemma}
\begin{corollary} 
  Under the same conditions, the $n$-th derivative of $f$ at $z_0$ is given by
  \[
    f^{(n)}(z_0) = \frac{n!}{2\pi i} \oint_C \frac{f(z)}{(z - z_0)^{n+1}} \dd z.
  \]
\end{corollary} 
We further explore holomorphic functions, focusing specifically on their global properties across the complex plane. To establish the necessary foundation, we first define entire functions.
\begin{definition}[Entire functions]
  An entire function is a complex-valued function that is holomorphic at all finite points in the whole complex plane.
\end{definition}
An entire function is also called an integral function. Such functions are differentiable everywhere, and each has a power-series representation with an infinite radius of convergence. We next define boundedness of a function.
\begin{definition}[Boundedness]
A function $f$ is bounded if $\exists M \in \bR^+ : \lvert f(z)\rvert < M$.
\end{definition}

Liouville’s theorem establishes that every bounded entire function must be constant. Formally:
\begin{lemma}[Liouville’s theorem on holomorphicity]
\vspace{3mm}
\[
  f: \bC\to \bC\  \textrm{holomorphic}\ / \exists M \in \bR^+: \lvert f(z)\rvert < M,\ \forall z \in \bC \Rightarrow f = c,\ c \in \bC.
\]
\end{lemma}
Equivalently, any non-constant holomorphic function on $\bC$ must have an unbounded image. While mathematically elegant, this presents an unsatisfactory limitation from a practical standpoint, particularly in machine learning architectures. The unbounded behavior of purely analytic functions makes them notoriously difficult to train, typically requiring highly sophisticated architectures and specialized activation functions to effectively exploit their phase structure. Fortunately, our framework does not demand strict holomorphicity, as we opt to replace standard differential operators with Wirtinger derivatives. We observe that non-holomorphic complex functions are as important as holomorphic functions, especially when solving parabolic-type PDEs.

\subsection{Wirtinger calculus}\label{subsec:wderiv}

To describe complex PDEs and CVSDEs involving complex functions consistently and clearly, we use the following definition of the Wirtinger derivatives \citep{kreutz2009complex, koor2023short}.
\begin{definition}[Wirtinger derivatives]\label{def:wderiv}
  Let $f: U \to \bC$ be a $2$-dimensional smooth function (when viewed as a map $f: U\to\bR^2$). Let us consider the case of $f(z) = r(x,y) + i \kappa(x,y)$ with $(x,y) \mapsto x+i y = z$. Define the derivatives with respect to the variable $z$ and its conjugate $\bar{z}$ as
  \[
    \frac{\partial}{\partial z} \coloneqq \frac{1}{2}\biggl(\frac{\partial}{\partial x} - i \frac{\partial}{\partial y} \biggr) \hspace{3em} \text{and} \hspace{3em} \frac{\partial}{\partial \bar{z}} \coloneqq \frac{1}{2}\biggl(\frac{\partial}{\partial x} + i\frac{\partial}{\partial y} \biggr),
  \]
  where $\frac{\partial f}{\partial x} = \frac{\partial r}{\partial x} + i \frac{\partial \kappa}{\partial x}$ and $\frac{\partial f}{\partial y} = \frac{\partial r}{\partial y} + i \frac{\partial \kappa}{\partial y}$. If $f$ is holomorphic, the derivative $\frac{\partial f}{\partial z}$ agrees with the complex derivative. 
\end{definition}
By this definition, the operators $\frac{\partial}{\partial z}$ and $\frac{\partial}{\partial {\bar{z}}}$ are often referred to as the holomorphic and antiholomorphic derivatives, respectively, since the Cauchy--Riemann equations express the condition that the antiholomorphic derivative is zero.
\begin{remark}[Cauchy--Riemann condition]\label{rem:cauchy-riemann}
  $\tfrac{\partial f}{\partial \bar{z}} = 0$ iff $f$ is \emph{holomorphic}.
\end{remark}
The Wirtinger derivative operators, $\frac{\partial}{\partial z}$ and $\frac{\partial}{\partial \bar{z}}$, are inherently $\bC$-linear. This property ensures that standard Wirtinger calculus respects the complex structure of the domain by coupling the real and imaginary axes rather than treating them as independent real variables.
\begin{remark}[$\bC$-linearity]\label{rem:c_linearity}
  For any complex-valued smooth function $f$ and any complex constant $c \in \bC$ (such as the imaginary unit $i$), the operators satisfy:
  \[
    \frac{\partial}{\partial z} (cf) = c \frac{\partial f}{\partial z}, \qquad \text{and} \qquad \frac{\partial}{\partial \bar{z}} (cf) = c \frac{\partial f}{\partial \bar{z}}.
  \]
\end{remark}
By this linearity, the complex conjugate of the gradient equals the gradient of the conjugate. 
\begin{remark}[Conjugation]\label{rem:conj} 
  By linearity, it holds that $\tfrac{\partial \bar{f}}{\partial \bar{z}} = \overline{\bigl(\tfrac{\partial f}{\partial {z}}\bigr)}$.
\end{remark}

For a complex number $z = r e^{i\vartheta} \in \bC$, one can naturally write a complex logarithm as  
\begin{equation*}
  \log\lvert z\rvert + i\arg(z), \quad \arg(z) =\{\vartheta \mid \vartheta = \Arg(z) +2\pi k, \ k\in\bZ\},
\end{equation*}
where $\lvert z\rvert$ is the magnitude of the complex number and equals $\lvert r\rvert$; $\arg(z)$ is the set of all possible angles for $z$, obtained by periodic shifts of the principal angular value $\Arg(z) \in (-\pi, \pi]$. The multivalued complex logarithm can be visualized as an infinite helicoid embedded in $\bR^2$. To obtain an appropriate single-valued function in $\bC$, we can restrict the angle and denote the resulting function by
\begin{equation*}
  \Log(z) = \log\lvert z\rvert + i\Arg(z).
\end{equation*}
Instead of restricting $\vartheta$ to a single interval, let it range freely over $\bR$. One can then take the Riemann surface (the helicoid) as the domain of the function and define $\log: \cR \to \bC^\ast$ by
\begin{equation}\label{eq:clog}
  \log(z) \coloneqq \log \vert r\rvert + i\vartheta.
\end{equation}
In the definition \eqref{eq:clog}, the function $\log(\cdot)$ is single-valued and holomorphic on this surface, with well-defined behavior of its derivatives on the manifold. 

To the best of our knowledge, this work has introduced a novel approach to the complexification of the Hopf--Cole transform \citep{hopf, cole, leger2021hopf}, in which the potentials have two different interpretations through the introduction of the Hamiltonian dynamics coefficient $\gamma$ and the imaginary unit ``$i$'' respectively: 
\[
  \Psi = e^{\psi/\gamma}{\enspace\ }\text{and}\enspace\ \Phi = e^{i\chi}{\quad\ }\implies{\quad\ }\nabla_{\!\vz}\Psi = \frac{\Psi}{\gamma}\nabla_{\!\vz}\mkern1mu\psi \enspace\text{and}{\enspace\ }\nabla_{\!\bar\vz}\Phi = i\Phi\nabla_{\!\bar\vz}\chi,
\]
where $\Psi$ corresponds to the wave interpretation in the limit leading to the Schr\"odinger equation  considered in \cref{rem:se}, whereas $\Phi$ is primarily used in the CSB-PDEs for deriving forward and backward SDEs. For the above reasons regarding the complex logarithmic function, we define the relationship using the exponential function as an extension of the Madelung wave function \citep{leger2021hopf}. We treat $\psi$ as the complex logarithm of $\Psi$ constructed from $\chi$ and $c_\gamma$. The Wirtinger derivatives work consistently with these logarithmic potentials, akin to the log-derivative trick for real-valued functions. Lastly, in Wirtinger calculus, the standard Laplacian is expressed in the following interchangeable ways using dagger and dot product notation
\begin{equation}
  \frac{1}{4}\Delta = -\nabla_{\!\vz}^\dagger \nabla_{\!\vz} = \nabla_{\!\bar\vz}\!\boldsymbol{\cdot}\!\nabla_{\!\vz} = \nabla_{\!\vz}\!\boldsymbol{\cdot}\!\nabla_{\bar{\vz}} = -\nabla_{\!\bar\vz}^\dagger \nabla_{\!\bar\vz}  
\end{equation}
where the dagger denotes the formal adjoint $\nabla^\dagger_{\!\vz}=-\nabla^\sT_{\!\bar\vz}$ and $\nabla^\dagger_{\!\bar\vz}=-\nabla^\sT_{\!\vz}$. In this work, these properties of complex operators are leveraged to derive the CSB formulations succinctly. Note that a semantically equivalent real-valued formulation requires writing at least four times as many operations without the introduction of an analogous number system. Beyond providing a concise exposition, deriving the formulations via complex operators significantly reduces human error during implementation on machine learning hardware optimized for fast complex-number computation.

\subsection{Complex Hopf--Cole transform}\label{subsec:hopfcole}

For a Hamiltonian $H(\cdot,\cdot,\nabla_{\!\vz}\psi, \nabla_{\!\bar\vz}\psi)$, we assume that the general dynamic formulation is described with a generalized Schr\"odinger-type system for $t\in[0,T]$. This is derived by the Feynman--Kac formula in the real-valued problem. In our complex setting, we propose to write the system in the following alternative form:  
\begin{equation}\label{eq:mf}
  \begin{cases}
    i\partial_t \psi = H\mkern-1mu(\vz,\bar\vz, \!\nabla_{\!\vz}\psi,  \!\nabla_{\!\bar\vz}\psi, \rho) - \frac{i}{4}c_\gamma\Delta(\sigma^2 \psi), \\
    \partial_t \rho = i\nabla_{\!\bar\vz}^\dagger\mspace{-1mu}\bigl(\nabla_{\!\alpha}H\mkern-1mu(\vz, \bar\vz,\!\nabla_{\!\vz}\psi,\!\nabla_{\!\bar\vz}\psi, \rho)\,\rho\bigr) - i\nabla_{\!\vz}^\dagger\mspace{-1mu}\bigl(\overline{\nabla_{\!\alpha}H\mkern-1mu(\vz, \bar\vz, \!\nabla_{\!\vz}\psi,\!\nabla_{\!\bar\vz}\psi, \rho)}\,\rho\bigr) +\frac{1}{4}\Delta(\sigma^2 \mkern-2mu \rho),
  \end{cases}
\end{equation}
where the diffusion coefficient $\sigma=\sigma(t)$ is spatially constant. We take $\rho>0$ and impose $\psi+\bar\psi=\gamma\log\mkern-1.2mu\rho$ and $\nabla_{\!\vz}\boldsymbol{\cdot}f=\nabla_{\!\bar\vz}\boldsymbol{\cdot}\bar{f}$. The PDEs can be viewed as a blending of two optimality conditions of diffusion bridges; the first row indicates a Hamilton--Jacobi (HJ) equation, and the second row indicates a Fokker--Planck (FP) equation \citep{mfsdepde, deepgsb}. The parameter $\gamma\in\bR$ enters through $c_\gamma=1+i\gamma$, the logarithmic derivative $h^\prime_\gamma(\ell)/h_\gamma(\ell)$ of the spiral map, and through the density-dependent term in the Hamiltonian below. We use the complex Hopf--Cole transform as follows
\begingroup
\addtolength{\belowdisplayskip}{-1pt} 
\begin{subequations}\label{eq:hopfcole}
\begin{align}
  \Phi\bigl(z(t), \bar{z}(t), t\bigr) &= \exp\Bigl(\mfrac{i}{c_\gamma}\bigl(\chi\bigl(z(t), \bar{z}(t), t\bigr) + \gamma\log\mkern-1.2mu\rho\bigr)\mkern-1.2mu\Bigr),\enspace z(t) = \vz_t,\\ 
  \hat{\Phi}\bigl(z(t), \bar{z}(t), t\bigr) &= \rho\exp\Bigl(\mfrac{i}{c_\gamma}\bar\chi\bigl(z(t), \bar{z}(t), t\bigr)\mkern-1.2mu\Bigr).
\end{align}
\end{subequations}
\endgroup
where $\psi=\chi+\gamma\log\mkern-1.2mu\rho$, $\bar\chi=-\psi$ and $\Phi\hat\Phi=\rho$. Apparently, the construction of $\hat\Phi$ here is a relaxation of complex conjugation that preserves the measure, and the density $\rho(\vz, \bar{\vz}, t)$ is recovered from the product $\Phi\hat\Phi$. Note that, for potentials $Y = e^{i\psi/c}$ and $\hat{Y} = \rho e^{i\bar\chi/c}$, applying the multivariate calculus yields
\begin{align*}
  \nabla Y &= e^{i\psi/c}\bigl\{ic^{-1}\nabla\psi\bigr\}, \\
  \nabla\hat{Y} &= \hat{Y}\bigl\{ic^{-1}\nabla \bar\chi+\nabla\log\rho\bigr\},\\
  \nabla^\dagger\nabla Y &= e^{i\psi/c}\big\{c^{-2}(\bar{\nabla}\psi)^\sT(\nabla \psi) + ic^{-1}\nabla^\dagger\nabla \psi\bigr\},\\
  \nabla^\dagger\nabla \hat{Y} &= \begin{aligned}[t]
    \hat{Y}\bigl\{-c^{-2}(i\bar\nabla\bar\chi+c\bar\nabla\log\rho)^\sT
(i\nabla\bar\chi+c\nabla\log\rho) + {}
    ic^{-1}\nabla^\dagger\nabla \bar\chi + \nabla^\dagger\nabla\log\rho \bigr\},
  \end{aligned}
\end{align*}
and the equations hold for the gradient operators appearing in the PDE: $(\nabla, \bar{\nabla}) = (\nabla_{\!\vz},\nabla_{\!\bar\vz})$ and $(\nabla, \bar{\nabla}) = (\nabla_{\!\bar\vz},\nabla_{\!\vz})$; the first two identities also hold for $\nabla=\partial_t$. Hence, one shall carry out the following derivations regarding a twisted Hamiltonian, 
\begin{equation}\label{eq:twistedham}
  H(\cdot,\cdot,\nabla_{\!\vz}\psi, \nabla_{\!\bar\vz}\psi, \rho) \coloneqq \sigma^2\bigl\langle\nabla_{\!\vz}\psi, \nabla_{\!\vz}\bar\psi\bigr\rangle + (\nabla_{\!\vz}\psi)^\sT f - (\nabla_{\!\bar\vz}\psi)^\sT\bar{f} + \frac{i\gamma\sigma^2}{2}\frac{\Delta\sqrt{\rho}}{\sqrt{\rho}},
\end{equation}
where $\langle \alpha,\beta\rangle=\alpha^\sT\bar\beta$ denotes the standard Hermitian inner product. Substituting this Hamiltonian into the expression for the forward time derivative of $\Phi$, we get
\begin{flalign*}
  \qquad \frac{\partial \Phi}{\partial t} = \frac{i\Phi}{c_\gamma}\frac{\partial\psi}{\partial t}&= \frac{\Phi}{c_\gamma}\biggl(\!\sigma^2(\nabla_{\!\vz}\psi)^\sT\nabla_{\!\bar\vz}\psi + (\nabla_{\!\vz}\psi)^\sT f - (\nabla_{\!\bar\vz}\psi)^\sT\bar{f} - \frac{i\sigma^2}{4}c_\gamma\Delta\psi + \frac{i\gamma\sigma^2}{2}\frac{\Delta\sqrt{\rho}}{\sqrt{\rho}}\!\biggr)&&\\
  &= -i\nabla_{\!\vz}\Phi^\sT f + i\nabla_{\!\bar\vz}\Phi^\sT\bar{f} - \frac{\sigma^2}{4}c_\gamma\Delta\Phi + \frac{i\gamma\sigma^2}{2c_\gamma}\frac{\Delta\sqrt{\rho}}{\sqrt{\rho}}\Phi,&&
\end{flalign*}
and using $\nabla_{\!\alpha}H=f+\sigma^2\nabla_{\!\bar\vz}\psi$ in the density equation in \cref{eq:mf} with $\psi+\bar\psi=\gamma\log\rho$, we get  
\begin{flalign*}
  \qquad \frac{\partial \hat{\Phi}}{\partial t} &= \frac{1}{\Phi}\biggl(\frac{\partial\rho}{\partial t} - \hat\Phi\frac{\partial\Phi}{\partial t}\biggr)&&\\
  &= \frac{1}{\Phi}\biggl\{\!-i\nabla_{\!\vz}\boldsymbol{\cdot}(\rho f) + i\nabla_{\!\bar\vz}\boldsymbol{\cdot}(\rho\bar{f}) + \frac{\sigma^2}{4}c_\gamma\bigl(\Phi\Delta\hat\Phi - \hat\Phi\Delta\Phi\bigr) - \hat\Phi\frac{\partial\Phi}{\partial t}\!\biggr\}&&\\
  &= -i\nabla_{\!\vz}\mkern-0.5mu\boldsymbol{\cdot}(\hat\Phi f) + i\nabla_{\!\bar\vz}\mkern-0.5mu\boldsymbol{\cdot}(\hat\Phi\bar{f}) + \frac{\sigma^2}{4}c_\gamma\Delta\hat\Phi - \frac{i\gamma\sigma^2}{2c_\gamma}\frac{\Delta\sqrt{\rho}}{\sqrt{\rho}}\hat\Phi.&&
\end{flalign*}
These equations give the corresponding CSB-PDEs. Furthermore, the Schr\"odinger equation \cref{eq:se} naturally guides us to the time-reversed version \cref{eq:mf}. Using the conjugation of our Hermitian function $H^\ast(\vz, \bar\vz, \nabla_{\!\vz}\psi, \nabla_{\!\bar\vz}\psi) = \overline{H(\bar\vz, \vz, \nabla_{\!\bar\vz}\psi, \nabla_{\!\vz}\psi)}$, we obtain alternative McKean--Vlasov CVSDE
\begin{equation}
  \dd Z_\ss = i\mkern3.2mu\nabla_{\!\alpha} H\mkern-1mu\bigl(\mkern-1.2mu Z_\ss, \bar{Z}_\ss, \nabla_{\!\vz}\mkern1mu\bar\psi_\ss, \nabla_{\!\bar\vz}\mkern1mu\bar\psi_\ss, \rho_\ss \mkern-1.2mu\bigr) \dd s + \sigma \dd W_\ss.
\end{equation}
Undoubtedly, one can apply the corresponding Hopf--Cole transform akin to \cref{eq:hopfcole} with the given real-valued density $\rho = \bar{\rho}$. Therefore, the symmetric CVSDEs \cref{eq:cvsb} are both mean-field SDEs and distinct optimal control solutions under time reversibility. In addition, we leave the following remark that relates our McKean--Vlasov-type CVSDEs to the Schr\"odinger equation \cref{eq:se}. 
\begin{remark}[Schr\"odinger equation] \label{rem:se}
  Consider a one-parameter family $\{(\gamma_\tau, \sigma_\tau)\}_{\tau\ge0}$ satisfying $\sigma_\tau \to 0$ and $\gamma_\tau \sigma_\tau \to 1$ as $\tau \to \infty$. Under the hypotheses of the SDE existence and uniqueness theorem \citep{sde}, the rotated potential $\Psi = e^{\psi/\gamma_\tau}$ converges to the harmonic oscillator \cref{eq:se} if and only if $\rho_{\scriptscriptstyle0} = \rho_{\scriptscriptstyle T}$.
\end{remark}
Consequently, we argue that solving the CSB problem sits between modelling a pure rotational Hamiltonian of the Schr\"odinger equation and learning for higher-order optimal control subject to an energy function that comes from an affine Hamiltonian prior.

%% file: text/B_cmap.tex
\section{Conformal Mappings and Complex-Valued Stochastic Processes}\label{sec:cmap}

This appendix discusses relevant topics regarding conformal mapping and stochastic processes. 

\textbf{Conformal mappings.}\enspace The conformal mapping preserves angular magnitudes and orientations between curves, carrying essential geometric structure underlying sample spaces. Because conformal mappings preserve angles locally, two straight lines in a plane may bend in such a way as to preserve the angles at which they intersect. The notion of conformality is closely related to the holomorphic function and the Cauchy--Riemann equation where a nonconstant holomorphic function $h$ is guaranteed to be locally conformal in $\bC^{1}$.
\begin{lemma}[Riemann mapping theorem; Theorem~14.8 of \citet{rca}]\label{lem:rmt}
  Every nonempty simply connected domain $\cZ\subsetneq\bC$ admits a biholomorphism onto the open unit disc $\bD\coloneqq\{z\in\bC\mid\lvert z\rvert<1\}$, \ie{}, a bijective holomorphic map with a holomorphic inverse.
\end{lemma}
\begin{figure}
  \centering
  \def\xx{0.175}\def\yy{1.34}\def\ss{0.63}
  \tikzset{inner sep=0, outer sep=0}
  \begin{tikzpicture}[tight background]
    \begin{scope}
      \node at (0, 0) {
        \includegraphics[height=69pt]{
          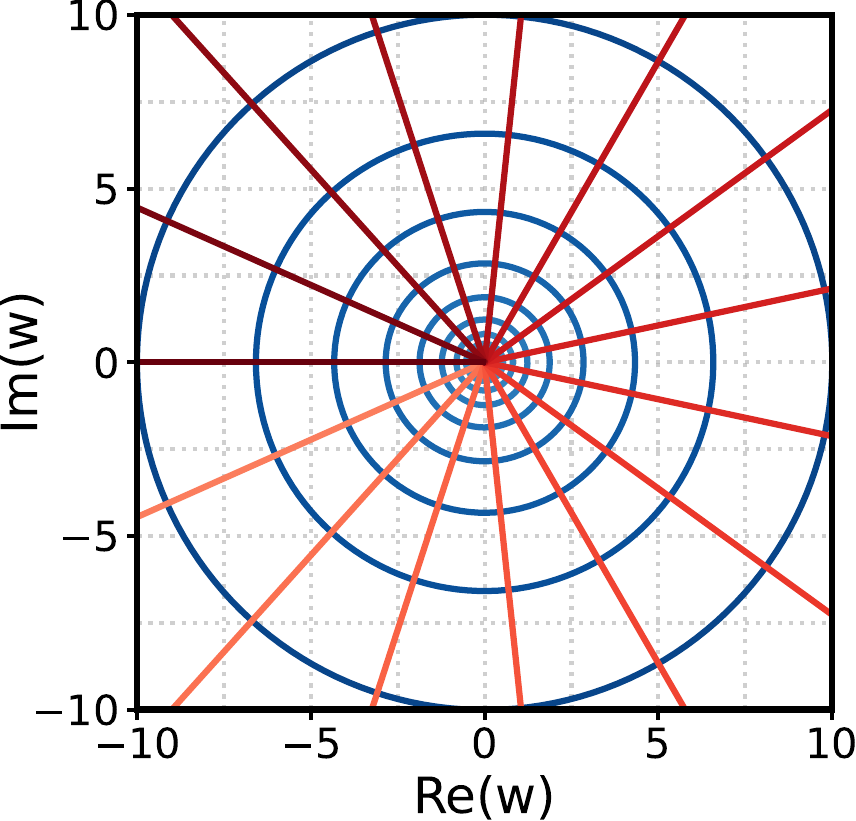}};          
      \node[scale=\ss] at (\xx, \yy) {$\mathsf{h(w) = e^w}$};
    \end{scope}
    \begin{scope}[xshift=3cm]
      \node at (0, 0) {
        \includegraphics[height=69pt]{
          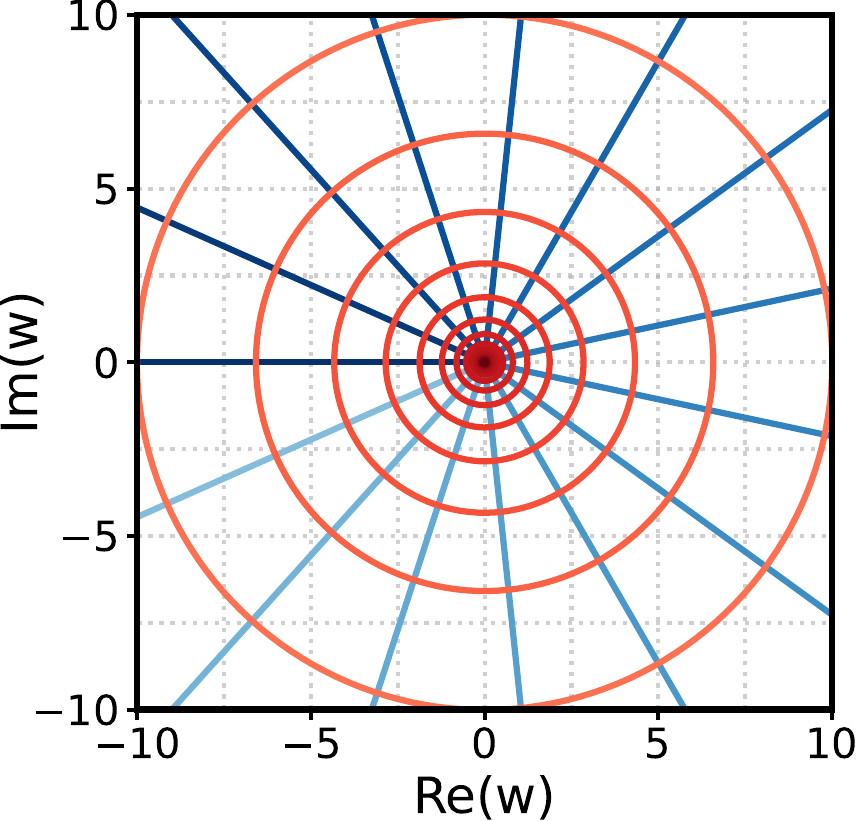}};          
      \node[scale=\ss] at (\xx, \yy) {$\mathsf{h(w) = e^{iw}}$};
    \end{scope}
    \begin{scope}[xshift=6cm]
      \node at (0, 0) {
        \includegraphics[height=69pt]{
          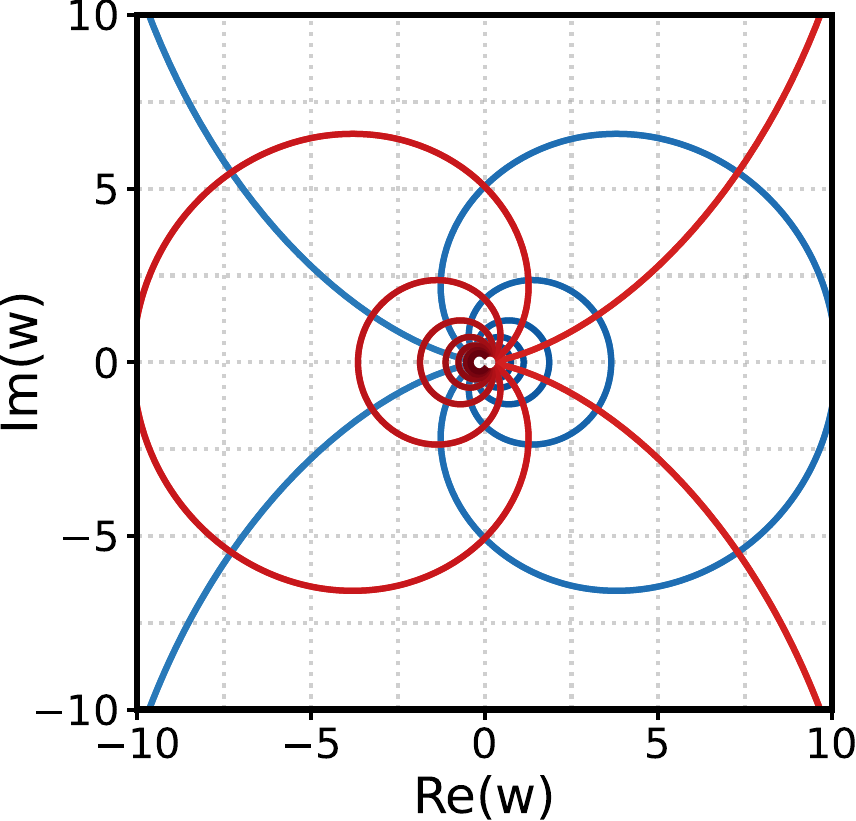}};
      \node[scale=\ss] at (\xx, \yy) {$\mathsf{h(w) = 4/w^2}$};
    \end{scope}
    \begin{scope}[xshift=9cm]
      \node at (0, 0) {
        \includegraphics[height=69pt]{
          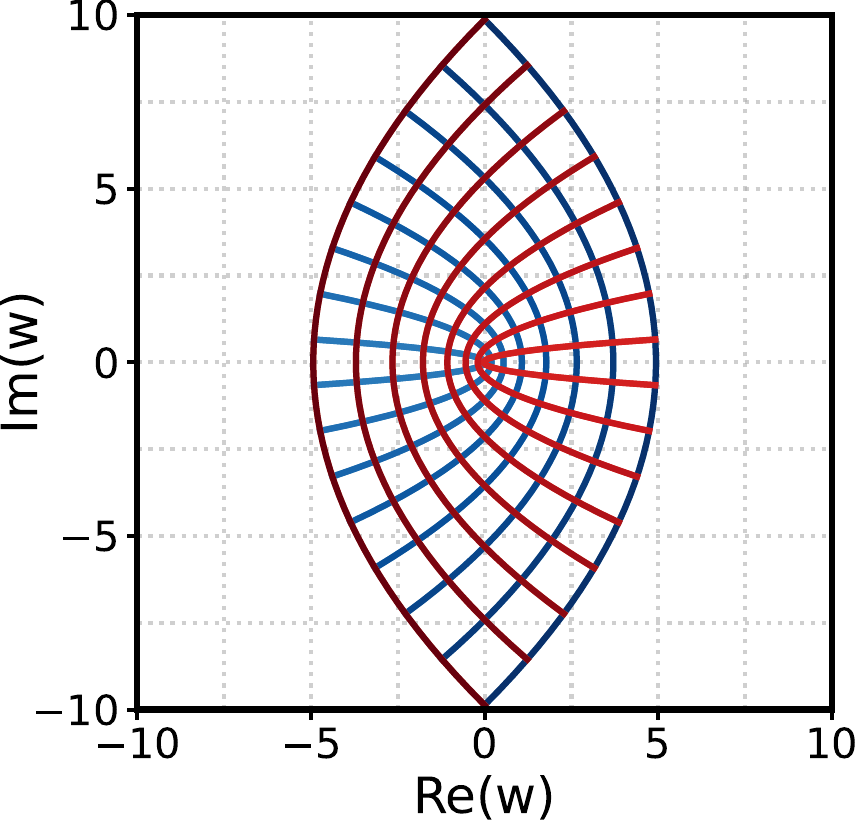}};
      \node[scale=\ss] at (\xx, \yy) {$\mathsf{h(w) = w^2/2}$};
    \end{scope}
    \path ([xshift=9.6pt]current bounding box.east);
  \end{tikzpicture}
  \vskip-3pt
  \caption{Conformal mappings in one complex dimension $\bC^1$ from \citet{moon2012field}.} \label{fig:confmap_others}
\end{figure}
Conformal mappings have been widely accepted in real two-dimensional ($d=1$) field problems \citep{moon2012field} and offer considerable flexibility in this setting. As various example functions in \cref{fig:confmap_others} demonstrate, the images of perpendicular Cartesian grid lines still meet at right angles. For a chosen origin $z_0\in\cZ$, the conditions $h(z_0)=0$ and $h^\prime(z_0)>0$ uniquely determine the Riemann map, as the following Schwarz lemma statement shows.
\begin{lemma}[Schwarz; Theorem~12.2 of \citet{rca}]\label{lem:schwarz}
  Let $h:\bD\to\bD$ be holomorphic with $h(0)=0$. Then $\lvert h(z)\rvert\le\lvert z\rvert$ for all $z\in\bD$ and $\lvert h^\prime(0)\rvert\le1$. If $\lvert h(z)\rvert=\lvert z\rvert$ for some nonzero $z\in\bD$ or $\lvert h^\prime(0)\rvert=1$, then $h(z)=az$ for a constant $a\in\bC$ with $\lvert a\rvert=1$.
\end{lemma}
Under the Schwarz lemma, simply $h^\prime(0)=1$ implies $h(z)=z$ in one complex dimension. As a higher-dimensional parallel, Cartan's uniqueness theorem extends this rigidity to bounded domains in $\bC^d$. In multidimensional settings, simply connected domains need not be biholomorphically equivalent, and biholomorphisms need not preserve Euclidean angles. In the following statements, $\vzero$ denotes a fixed point chosen as the origin and $\vJ_{\!f}$ denotes the holomorphic Jacobian matrix of $f$, explicitly defined with the Wirtinger operator as $\vJ_{\!f} \coloneqq [\nabla_{\!\vz}^\sT f_1, \nabla_{\!\vz}^\sT f_2, \cdots,\nabla_{\!\vz}^\sT f_d]^\sT$.
\begin{lemma}[Cartan; Theorem~1.3.1 of \citep{greene2011geometry}]
  Let $\cZ\subset\bC^d$ be bounded and $f:\cZ\to\cZ$ be a holomorphic function. If $f(\vzero)=\vzero$ and $\vJ_{\mkern-1.2muf}(\vzero)=\vI_d$ at some $\vzero\in\cZ$, then $f = \operatorname{id}_\cZ$ holds.
\end{lemma}
\begin{corollary}
  If $\cZ$ is a bounded domain in $\bC^d$ and $p \in \cZ$, then the mapping $f \mapsto \vJ_{\!f}(p)$ is an injective homomorphism of the isotropy group $\cQ_p \coloneqq \{ f \in \Aut(\cZ) : f(p) = p\}$ into $GL(d, \bC)$.
\end{corollary}
\begin{corollary} 
  The injective homomorphism $f \mapsto \vJ_{\!f}(p)$ of $\cQ_p$ onto $d\cQ_p \coloneqq \{ \vJ_{\!f}(p) : f \in \cQ_p \}$ is a homeomorphism of $\cQ_p$ (in the compact-open topology) onto a compact subgroup of $GL(d, \bC)$.
\end{corollary}
The uniqueness theorem and associated corollaries provided by \citet{greene2011geometry} imply that every automorphism fixing $\vzero$ ($h(\vzero) = \vzero$) is uniquely determined by its Jacobian at $\vzero$. They also imply that there is a Hermitian metric for which $\cQ_p$ acts isometrically on the tangent space at $p$ such that the corresponding map $f \mapsto \vJ_{\!f}(p)$ is also a homeomorphism onto its image, and that the matrix group $\vJ_{\!f}(p)$ is compact. Consequently, admitting a common conformal chart additionally requires $\vJ_h(\vz)\vJ^\dagger_h(\vz)=\Omega(\vz)^2\vI_d$ with a real field $\Omega$. 

\textbf{Conformal martingales.}\enspace Conformality plays a central role in formulating diffusions under generalized geometric priors, as it allows us to analyze diffusion bridge processes across both the planar and transformed domains in a unified manner. To characterize these processes, we first introduce the following definition of martingales.
\begin{definition}[Martingales]
  Let $(\cZ, \cF, P, \{\cF_t\}_{t\in I})$ for an ordered time set $I\subset[0,\infty)$ be a filtered probability space. A real-valued stochastic process $\{X_t\}_{t\in I}$ is a \emph{martingale} with respect to the filtration $\{\cF_t\}_{t\in I}$ if \one{} $X_t$ is $\cF_t$-measurable for all $t\in I$, \two{} $\bE\lVert X_t\rVert < \infty$ for all $t\in I$, and \three{} for all $\tau, t\in I$ with $\tau \le t$, the conditional expectation satisfies $\bE[X_t\vert \cF_\tau] = X_\tau$, $P$-almost surely. 
\end{definition}
A martingale is a process whose conditional expectation at any future time, given all past information, equals the process's current value. Notably, score-based diffusion models rely not only on the properties of BM, but also on Girsanov's theorem to govern changes of probability measure. Central to our framework is the following result on the conformal invariance of complex BM. Under proper normalization of time, $W_t$ is a proper complex Wiener process satisfying $\bE[\dd W_t\dd W_t^\dagger]=\vI_d\dd t$ and $\bE[\dd W_t\dd W_t^\sT]=0$.
\begin{proposition}[Conformal invariance; Theorem~V.2.5 of \citet{revuz2013continuous}]\label{prop:confinv}
  Let $h$ be an entire and non-constant function, and consider a complex Wiener process, \ie{}, $W_t = (\vw_t + i \hat\vw_t)/\sqrt{2}$ with independent standard real BMs $\vw$ and $\hat\vw$. Then, $\mathfrak{B}_t = h(W_t)$ is a standard BM with a time change. More precisely, there exist complex BMs $W$ and $\tilde{\mathfrak{B}}$ on the probability space such that 
  \begin{equation}
    h(W_t) = h(W_0) + \tilde{\mathfrak{B}}_{\langle Z, Z\rangle_t},
  \end{equation}
  where $\langle Z, Z\rangle_t = \int_0^t \lVert h^\prime (W_\tau) \rVert^2 \dd \tau$; $\langle Z, Z\rangle_\infty=\infty$ almost surely, with the clock denoted by $\langle Z, Z\rangle_t$.
\end{proposition} 
Note that the monotonically increasing term $\langle Z, Z\rangle_t$ serves as an alternative time scale for the transformed trajectories. Therefore, mappings preserve continuous conformal local martingales $(X+iY)$ up to domain exit, with $\langle X,X\rangle=\langle Y,Y\rangle$ and $\langle X,Y\rangle=0$; additional integrability conditions are required for true martingales.

\textbf{Quasiconformal mappings.}\enspace Due to the geometrical rigidity, every conformal mapping for $d\ge2$ reduces to a M\"obius transformation, a composition of basic operations---namely translations, scalings, rotations, reflections, and inversions. In this work, we resort to coordinatewise normalized operations, which are more flexible than global conformal mapping and are sufficient to support generative modeling and simulation under Hamiltonian dynamics. As a result, we relax the requirement to that of quasiconformal mappings \citep{ahlfors2006lectures}, orientation-preserving homeomorphisms between open sets. 
\begin{definition}[Quasiconformal mappings]
  Suppose that there is a smooth function $f: \cZ \to \cZ^\prime$ where $\cZ, \cZ^\prime \in \bC^1$. Then $f$ is quasiconformal if its partial derivatives satisfy the Beltrami equation
  \begin{equation}\label{eq:beltrami}
    \frac{\partial f}{\partial \bar{z}} = \mu(z) \frac{\partial f}{\partial z}
  \end{equation}
  for some complex-valued Lebesgue measurable function $\mu$ satisfying $\lvert\mu\rvert_\infty < 1$.
\end{definition}
The definition describes a $1$-dimensional complex map whose local distortion is prescribed by a complex-valued function $\mu$, called the Beltrami coefficient. The Beltrami equation is a first-order partial differential equation that can be viewed as a generalization of the Cauchy–Riemann equations.\footnote{Plugging in $\mu=0$ recovers the Cauchy--Riemann equations and also holomorphicity.}

Quasiconformal mappings admit a natural geometric interpretation related to quasihyperbolic planes. Suppose we equip $\cZ$ with the Riemannian metric 
\begin{equation}
  (\dd s)^2 = \Omega(z)^2\bigl\lvert \dd z + \mu(z) \dd \bar{z}\bigr\rvert^2
\end{equation}
where the conformal factor $\Omega(z) > 0$ is an arbitrary smooth function that sets the local length scale. Then a mapping $h$ satisfies \eqref{eq:beltrami} if and only if it is conformal from  $(\cZ, \dd s)$ to the target domain $\cZ^\prime$ endowed with the standard Euclidean metric; accordingly, $h$ is said to be $\mu$-conformal.

A quasiconformal map permits directional stretching with a uniformly bounded ratio: at any regular point of differentiability, its real differential maps infinitesimal circles to ellipses with an axis ratio bounded by the maximal dilatation $K=\frac{1+k}{1-k}$, $k=\lvert\mu\rvert_\infty$. By the measurable Riemann mapping theorem (the quasiconformal analogue of the Riemann mapping theorem), for every measurable Beltrami coefficient $\mu$ on $\bC$ satisfying $\lvert\mu\rvert_\infty<1$, there exists a quasiconformal homeomorphism that solves the Beltrami equation almost everywhere. This indicates even a discontinuous measurable field of admissible local distortions can be realized by a global map and such a homeomorphism is uniquely determined under the normalization.

Based on these details, this work considers a path-normalized quasiconformal homeomorphism $H_\gamma: C([0, T]; \bC^d)\to C([0, T]; \bC^d)$ in the uniform topology, where we take conformal spatial slices into consideration at fixed $t<T$ and a fixed endpoint:
\begin{equation}
  (\cH_\gamma \vz)_t = \vz_T +\frac{h_\gamma(\log(1-t/T))}{h_0(\log(1-t/T))}(\vz_t - \vz_T),\, \quad (\cH_\gamma \vz)_T = \vz_T.
\end{equation}
Apart from standard treatment using a quasihyperbolic metric \citep{koskela1998old, astala2015bilipschitz}, our construction is conditioned on the respective endpoint $\vz_T$ due to the marginal constraint $\bP_0 = \bP^\gamma_0 = \rho_{\scriptscriptstyle0}$ and $\bP_T = \bP^\gamma_T = \rho_{\scriptscriptstyle T}$. Still, it is a continuous, bijective function between two topological spaces that has a continuous inverse function, with properties such as $\cH^{-1}_\gamma = \cH_{-\gamma}$, $\cH_0 = \operatorname{id}$, and $\cH_{\gamma_1}\circ\cH_{\gamma_2}=\cH_{\gamma_1+\gamma_2}$ for $\gamma_1,\gamma_2\in\bR$.

\begin{figure}
  \centering
  \def\figh{145pt} \def\gaph{6pt}
  \tikzset{inner sep=0, outer sep=0}
  \begin{tikzpicture}[tight background] 
    \begin{scope}
      \node (Fig1) at (0,0) {
        \includegraphics[height=\figh]{
          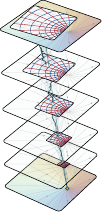}};
    \end{scope}
    \begin{scope}[xshift=2.85cm]
      \node (Fig2) at (0,0) {
        \includegraphics[height=\figh]{
          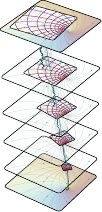}};
    \end{scope}
    \begin{scope}[xshift=5.7cm]
      \node (Fig3) at (0,0) {
        \includegraphics[height=\figh]{
          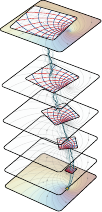}};
    \end{scope}
    \begin{scope}[xshift=8.55cm]
      \node (Fig4) at (0,0) {
        \includegraphics[height=\figh]{
          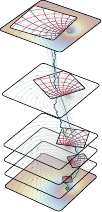}};
    \end{scope}

    \draw[thick,-{latex}]
      ([xshift=14pt]$(Fig4.north east)!0.1!(Fig4.south east)$)
      -- node[midway,right=\gaph] {$t$}
      ([xshift=14pt]$(Fig4.north east)!0.9!(Fig4.south east)$);
    \path ($(Fig1.west)+(Fig4.east)-(current bounding box.east)$);
     
    \begin{scope}[every node/.style={anchor=south},
                  font=\fontsize{10}{10}\selectfont]
      \node at ([yshift=\gaph]Fig1.north) {$\gamma=0.1$};
      \node at ([yshift=\gaph]Fig2.north) {$\gamma=0.5$};
      \node at ([yshift=\gaph]Fig3.north) {$\gamma=1.0$};
      \node at ([yshift=\gaph]Fig4.north) {$\gamma=2.5$};
    \end{scope}
  \end{tikzpicture}
  \caption{Conformal optimal transport geometry for $\gamma=\{0.1,0.5,1.0,2.5\}$.  Time increases downward. Increasing $\gamma$ produces greater angular winding through $e^{i\gamma\log(1-t)}$, while the radial contraction remains $1-t$, preserving the self-similarity of the transported geometry.}\label{fig:cot} 
\end{figure}

\cref{fig:cot} illustrates the self-similar geometry of COT in \eqref{eq:cot}. Since general conformality alone does not guarantee global invertibility, equivalence of path measures, or preservation of an arbitrary transport objective, we focus on this explicit similarity of our instance-based quasiconformal spiral mapping, which contains no reflection or folding. The time-dependent similarity in the figure describes its rotating contraction. Let the terminal time be $T=1$, and let $\vz\in\bC^d\simeq\bR^{2d}$. Conditional on a fixed endpoint $\vz_T$, our construction with $\gamma>0$ uses $\Upsilon_t^\gamma(\vz)=(\cH_\gamma\vz;\vz_T)_t,\,\Upsilon_t^\gamma(\vz)\coloneqq\vz_T+(1-t)^{1+i\gamma}(\vz-\vz_T)$, with $\Upsilon_1^\gamma(\vz)=\vz_T$ by continuity. Since $(1-t)^{1+i\gamma}=(1-t)e^{i\gamma\log(1-t)}$, the map is the identity at $t=0$ and a rotation followed by a strict contraction toward $\vz_T$ for $0<t<1$; for every fixed $t<1$, it is a global conformal similarity of $\bC^d\simeq\bR^{2d}$. For $\vz\ne\vz_T$, the spiral satisfies $\lVert\Upsilon_t^\gamma(\vz)-\vz_T\rVert=(1-t)\lVert\vz-\vz_T\rVert$. Its distance from the endpoint decreases strictly, so the trajectory cannot self-intersect before reaching $\vz_T$. Because the transformation fixes the boundary states, consistently transforming the reference measure, admissible measures, and path space yields equivalent SB problems. Likewise, for every fixed $t<1$ and a common fixed endpoint $\vz_T$, applying $\Upsilon_t^\gamma$ to both arguments of a quadratic transport cost multiplies all squared distances by $(1-t)^2$, leaving the minimizing coupling unchanged. 

Let $\bP^{\gamma,\ast}=(\cH_\gamma)_\#\bP$ be the optimal path law in
\cref{eq:klobj}, with terminal marginal $\rho_T$. For fixed $t<T$ and
terminal point $\vz_T$, write
\[
  \mu^{\delta(\vz_T)}_{\gamma,t}
  \coloneqq \frac{\partial_{\bar z}(\cH_\gamma)_t}
                 {\partial_z(\cH_\gamma)_t}
  = \frac{0}{(1-t/T)^{i\gamma}} = 0,
  \qquad
  \mu^{\rho_T}_{\gamma,t}(\dd\vz)
  \coloneqq (\bP^{\gamma,\ast})_t(\dd\vz)
  = \rho_t^\ast(\vz)\,\dd^{2d}\vz.
\]
Here $\delta(\vz_T)$ denotes the Dirac measure at $\vz_T$. The first quantity
is the Beltrami coefficient in one complex spatial coordinate. Its vanishing
shows that each fixed-time spatial slice is conformal for every $\gamma\in\bR$.
The second quantity is the time-$t$ marginal of the optimal path law, with
density $\rho_t^\ast$.

\textbf{It\^o formulas.}\enspace We recall the real and complex It\^o lemmas for stochastic differential equations.
\begin{lemma}[It\^o's formula \citep{itolemma}]\label{lem:ito}
  Let $\vx_t$ be the solution to a real-valued SDE:
  \begin{equation}
    \dd X_t = f(X_t, t)\dd t + g(X_t, t)\dd W_t.
  \end{equation}
  Then, the stochastic process $u(\vx_t, t)$, where $u\in C^{2,1}(\bR^d, [0, T])$, is also an It\^o process satisfying
  \begin{equation}\label{eq:itolemma}
  \begin{aligned}
    \rd u(X_t, t) = \frac{\partial u}{\partial t}\dd t + \Bigl\{&\nabla_{\!\vx}\mkern1.2mu u(X_t, t)^\sT f(X_t, t) +{} \\[-2pt]&  \mfrac{1}{2}\Tr\bigl[gg^{\!\ssT}\mspace{-1mu}(X_t, t)\mathop{\mathrm{Hess}} u(X_t,t)\bigr]  \Bigr\} \dd t + \bigl\{\nabla_{\!\vx}\mkern1.2mu u(X_t, t)^\sT g(X_t, t)\bigr\} \dd W_t
  \end{aligned}
  \end{equation}
  where $\nabla_{\!\vx}$ is the gradient operator with respect to $\vx_t$.
\end{lemma}

The complex It\^o lemma (also known as the quantum It\^o formula in the literature) \citep{ito1952complex, hudson1984quantum, freeman2015ito} describes the progression of complex variations. In the one-dimensional case, the formula gives the differential $\dd \varphi(z_t, \bar{z_t}, t)$ through a second-order Taylor expansion:
\[
  \dd \varphi = \frac{\partial \varphi}{\partial t} \dd t  + \frac{\partial \varphi}{\partial z} \dd Z_t + \frac{\partial \varphi}{\partial \bar{z}} \dd \bar{Z}_t + \frac{1}{2}\biggl(\frac{\partial^2 \varphi}{\partial z^2} (\dd Z_t)^2 + 2 \frac{\partial^2 \varphi}{\partial z \partial \bar{z}} \dd Z_t \dd \bar{Z}_t + \frac{\partial^2 \varphi}{\partial \bar{z}^2} (\dd \bar{Z}_t)^2 \biggr)
\]
The second-order terms (the quadratic variations) constitute the main difference between stochastic calculus and standard calculus. To evaluate $(\dd Z_t)^2$, $(\dd \bar{Z}_t)^2$, and $\dd Z_t\dd \bar{Z}_t$, we expand them using standard real-valued It\^o multiplication rules (where $\dd t^2 = 0$, $\dd t \dd W_t = 0$, and $\dd W^2_t= \dd t$) as follows:
\begin{align*}
  (\dd Z_t)^2 &= (\dd X_t)^2 - (\dd Y_t)^2 + 2 i\dd \langle  X, Y\rangle_t \\
  (\dd \bar{Z}_t)^2 &= (\dd X_t)^2 - (\dd Y_t)^2 - 2 i\dd  \langle X, Y\rangle_t \\
  (\dd Z_t) (\dd \bar{Z}_t) &= (\dd X_t)^2 + (\dd Y_t)^2 
\end{align*}
In the simple case of Brownian motion we can replace the quadratic It\^o correction term with the Laplacian by the following remark.
\begin{remark}[Complex quadratic variations]\label{rem:cpx_quad}
Let $W_t = (\vw_t + i \hat{\vw}_t)/\sqrt{2}$ be a standard complex Brownian motion constructed from two independent real Brownian motions $\vw_t$ and $\hat{\vw}_t$. Then $\dd\langle W, W\rangle_t = 0$. Let $Z_t$ be a stochastic process satisfying $\dd Z_t = \mu \dd t + \sigma \dd W_t$. A well-known consequence of their independence and equal variance is that the complex quadratic variation vanishes \citep{le2016brownian}
\[
  (\dd Z_t)^2 = (\mu \dd t+ \sigma \dd W_t )^2  = 0,
\]
and $(\dd \bar{Z}_t)^2 = 0$ also holds.
\end{remark}
As a natural consequence of \cref{rem:cpx_quad}, we state a multivariate complex It\^o formula herein.
\begin{lemma}[Complex It\^o formula \citep{ito1952complex}]\label{lem:cpxito}
  Let $Z_t$ be the solution to a CVSDE:
  \[
    \dd Z_t = \mu(Z_t, \bar{Z}_t, t) \dd t + \sigma \dd W_t
  \]
  where $W_t$ is the complex Wiener process and $\mu$ and $\sigma$ are sufficiently smooth. Then, the stochastic process defined by an arbitrary time-dependent smooth function $\varphi(\vz_t, \bar\vz_t, t)$ satisfies the SDE:
  \begin{align}
    \dd \varphi (Z_t, \bar{Z}_t, t) = &\biggl\{\!\frac{\partial \varphi}{\partial t} + \nabla_{\!\vz}\mkern1.2mu \varphi(Z_t, \bar{Z}_t, t)^\sT \mu(Z_t, \bar{Z}_t, t) + \nabla_{\!\bar\vz}\mkern1.2mu \varphi(Z_t, \bar{Z}_t, t)^\sT \bar{\mu}(\bar{Z}_t, Z_t, t)\!\biggr\} \dd t+{} \label{eq:cpxitolemma}\\&\Bigl\{\sigma^2\nabla_{\!\bar{\vz}}\!\boldsymbol{\cdot}\!\nabla_{\!\vz}\mkern1.2mu\varphi(Z_t, \bar{Z}_t, t)\Bigr\}\dd t
    + \sigma \nabla_{\!\vz}\mkern1.2mu \varphi(Z_t, \bar{Z}_t, t)^\sT \dd W_t + \sigma\nabla_{\!\bar{\vz}}\mkern1.2mu\varphi(Z_t, \bar{Z}_t, t)^\sT \dd \bar{W}_t \nonumber
  \end{align}
  where $(\nabla_\vz, \nabla_{\bar{\vz}})$ are multivariate Wirtinger gradient operators with respect to $(\vz, \bar{\vz})$. 
\end{lemma}
The time-dependent rotation in $\cH_\gamma$ modifies the drift and preserves the quadratic variation. In our general bridge setting, we explicitly use the definitions of both potentials and their mixing operator, with the resulting drift derived from the corrected complex It\^o formula.

%% file: text/C_proof.tex
\section{Proofs}\label{sec:proof}

We introduce a complete, multivariate derivation of SB-CVSDEs, which provide an explicit complexified formulation of dynamic Schr\"odinger bridge models. The derivation is motivated by relevant topics in statistical physics, differential equations and controls \citep{mckean1966class, vlasov1968vibrational}, as well as recent advances in score-based diffusion models and Schr\"odinger bridges \citep{sgmsde, likesb, deepgsb}. 

\subsection{Proof of Lemma~\texorpdfstring{\ref{lem:cld}}{1} (complex Langevin dynamics)}

With $\vz = \vx + i\vv$, let us write the scores, the real gradients of the log-density, as follows:
\[
  s_\vx(\vx, \vv) = \nabla_{\!\vx} \log p\mkern1.2mu(\vx, \vv)  \qquad \text{and} \qquad s_\vv(\vx, \vv) = \nabla_{\!\vv} \log p\mkern1.2mu(\vx, \vv). 
\]
This naturally yields
\begin{equation}\label{eq:sz}
  s_{{\vz}}(\vz, \bar{\vz}) = \nabla_{\!\vz} \log\rho\mkern1.2mu(\vz, \bar{\vz})  = \frac{s_\vx(\vx, \vv)}{2} - i\frac{\mkern1.2mus_\vv(\vx,\vv)}{2} 
\end{equation}
One can naturally define their Wirtinger calculus counterparts as 
\begin{equation}\label{eq:sbarz}
  s_{\bar{\vz}}(\vz, \bar{\vz}) = \nabla_{\!\bar\vz} \log\rho\mkern1.2mu(\vz, \bar{\vz})  = \frac{s_\vx(\vx, \vv)}{2} + i\frac{\mkern1.2mus_\vv(\vx,\vv)}{2} 
\end{equation}
Then, the suggested forward drift coefficient is written with the Wirtinger gradients \eqref{eq:sbarz} as 
\[
  i\sigma^2\nabla_{\!\bar\vz}\mkern1mu\psi = i\sigma^2\frac{\gamma s_{\bar{\vz}}-i\mkern1.2mu s_{\bar{\vz}}}{2} = i\sigma^2\biggl(\gamma \frac{s_{\mkern-1.2mu\vx} + is_{\mkern-1.2mu\vv}}{4} - i\frac{s_{\mkern-1.2mu\vx} + is_{\mkern-1.2mu\vv}}{4}\biggr) = \sigma^2\frac{ s_{\mkern-1.2mu\vx} - \gamma s_{\mkern-1.2mu\vv}}{4} +i\sigma^2\frac{\gamma s_{\mkern-1.2mu\vx} + s_{\mkern-1.2mu\vv}}{4},
\]
with omitted inputs $(\vz, \bar\vz)$. Therefore, the SDE for $[\vx_t, \vv_t]^\sT$ takes the form 
\[
  \begin{bmatrix}\dd\vx_t \\ \dd\vv_t\end{bmatrix} = \Biggl(\underbrace{\frac{\sigma^2}{4}\!\begin{bmatrix}
    \nabla_{\vx}\log p (\vx_t, \vv_t)\\
    \nabla_{\vv}\log p (\vx_t, \vv_t) 
  \end{bmatrix}}_{\text{Contraction}}\mkern-1.2mu + 
    \underbrace{\frac{\sigma^2}{4}\gamma\!\begin{bmatrix}
      -\nabla_{\vv}\log p(\vx_t, \vv_t) \\
      \nabla_{\vx}\log p(\vx_t, \vv_t)
  \end{bmatrix}}_{\text{Forward dynamics}}\Biggr)\dd t + 
    \underbrace{
    \frac{\sigma}{\sqrt{2}}\!\begin{bmatrix}
      \dd \vw_t\\
      \dd \hat{\vw}_t
  \end{bmatrix}}_{\text{Diffusion}}.
\]
For each $\vx$ and $\vv$, this recovers Langevin dynamics using a real-valued score function $\log p(\vx, \vv) \propto -E(\vx, \vv)$ for some energy $E(\cdot, \cdot)$. 

Conversely, for $[\vx_t, -\vv_t]^\sT$ and $s_\vz = \nabla_{\!\vz}\log\rho$, applying \eqref{eq:sz} yields
\[
\begin{aligned}
  i\sigma^2\nabla_{\!\vz}\mkern1mu\psi &= i\sigma^2\frac{\gamma s_{\mkern-1.2mu\vz}-i\mkern1.2mus_{\mkern-1.2mu\vz}}{2} = i\sigma^2 \biggl( \frac{\gamma s_{\mkern-1.2mu\vx}- s_{\mkern-1.2mu\vv}}{4} - i\frac{ s_{\mkern-1.2mu\vx} + \gamma s_{\mkern-1.2mu\vv}}{4} \biggr) \\
  &= \sigma^2\frac{ s_{\mkern-1.2mu\vx} + \gamma s_{\mkern-1.2mu\vv}}{4} +i\sigma^2 \frac{\gamma s_{\mkern-1.2mu\vx} - s_{\mkern-1.2mu\vv}}{4} = \sigma^2\frac{ \cev{s}_{\mkern-1.2mu\vx} - \gamma \cev{s}_{\mkern-1.2mu-\vv}}{4} +i\sigma^2 \frac{\gamma \cev{s}_{\mkern-1.2mu\vx} + \cev{s}_{\mkern-1.2mu-\vv}}{4},
\end{aligned}
\]
where $\cev{s}_\vx = \nabla_\vx\log\cev{p}(\vx, -\vv)$ and $\cev{s}_{-\vv} = \nabla_{(-\vv)}\log\cev{p}(\vx, -\vv) = -s_\vv$ denote the score functions of the reversed time process. Thus, this yields
\[ 
  \begin{bmatrix}\mkern14mu \dd\vx_t \\ -\dd\vv_t\end{bmatrix} =  
  \Biggl(\frac{\sigma^2}{4}\!\begin{bmatrix}
    \nabla_\vx \log \cev{p}(\vx_t, -\vv_t)\\
    \nabla_{(-\vv)} \log \cev{p}(\vx_t, -\vv_t)\\
  \end{bmatrix}\mkern-1.2mu +
  \frac{\sigma^2}{4}\gamma \!\begin{bmatrix}
    -\nabla_{(-\vv)} \log \cev{p}(\vx_t, -\vv_t)\\
    \nabla_\vx \log \cev{p}(\vx_t, -\vv_t) 
  \end{bmatrix}\Biggr)\dd s +
  \frac{\sigma}{\sqrt{2}}\!\begin{bmatrix}
    \dd \vw_\ss\\
    \dd \hat{\vw}_\ss
  \end{bmatrix}
\]
The proof is completed by considering ergodicity and the time reversibility of Hamiltonian Markov chains \citep{girolami2011riemann}. \QED{}

\subsection{Proof of Proposition~\texorpdfstring{\ref{prop:csb-pde}}{1} (CSB-PDEs)}

For a time-dependent smooth complex function $\psi$ under the conditions of Appendix~\ref{subsec:hopfcole}, consider the MF Hamiltonian:
\begin{equation}\label{eq:mf-hamiltonian}
  H(\vz, \bar\vz, \nabla_{\!\vz} \psi, \nabla_{\!\bar\vz}  \psi, \rho, \sigma) = \sigma^2\bigl\langle\nabla_{\!\vz}\psi, \nabla_{\!\vz}\bar\psi\bigr\rangle + (\nabla_{\!\vz}\psi)^\sT f - (\nabla_{\!\bar\vz}\psi)^\sT\bar{f} + \frac{i\gamma\sigma^2}{2}\frac{\Delta\sqrt{\rho}}{\sqrt{\rho}}.
\end{equation}
The potentials $\Phi, \hat{\Phi}\in C^{2,1}(\bC^d, [0,T])$ are parameterized via the complex Hopf--Cole transform~\eqref{eq:hopfcole} in Appendix~\ref{subsec:hopfcole} for a positive, real-valued density $\rho_t$ satisfying $\rho_{\scriptscriptstyle t} = \Phi(\cdot,\cdot,t)\hat{\Phi}(\cdot,\cdot,t)$. Applying the results from Appendix~\ref{subsec:hopfcole}, with $F$ added to the HJ equation, we derive $(\frac{\partial \Phi}{\partial t}, \frac{\partial \hat{\Phi}}{\partial t})$. Substituting both equations, we obtain the following PDEs:
\begingroup
\addtolength{\abovedisplayskip}{-1pt}
\addtolength{\belowdisplayskip}{-1pt}
\begin{subequations}\label{eq:csb-pde_display}
  \begin{empheq}[left={\empheqlbrace\mspace{3mu}}]{align}
    \frac{\partial \Phi\mkern-1mu(\vz,\bar{\vz}, t)}{\partial t} &= -i\nabla_\vz\Phi^\sT f + i\nabla_{\!\bar{\vz}}\Phi^\sT \bar{f} - \frac{\sigma^2}{4}c_\gamma\Delta\Phi + \frac{i\gamma\sigma^2}{2c_\gamma}\frac{\Delta\sqrt{\rho}}{\sqrt{\rho}}\Phi + c_\gamma^{-1}F\Phi, \label{subeq:csb-pde_fwd} \\[-2pt]
    \frac{\partial \hat{\Phi}\mkern-1mu(\vz, \bar{\vz}, t)}{\partial t} &= -i\nabla_{\!\vz}\boldsymbol{\cdot}(\hat{\Phi}f)+ i\nabla_{\!\bar{\vz}}\boldsymbol{\cdot}(\hat{\Phi}\bar{f}) + \frac{\sigma^2}{4}c_\gamma\Delta{\hat{\Phi}} - \frac{i\gamma\sigma^2}{2c_\gamma}\frac{\Delta\sqrt{\rho}}{\sqrt{\rho}}\hat{\Phi} - c_\gamma^{-1}F\hat{\Phi} \label{subeq:csb-pde_bwd}
  \end{empheq}
\end{subequations}
\endgroup
which hold for $t\in[0,T]$, where $\Delta$ denotes the Laplacian and $F$ is a real-valued functional modeling the mean-field interaction. Furthermore, $\hat{\Phi}$ is defined by $\rho/\Phi$, and the boundary conditions $(\Phi\hat{\Phi})(\,\cdot\mkern1.2mu, \,\cdot\mkern1.2mu, 0) = \rho_0$ and $(\Phi\hat{\Phi})(\,\cdot\mkern1.2mu, \,\cdot\mkern1.2mu, T) = \rho_T$ are imposed. This completes the proof. \QED{}

For the reverse Hopf--Cole transform, denote the conjugate of the backward potential in \eqref{eq:bwd} by $\varphi\coloneqq\bar\eta+\gamma\log\mkern-1.2mu\rho=-\eta$. In the reversed time coordinate $s=T-t\in[0,T]$, the Hopf--Cole factors are
\begin{subequations}\label{eq:hopfcole-rev}
\begin{align}
  \Phi^\prime(\vz_\ss, \bar{\vz}_\ss, s) &= \rho\mkern.5mu\exp\Bigl(-\mfrac{i}{c_\gamma}\bar\varphi(\vz_\ss, \bar{\vz}_\ss, T-s)\mkern-1.2mu\Bigr),\\
  \hat{\Phi}^\prime(\vz_\ss, \bar{\vz}_\ss, s) &= \exp\Bigl(\mfrac{i}{c_\gamma}\bar\varphi(\vz_\ss, \bar{\vz}_\ss, T-s)\mkern-1.2mu\Bigr),
\end{align}
\end{subequations}
so $\Phi^\prime\hat\Phi^\prime=\rho$. The primed factors denote the Hopf--Cole pair in \eqref{eq:csb-pde_display} reindexed by $t=T-s$, with all coefficients evaluated at $t=T-s$. For $F=0$, applying $\partial_s=-\partial_t$ and the drift-divergence condition constructed in Appendix~\ref{subsec:hopfcole} gives us a system of reversed-time PDEs
\begin{subequations}
\begin{align}
  \frac{\partial \hat\Phi^\prime}{\partial s} &= -i\nabla_{\!\bar\vz}\hat\Phi^{\prime\sT} \bar{f} + i\nabla_{\!\vz}\hat\Phi^{\prime\sT} f - \frac{\sigma^2}{4}c_\gamma\Delta\hat\Phi^\prime + \frac{i\gamma\sigma^2}{2c_\gamma}\frac{\Delta\sqrt{\rho}}{\sqrt{\rho}}\hat\Phi^\prime,\\[-2pt]
  \frac{\partial \Phi^\prime}{\partial s} &= -i\nabla_{\!\bar\vz}\!\boldsymbol{\cdot}\mkern-1.2mu(\Phi^\prime \bar{f}) + i\nabla_{\!\vz}\!\boldsymbol{\cdot}\mkern-1.2mu(\Phi^\prime f) + \frac{\sigma^2}{4}c_\gamma\Delta\Phi^\prime - \frac{i\gamma\sigma^2}{2c_\gamma}\frac{\Delta\sqrt{\rho}}{\sqrt{\rho}}\Phi^\prime.
\end{align}
\end{subequations}
This yields the time-reversed system as the following corollary.
\begin{corollary}\label{cor:csb-pde-rev} For the MF Hamiltonian \cref{eq:mf-hamiltonian}, applying the Hopf--Cole transform \cref{eq:hopfcole-rev} in the reversed time coordinate yields the following PDEs:
\begingroup
\addtolength{\abovedisplayskip}{-1pt}
\addtolength{\belowdisplayskip}{-1pt}
\begin{subequations}\label{eq:csb-pde-rev_display}
  \begin{empheq}[left={\empheqlbrace\mspace{3mu}}]{align}
    \frac{\partial \hat\Phi^\prime\mkern-1mu(\vz,\bar{\vz}, s)}{\partial s} &= -i\nabla_{\!\bar\vz}\hat\Phi^{\prime\sT} \bar{f} + i\nabla_{\!\vz}\hat\Phi^{\prime\sT} f - \frac{\sigma^2}{4}c_\gamma\Delta\hat\Phi^\prime + \frac{i\gamma\sigma^2}{2c_\gamma}\frac{\Delta\sqrt{\rho}}{\sqrt{\rho}}\hat\Phi^\prime + c_\gamma^{-1}F\hat\Phi^\prime, \label{subeq:csb-pde-rev_fwd}  \\[-2pt]
    \frac{\partial\Phi^\prime\mkern-1mu(\vz, \bar{\vz}, s)}{\partial s} &= -i\nabla_{\!\bar\vz}\!\boldsymbol{\cdot}\mkern-1.2mu(\Phi^\prime \bar{f})+ i\nabla_{\!\vz}\!\boldsymbol{\cdot}\mkern-1.2mu(\Phi^\prime f) + \frac{\sigma^2}{4}c_\gamma\Delta{\Phi^\prime} - \frac{i\gamma\sigma^2}{2c_\gamma}\frac{\Delta\sqrt{\rho}}{\sqrt{\rho}}\Phi^\prime - c_\gamma^{-1}F\Phi^\prime, \label{subeq:csb-pde-rev_bwd}
  \end{empheq}
\end{subequations}
\endgroup
where the operator $\nabla_{\!\bar\vz}$ here denotes the Wirtinger gradient of $\hat{\Phi}^\prime(\vz_\ss, \bar{\vz}_\ss, s)$ according to \cref{def:multiwderv}, \ie{} differentiation with respect to  $\bar{\vz}_\ss$.  
\end{corollary}
Utilizing two PDE systems of \cref{prop:csb-pde} and \cref{cor:csb-pde-rev}, we can now derive SB-CVSDEs in the following proof of \cref{thm:sb-cvsde}.

\subsection{Proof of Theorem~\texorpdfstring{\ref{thm:sb-cvsde}}{1} (SB-CVSDEs)}\label{subsec:sb-cvsde_proof}

We first state an expanded version of the SB theorem for the reverse time coordinate $s$. 
\begingroup
\renewcommand{\thetheorem}{\ref{thm:sb-cvsde}}
\begin{theorem}[SB-CVSDEs]  
  Suppose that $(Y_t, \widehat{Y}_t)$ are the potentials with $Y_t \coloneqq \chi(\vz_t, \bar{\vz}_t, t)$ and $\widehat{Y}_t \coloneqq -\eta(\vz_t, \bar{\vz}_t, t)$. Let the base drift $f$ be analytic and satisfy $\nabla_{\!\vz}\!\boldsymbol{\cdot}\!f_t = \nabla_{\!\bar\vz}\!\boldsymbol{\cdot}\!\bar{f}_t$. Then, the CVSDEs corresponding to the CSB-PDE in \cref{prop:csb-pde} for the evolution of \cref{eq:fwd} are given by  
  \begingroup
  \addtolength{\abovedisplayskip}{-1.5pt}
  \addtolength{\belowdisplayskip}{-1.5pt}
  \begin{subequations}\label{appeq:cvsde1}
    \begin{empheq}[left={\begin{gathered}\textrm{SB}\\[-0.4em]{\textrm{\small CVSDEs}}\\[-0.4em]\textrm{wrt \eqref{eq:fwd}}\end{gathered}\mspace{-6mu}:
    \empheqlbrace}]{align}
      \dd Z_t &= i\mkern.5mu\bigl(f_t+\sigma^2\nabla_{\!\bar\vz}\mkern1.2mu\psi_t\bigr) \dd t + \sigma \dd W_{t}, \label{appeq:cvsde1a}\\ 
      \dd Y_t &= -i\mkern.5mu\bigl(\mkern-1.2mu \lVert v_t \mkern-0.5mu\rVert^2 + F_t + \gamma\sigma\nabla_{\!\vz}\!\boldsymbol{\cdot}\mkern-1.2mu \check{v}_t - \tfrac{i\gamma}{2}\lVert \hat{v}_t\!+\!\check{v}_t\mkern-0.5mu\rVert^2 \mkern-1.2mu\bigr) \dd t + v^\ssT_t \dd W_t + \hat{v}^\ssT_t \dd \bar{W}_t,\label{appeq:cvsde1b}\\
      \dd\widehat{Y}_t &= \begin{aligned}[t]
        -i\mkern.5mu &\bigl(\lVert v_t\mkern-0.5mu\rVert^2 + F_t - \gamma\sigma\nabla_{\!\vz}\!\boldsymbol{\cdot}\mkern-1.2mu \hat{v}_t - (1\!+\!\tfrac{i\gamma}{2})\lVert \hat{v}_t\!+\!\check{v}_t\mkern-0.5mu\rVert^2\mkern-1.2mu\bigr) \dd t - {}\\
        &\bigl(\sigma\nabla_{\!\vz}\!\boldsymbol{\cdot}\mkern-1.2mu\check{v}_t + \sigma\nabla_{\!\bar\vz}\!\boldsymbol{\cdot}\mkern-1.2mu\bar{v}_t \mkern-0.5mu\bigr) \dd t - \bar{v}^\ssT_t\dd W_t - \check{v}^\ssT_t\dd \bar{W}_t,
      \end{aligned} \label{appeq:cvsde1c}
    \end{empheq}
  \end{subequations}
  \endgroup
  where $\psi=\chi+\gamma\log\mkern-1.2mu\rho$, $\Phi=e^{i\psi/c_\gamma}$, $\hat{\Phi}=\rho\mkern.5mu e^{i\bar\chi/c_\gamma}=e^{i\eta/c_\gamma}$ and $\Phi\hat{\Phi}=\rho$; $(v_t, \hat{v}_t, \check{v}_t, \bar{v}_t)$ denote the Wirtinger gradients for $\chi$ and $\eta$  
  \[
    v_t \coloneqq \sigma\nabla_{\!\vz}\mkern1mu\chi(\vz_t, \bar{\vz}_t, t),\enspace \hat{v}_t \coloneqq \sigma\nabla_{\!\bar\vz}\mkern1mu \chi(\vz_t, \bar{\vz}_t, t), \enspace  
    \check{v}_t \coloneqq \sigma\nabla_{\!\bar\vz}\mkern1mu\eta(\vz_t, \bar{\vz}_t, t), \enspace
    \bar{v}_t \coloneqq \sigma\nabla_{\!\vz}\mkern1mu\eta(\vz_t, \bar{\vz}_t, t). 
  \]
  Conversely, conjugating \eqref{eq:bwd} gives the evolution of $Z_\ss$ in increasing reverse time $s=T-t$. The CSB-PDEs in \cref{cor:csb-pde-rev} then yield
  \begingroup
  \addtolength{\abovedisplayskip}{-1.5pt}
  \addtolength{\belowdisplayskip}{-1.5pt}
  \begin{subequations}\label{appeq:cvsde2}
  \begin{empheq}[left={\begin{gathered}\textrm{SB}\\[-0.4em]{\textrm{\small CVSDEs}}\\[-0.4em]\textrm{wrt \eqref{eq:bwd}
  }\end{gathered}\mspace{-6mu}:\empheqlbrace}]{align}
    \dd Z_\ss &= i\mkern.5mu\bigl(-f_\ss - \sigma^2\nabla_{\!\bar\vz}\mkern1.2mu\varphi_\ss\bigr) \dd s + \sigma \dd W_\ss, \label{appeq:cvsde2a}\\ 
    \dd\widehat{Y}_\ss &= i\mkern.5mu\bigl(\mkern-1.2mu \lVert \check{v}_\ss\mkern-0.5mu\rVert^2 + F_\ss + \gamma\sigma\nabla_{\!\bar\vz}\!\boldsymbol{\cdot}\mkern-1.2mu v_\ss - \tfrac{i\gamma}{2}\lVert v_\ss\!+\!\bar{v}_\ss\mkern-0.5mu\rVert^2 \mkern-1.2mu\bigr) \dd s - \check{v}^\ssT_\ss \dd \bar{W}_\ss - \bar{v}^\ssT_\ss \dd W_\ss,\label{appeq:cvsde2b}\\
    \dd Y_\ss &= \begin{aligned}[t]
      i &\bigl(\lVert \check{v}_\ss\mkern-0.5mu\rVert^2 + F_\ss - \gamma\sigma\nabla_{\!\vz}\!\boldsymbol{\cdot}\mkern-1.2mu \check{v}_\ss - (1\!+\!\tfrac{i\gamma}{2})\lVert v_\ss\!+\!\bar{v}_\ss\mkern-0.5mu\rVert^2\mkern-1.2mu\bigr)\dd s + {}\\
      &\bigl(\sigma\nabla_{\!\vz}\!\boldsymbol{\cdot}\mkern-1.2mu\hat{v}_\ss + \sigma\nabla_{\!\bar\vz}\!\boldsymbol{\cdot}\mkern-1.2muv_\ss\mkern-0.5mu\bigr)\dd s + v^\ssT_\ss \dd W_\ss + \hat{v}^\ssT_\ss \dd \bar{W}_\ss ,
    \end{aligned} \label{appeq:cvsde2c}
  \end{empheq}
  \end{subequations}
  \endgroup
  where $\varphi=\bar\eta+\gamma\log\mkern-1.2mu\rho=-\eta$, $\hat{\Phi}^\prime=e^{i\bar\varphi/c_\gamma}$, $\Phi^\prime=\rho\mkern.5mu e^{i\bar\eta/c_\gamma}=e^{i\chi/c_\gamma}$ and $\Phi^\prime\hat{\Phi}^\prime=\rho$.
  Here $(Y_\ss,\widehat{Y}_\ss)\coloneqq(\chi,-\eta)$, and the Wirtinger gradients are
  \[ 
    v_\ss = \sigma\nabla_{\!\vz}\mkern1mu\chi(\vz_\ss, \bar{\vz}_\ss, \text{\small {$T\shortminus s$}}), 
    \hat{v}_\ss = \sigma\nabla_{\!\bar\vz}\mkern1mu\chi(\vz_\ss, \bar{\vz}_\ss, \text{\small {$T\shortminus s$}}), \check{v}_\ss = \sigma\nabla_{\!\bar\vz}\mkern1mu\eta(\vz_\ss, \bar{\vz}_\ss, \text{\small {$T\shortminus s$}}), \bar{v}_\ss = \sigma\nabla_{\!\vz}\mkern1mu\eta(\vz_\ss, \bar{\vz}_\ss, \text{\small {$T\shortminus s$}}).
  \]
  When the forward and reverse potentials parameterize the same Hopf--Cole pair and the reverse process starts from $\rho_T$, the law of $Z_\ss$ in \eqref{appeq:cvsde2a} is $\rho_{T-s}$.  
\end{theorem}
\addtocounter{theorem}{-1}
\endgroup

\par\noindent\textit{Proof.}\enspace To apply the complex It\^o formula in \cref{lem:cpxito}, let us first apply the complex Hopf--Cole transform \cref{eq:hopfcole} \citep{hopf, cole, leger2021hopf} as observed in \cref{prop:csb-pde}, and let $\chi, \eta \in C^{2,1}(\bC^d,[0,T])$ be functions that represent the variables $Y=\chi$ and $\widehat{Y}=-\eta$ with the relation $\widehat{Y}_t = Y_t+i\log\rho_t$, using compatible logarithm branches. The last equality of each combined formula below uses the stated equality of the base-drift divergences. We first verify that the heat-type CSB-PDEs simplify the second-order term of the complex It\^o formula.

Next, apply the complex It\^o formula (\cref{lem:cpxito}) to the forward drift \eqref{appeq:cvsde1a}
\begin{equation}\label{eq:psifwdito}
\begin{aligned}
  \dd\chi = &\frac{\partial \chi}{\partial t} \dd t + \biggl[i(\nabla_{\!\vz}\mkern1.2mu\chi)^\sT\bigl(\,f + \sigma^2\nabla_{\!\bar\vz}\mkern1.2mu\psi\,\bigr) - i(\nabla_{\!\bar\vz}\mkern1.2mu\chi)^\sT\bigl(\,\overline{f +\sigma^2\nabla_{\!\bar{\vz}}\mkern1.2mu\psi}\,\bigr) + \mfrac{\sigma^2}{4} \Delta\chi \biggr] \dd t + \\&\qquad \sigma(\nabla_{\!\vz}\mkern1.2mu\chi)^\sT\mkern-1.2mu\dd W_t + \sigma(\nabla_{\!\bar\vz}\mkern1.2mu\chi)^\sT\mkern-1.2mu\dd\bar{W}_t,
\end{aligned}
\end{equation}
where the above expression uses simplified notation, such as $\psi$ for $\psi_t$. Note that, from PDEs~\eqref{subeq:csb-pde_fwd} and \eqref{subeq:csb-pde_bwd} and the Hopf--Cole relations above, which give $\chi+\eta=-i\log\rho$ and $\partial_t\log\rho=\Phi^{-1}\partial_t\Phi+\hat\Phi^{-1}\partial_t\hat\Phi$, the general expression for $\frac{\partial\chi}{\partial t}$ is derived as
\begin{align*}
  \frac{\partial\chi}{\partial t} &= -\frac{i c_\gamma}{\Phi} \biggl(-i\nabla_{\!\vz}\mkern1.2mu\Phi^\sT\!f +i\nabla_{\!\bar\vz}\mkern1.2mu\Phi^\sT\! \bar{f} - \frac{\sigma^2}{4}c_\gamma\Delta\Phi + \frac{i\gamma\sigma^2}{2c_\gamma}\frac{\Delta\sqrt{\rho}}{\sqrt{\rho}}\Phi + c_\gamma^{-1}F\Phi \biggr) - \gamma\frac{\partial}{\partial t}\log\mkern-1.2mu\rho \\
  &= -i(\nabla_{\!\vz}\mkern1.2mu\psi)^{\mkern-.5mu\sT}\mkern-.5mu f+i(\nabla_{\!\bar\vz}\mkern1.2mu\psi)^{\mkern-.5mu\sT}\mkern-.5mu \bar{f} + \sigma^2\frac{ic^2_\gamma}{\Phi}\nabla_{\!\bar{\vz}}\boldsymbol{\cdot} \biggl(\frac{i}{c_\gamma} \Phi \nabla_{\!\vz} \psi \biggr) + \frac{\gamma\sigma^2}{2}\frac{\Delta\sqrt{\rho}}{\sqrt{\rho}} - iF - \gamma\frac{\partial}{\partial t}\log\mkern-1.2mu\rho \\
  &= -i(\nabla_{\!\vz}\mkern1.2mu\chi)^{\mkern-.5mu\sT}\mkern-.5mu f+i(\nabla_{\!\bar\vz}\mkern1.2mu\chi)^{\mkern-.5mu\sT}\mkern-.5mu \bar{f} + i\gamma\bigl(\nabla_{\!\vz}\boldsymbol{\cdot}f - \nabla_{\!\bar\vz}\boldsymbol{\cdot}\bar{f}\bigr) \\
  &\qquad - i\sigma^2\nabla_{\!\bar\vz}\mkern1.2mu\chi^\sT \nabla_{\!\vz}\chi - \frac{\sigma^2}{4}\Delta(\chi+i\gamma\eta) - \frac{\gamma}{2}\lVert \sigma\nabla_{\!\bar\vz}\mkern1.2mu(\chi+\eta)\rVert^2 - iF. 
\end{align*}
Combining the time derivative with the whole It\^o formula and using the conjugation and linearity properties, we obtain
\begin{flalign*}
  \qquad \dd \chi &= -i\Bigl(-\sigma^2\nabla_{\!\vz}\mkern1.2mu\chi^\sT\nabla_{\!\bar\vz}\mkern1.2mu\psi + \sigma^2\nabla_{\bar{\vz}}\chi^\sT\overline{\nabla_{\!\bar\vz}\mkern1.2mu{\psi}} +\sigma^2\nabla_{\!\bar\vz}\mkern1.2mu\chi^\sT\nabla_{\!\vz}\mkern1.2mu\chi + F + \mfrac{\gamma\sigma^2}{4}\Delta\eta &&\\
  &\quad {}- \mfrac{i\gamma}{2}\lVert \sigma\nabla_{\!\bar\vz}\mkern1.2mu(\chi+\eta)\rVert^2\Bigr) \dd t + i\gamma\bigl(\nabla_{\!\vz}\boldsymbol{\cdot}f - \nabla_{\!\bar\vz}\boldsymbol{\cdot}\bar{f}\bigr) \dd t + \sigma(\nabla_{\!\vz}\chi)^\sT\mkern-1.2mu \dd W_t + \sigma (\nabla_{\!\bar\vz}\chi)^\sT\mkern-1.2mu \dd \bar{W}_t &&\\
  &= -i\Bigl(\lVert \sigma\nabla_{\!\vz}\mkern1.2mu\chi\rVert^2 + F + \mfrac{\gamma\sigma^2}{4}\Delta\eta - \mfrac{i\gamma}{2}\lVert \sigma\nabla_{\!\bar\vz}\mkern1.2mu(\chi+\eta)\rVert^2\Bigr) \dd t + &&\\
  &\qquad i\gamma\bigl(\nabla_{\!\vz}\boldsymbol{\cdot}f - \nabla_{\!\bar\vz}\boldsymbol{\cdot}\bar{f}\bigr) \dd t + \sigma(\nabla_{\!\vz}\chi)^\sT\mkern-1.2mu \dd W_t + \sigma (\nabla_{\!\bar\vz}\chi)^\sT\mkern-1.2mu \dd \bar{W}_t &&\\
  &= -i\bigl(\lVert v_t\rVert^2+F_t + \gamma\sigma\nabla_{\!\vz}\boldsymbol{\cdot} \check{v}_t - \mfrac{i\gamma}{2}\lVert \hat{v}_t+\check{v}_t\rVert^2\bigr) \dd t + v^\ssT_t \dd W_t +  \hat{v}^\ssT_t \dd \bar{W}_t. &&
\end{flalign*}
This implies that $\psi$ and $\chi$ are the associated mathematical objects in forward SDE integration. Applying the complex It\^o lemma to $\eta$, we obtain
\begin{equation}
  \begin{aligned}
  \dd\eta = &\frac{\partial \eta}{\partial t} \dd t + \biggl[i(\nabla_{\!\vz}\mkern1.2mu\eta)^\sT\bigl(\,f + \sigma^2\nabla_{\!\bar\vz}\mkern1.2mu\psi\,\bigr) - i(\nabla_{\!\bar\vz}\mkern1.2mu\eta)^\sT\bigl(\,\overline{f +\sigma^2\nabla_{\!\bar{\vz}}\mkern1.2mu\psi}\,\bigr) + \mfrac{\sigma^2}{4} \Delta\eta \biggr] \dd t + \\&\qquad \sigma(\nabla_{\!\vz}\mkern1.2mu\eta)^\sT\mkern-1.2mu\dd W_t + \sigma(\nabla_{\!\bar\vz}\mkern1.2mu\eta)^\sT\mkern-1.2mu\dd\bar{W}_t,
  \end{aligned}
\end{equation}
and for PDE~\eqref{subeq:csb-pde_bwd} the general expression for $\frac{\partial\eta}{\partial t}$ is derived as 
\[
  \begin{aligned}
    \frac{\partial\eta}{\partial t} &= -\frac{i c_\gamma}{\hat{\Phi}} \biggl(-i\nabla_{\!\vz} \boldsymbol{\cdot}\bigl(\,\hat{\Phi} f\,\bigr) + i\nabla_{\!\bar\vz}  \boldsymbol{\cdot} \bigl(\,\hat{\Phi}\bar{f}\,\bigr) + \frac{\sigma^2}{4}c_\gamma\Delta\hat{\Phi} - \frac{i\gamma\sigma^2}{2c_\gamma}\frac{\Delta\sqrt{\rho}}{\sqrt{\rho}}\hat{\Phi} - c_\gamma^{-1}F\hat{\Phi} \biggr) \\
    &= -i(\nabla_{\!\vz}\mkern1.2mu\eta)^{\mkern-.5mu\sT}\mkern-.5mu f - c_\gamma \nabla_{\!\vz} \boldsymbol{\cdot} f + i(\nabla_{\!\bar\vz}\mkern1.2mu\eta)^{\mkern-.5mu\sT}\mkern-.5mu\bar{f} + c_\gamma \nabla_{\!\bar\vz}\mkern1.2mu \boldsymbol{\cdot} \bar{f} \\
    &\qquad -\sigma^2\frac{ic_\gamma^2}{\hat{\Phi}} \nabla_{\!\vz}\boldsymbol{\cdot} \biggl(\frac{i}{c_\gamma} \hat{\Phi} \nabla_{\!\bar{\vz}}\eta \biggr) - \frac{\gamma\sigma^2}{2}\frac{\Delta\sqrt{\rho}}{\sqrt{\rho}} + iF\\
    &= -i(\nabla_{\!\vz}\mkern1.2mu\eta)^{\mkern-.5mu\sT}\mkern-.5mu f+i(\nabla_{\!\bar\vz}\mkern1.2mu\eta)^{\mkern-.5mu\sT}\mkern-.5mu \bar{f} - c_\gamma\bigl(\nabla_{\!\vz}\boldsymbol{\cdot}f - \nabla_{\!\bar\vz}\boldsymbol{\cdot}\bar{f}\bigr) \\
    &\qquad + i\sigma^2\nabla_{\!\bar\vz}\mkern1.2mu\eta^\sT \nabla_{\!\vz}\eta + \frac{\sigma^2}{4}\Delta(\eta-i\gamma\chi) - \frac{\gamma}{2}\lVert \sigma\nabla_{\!\bar\vz}\mkern1.2mu(\chi+\eta)\rVert^2 + iF.
  \end{aligned}
\]
Combining the above formulas yields
\begin{flalign*}
  \qquad -\dd \eta &= -i\Bigl(\sigma^2\nabla_{\!\vz}\mkern1.2mu\eta^\sT\nabla_{\!\bar\vz}\mkern1.2mu\psi - \sigma^2\nabla_{\bar{\vz}}\eta^\sT\overline{\nabla_{\!\bar\vz}\mkern1.2mu{\psi}} +\sigma^2\nabla_{\!\bar\vz}\mkern1.2mu\eta^\sT\nabla_{\!\vz}\mkern1.2mu\eta &&\\[-2pt]
  &\qquad\quad+F - \mfrac{\gamma\sigma^2}{4}\Delta\chi + \mfrac{i\gamma}{2}\lVert \sigma\nabla_{\!\bar\vz}\mkern1.2mu(\chi+\eta)\rVert^2\Bigr) \dd t + &&\\[-2pt]
  &\qquad \bigl(c_\gamma(\nabla_{\!\vz}\boldsymbol{\cdot}f - \nabla_{\!\bar\vz}\boldsymbol{\cdot}\bar{f}) - \mfrac{\sigma^2}{2}\Delta\eta\bigr) \dd t - \sigma(\nabla_{\!\vz}\eta)^\sT\mkern-1.2mu \dd W_t - \sigma (\nabla_{\!\bar\vz}\eta)^\sT\mkern-1.2mu \dd \bar{W}_t &&\\
  &= -i\Bigl(\lVert\sigma\nabla_{\!\vz}\mkern1.2mu\chi\rVert^2 + F - \mfrac{\gamma\sigma^2}{4}\Delta\chi - \bigl(1+\mfrac{i\gamma}{2}\bigr)\lVert \sigma\nabla_{\!\bar\vz}\mkern1.2mu(\chi+\eta)\rVert^2\Bigr) \dd t + &&\\
  &\qquad \bigl(c_\gamma(\nabla_{\!\vz}\boldsymbol{\cdot}f - \nabla_{\!\bar\vz}\boldsymbol{\cdot}\bar{f}) - \mfrac{\sigma^2}{2}\Delta\eta\bigr) \dd t - \sigma(\nabla_{\!\vz}\mkern1.2mu\eta)^\sT\mkern-1.2mu \dd W_t - \sigma(\nabla_{\!\bar\vz}\mkern1.2mu\eta)^\sT\mkern-1.2mu \dd\bar{W}_t &&\\
  &= -i\bigl(\lVert v_t\rVert^2+F_t - \gamma\sigma\nabla_{\!\vz}\boldsymbol{\cdot} \hat{v}_t - \bigl(1+\mfrac{i\gamma}{2}\bigr)\lVert \hat{v}_t+\check{v}_t\rVert^2\bigr) \dd t - &&\\
  &\qquad \sigma\bigl(\nabla_{\!\vz}\boldsymbol{\cdot}\check{v}_t + \nabla_{\!\bar\vz}\boldsymbol{\cdot}\bar{v}_t\bigr) \dd t - \bar{v}^\ssT_t \dd W_t - \check{v}^\ssT_t \dd \bar{W}_t. &&
\end{flalign*}

At corresponding times $t=T-s$, the matched Hopf--Cole factors satisfy $\Phi^\prime=\Phi$ and $\hat\Phi^\prime=\hat\Phi$. Comparing their exponents in \eqref{eq:hopfcole} and \eqref{eq:hopfcole-rev} on compatible logarithm branches gives
\[
  (\chi,\eta)=
  \begin{cases}
    (-\bar\psi,\bar\varphi) & \text{in the forward parameterization},\\
    (\psi,-\varphi) & \text{in the backward parameterization}.
  \end{cases}
\]
Conjugating the forward identity $\chi+\eta=-i\log\rho$ gives $\varphi=\psi+i\log\rho$. Taking its $\nabla_{\!\bar\vz}$ gradient gives the drift of $Z_\ss$ in \eqref{appeq:cvsde2a} directly from \eqref{appeq:cvsde1a},
\[
  i\bigl(-f-\sigma^2\nabla_{\!\bar\vz}\varphi\bigr)
  =-i\bigl(f+\sigma^2\nabla_{\!\bar\vz}\psi\bigr)+\sigma^2\nabla_{\!\bar\vz}\log\rho.
\]
The reverse process initialized with the forward terminal law $\rho_T$ therefore has marginal law $\rho_{T-s}$.

For independently fitted potentials, let $\mathrm f,\mathrm b$ denote the forward and backward parameterizations and $\rho$ the marginal density of the fixed sampling law. Under $\bP_\theta$, $\eta^\phi_{\mathrm f}-\eta^\phi_{\mathrm b}=\gamma\log\rho=-2\Re\chi^\theta_{\mathrm f}$, while under $\bQ_\phi$, $\chi^\theta_{\mathrm b}-\chi^\theta_{\mathrm f}=\gamma\log\rho=-2\Re\eta^\phi_{\mathrm b}$. Taking Wirtinger gradients gives
\[
  \overline{v^\theta_{\mathrm f}}-\check{v}^\phi_{\mathrm b}=
  \begin{cases}
    -\hat{v}^\theta_{\mathrm f}-\check{v}^\phi_{\mathrm f} & \text{under }\bP_\theta,\\
    \overline{v^\theta_{\mathrm b}+\bar{v}^\phi_{\mathrm b}} & \text{under }\bQ_\phi.
  \end{cases}
\]
Multiplying by $-i\sigma$ gives the drift sum in $\vz$-coordinates. Each shift adds $\tfrac{\gamma\sigma}{4}\Delta\log\rho\in\bR$ to the corresponding divergence, preserving its imaginary part.

Below we suppress the backward subscripts on $\chi$ and $\eta$, and $\partial_s$ differentiates the time-reindexed potentials. Applying the complex It\^o lemma to $\eta$ using \eqref{appeq:cvsde2a}, we obtain
\begin{equation}
\begin{aligned}
  \dd\eta = \frac{\partial\eta}{\partial s}\dd s &+ \biggl[i(\nabla_{\!\bar\vz}\mkern1.2mu\eta)^\sT\bigl(\,\bar{f}+\sigma^2\nabla_{\!\vz}\mkern1.2mu\bar\varphi\,\bigr) - i(\nabla_{\!\vz}\mkern1.2mu\eta)^\sT\bigl(\,f + \sigma^2\nabla_{\!\bar\vz}\mkern1.2mu\varphi\,\bigr) + \frac{\sigma^2}{4} \Delta\eta \biggr] \dd s \\&+ \sigma(\nabla_{\!\bar\vz}\mkern1.2mu\eta)^\sT\mkern-1.2mu\dd \bar{W}_\ss + \sigma (\nabla_{\!\vz}\eta)^\sT\mkern-1.2mu\dd W_\ss.
\end{aligned}
\end{equation}
From PDEs~\eqref{subeq:csb-pde-rev_fwd} and \eqref{subeq:csb-pde-rev_bwd} in \cref{cor:csb-pde-rev} and the Hopf--Cole relations above, which give $\chi+\eta=-i\log\rho$ and $\partial_s\log\rho=(\Phi^\prime)^{-1}\partial_s\Phi^\prime+(\hat\Phi^\prime)^{-1}\partial_s\hat\Phi^\prime$, the general expression for $\frac{\partial\eta}{\partial s}$ is  
\begin{align*}
  \frac{\partial\eta}{\partial s} &= -\frac{i c_\gamma}{\hat{\Phi}^\prime} \biggl(-i\nabla_{\!\bar\vz}\mkern1.2mu\hat{\Phi}^{\prime\sT}\!\bar{f} +i\nabla_{\!\vz}\mkern1.2mu\hat{\Phi}^{\prime\sT}\!f - \frac{\sigma^2}{4}c_\gamma\Delta\hat{\Phi}^\prime + \frac{i\gamma\sigma^2}{2c_\gamma}\frac{\Delta\sqrt{\rho}}{\sqrt{\rho}}\hat{\Phi}^\prime + c_\gamma^{-1}F\hat{\Phi}^\prime \biggr) - \gamma\frac{\partial}{\partial s}\log\mkern-1.2mu\rho \\
  &= i(\nabla_{\!\vz}\mkern1.2mu\bar\varphi)^{\mkern-.5mu\sT}\mkern-.5mu f-i(\nabla_{\!\bar\vz}\mkern1.2mu\bar\varphi)^{\mkern-.5mu\sT}\mkern-.5mu \bar{f} + \sigma^2\frac{ic^2_\gamma}{\hat{\Phi}^\prime}\nabla_{\!\bar{\vz}}\boldsymbol{\cdot} \biggl(\frac{i}{c_\gamma} \hat{\Phi}^\prime \nabla_{\!\vz} \bar\varphi \biggr) + \frac{\gamma\sigma^2}{2}\frac{\Delta\sqrt{\rho}}{\sqrt{\rho}} - iF - \gamma\frac{\partial}{\partial s}\log\mkern-1.2mu\rho \\
  &= i(\nabla_{\!\vz}\mkern1.2mu\eta)^{\mkern-.5mu\sT}\mkern-.5mu f-i(\nabla_{\!\bar\vz}\mkern1.2mu\eta)^{\mkern-.5mu\sT}\mkern-.5mu \bar{f} - i\gamma\bigl(\nabla_{\!\vz}\boldsymbol{\cdot}f - \nabla_{\!\bar\vz}\boldsymbol{\cdot}\bar{f}\bigr) \\
  &\qquad - i\sigma^2\nabla_{\!\bar\vz}\mkern1.2mu\eta^\sT \nabla_{\!\vz}\eta - \frac{\sigma^2}{4}\Delta(\eta+i\gamma\chi) - \frac{\gamma}{2}\lVert \sigma\nabla_{\!\bar\vz}\mkern1.2mu(\chi+\eta)\rVert^2 - iF. 
\end{align*}
By repeating a similar process, we obtain
\begin{flalign*}
  \qquad \dd \eta &= -i\Bigl(-\sigma^2\nabla_{\!\bar\vz}\mkern1.2mu\eta^\sT\nabla_{\!\vz}\mkern1.2mu\bar\varphi + \sigma^2\nabla_{\!\vz}\mkern1.2mu\eta^\sT\nabla_{\!\bar\vz}\mkern1.2mu\varphi +\sigma^2\nabla_{\!\bar\vz}\mkern1.2mu\eta^\sT\nabla_{\!\vz}\mkern1.2mu\eta + F + \mfrac{\gamma\sigma^2}{4}\Delta\chi &&\\
  &\quad {}- \mfrac{i\gamma}{2}\lVert \sigma\nabla_{\!\bar\vz}\mkern1.2mu(\chi+\eta)\rVert^2\Bigr) \dd s {}-i\gamma\bigl(\nabla_{\!\vz}\boldsymbol{\cdot}f - \nabla_{\!\bar\vz}\boldsymbol{\cdot}\bar{f}\bigr) \dd s + \sigma(\nabla_{\!\bar\vz}\eta)^\sT\mkern-1.2mu \dd \bar{W}_\ss + \sigma (\nabla_{\!\vz}\eta)^\sT\mkern-1.2mu \dd W_\ss &&\\
  &= -i\Bigl(\lVert \sigma\nabla_{\!\bar\vz}\mkern1.2mu\eta\rVert^2 + F + \mfrac{\gamma\sigma^2}{4}\Delta\chi - \mfrac{i\gamma}{2}\lVert \sigma\nabla_{\!\vz}\mkern1.2mu(\chi+\eta)\rVert^2\Bigr) \dd s &&\\
  &\qquad {}-i\gamma\bigl(\nabla_{\!\vz}\boldsymbol{\cdot}f - \nabla_{\!\bar\vz}\boldsymbol{\cdot}\bar{f}\bigr) \dd s + \sigma(\nabla_{\!\bar\vz}\eta)^\sT\mkern-1.2mu \dd \bar{W}_\ss + \sigma (\nabla_{\!\vz}\eta)^\sT\mkern-1.2mu \dd W_\ss &&\\
  &= -i\bigl(\lVert \check{v}_\ss\rVert^2 + F_\ss + \gamma\sigma\nabla_{\!\bar\vz}\boldsymbol{\cdot}v_\ss - \mfrac{i\gamma}{2}\lVert v_\ss+\bar{v}_\ss\rVert^2\bigr) \dd s + \check{v}^\ssT_\ss \dd \bar{W}_\ss + \bar{v}^\ssT_\ss \dd W_\ss. &&
\end{flalign*}
Since $\widehat{Y}_\ss=-\eta$, we have $\dd\widehat{Y}_\ss=-\dd\eta$.

Lastly, applying the complex It\^o lemma to $\chi$ using \eqref{appeq:cvsde2a}, we obtain
\begin{equation}
\begin{aligned}
  \dd\chi = \frac{\partial\chi}{\partial s}\dd s &+ \biggl[i(\nabla_{\!\bar\vz}\mkern1.2mu\chi)^\sT\bigl(\,\bar{f}+\sigma^2\nabla_{\!\vz}\mkern1.2mu\bar\varphi\,\bigr) - i(\nabla_{\!\vz}\mkern1.2mu\chi)^\sT\bigl(\,f + \sigma^2\nabla_{\!\bar\vz}\mkern1.2mu\varphi\,\bigr) + \frac{\sigma^2}{4} \Delta\chi \biggr] \dd s \\&+ \sigma(\nabla_{\!\bar\vz}\mkern1.2mu\chi)^\sT\mkern-1.2mu\dd \bar{W}_\ss + \sigma (\nabla_{\!\vz}\chi)^\sT\mkern-1.2mu\dd W_\ss.
\end{aligned}
\end{equation}
From PDE~\eqref{subeq:csb-pde-rev_bwd}, the general expression for $\frac{\partial \chi}{\partial s}$ is derived as 
\[
  \begin{aligned}
    \frac{\partial\chi}{\partial s} &= -\frac{i c_\gamma}{\Phi^\prime} \biggl(-i\nabla_{\!\bar\vz} \boldsymbol{\cdot}\bigl(\,\Phi^\prime \bar{f}\,\bigr) + i\nabla_{\!\vz} \boldsymbol{\cdot} \bigl(\,\Phi^\prime f\,\bigr) + \frac{\sigma^2}{4}c_\gamma\Delta\Phi^\prime - \frac{i\gamma\sigma^2}{2c_\gamma}\frac{\Delta\sqrt{\rho}}{\sqrt{\rho}}\Phi^\prime - c_\gamma^{-1}F\Phi^\prime \biggr) \\
    &= i(\nabla_{\!\vz}\mkern1.2mu\chi)^{\mkern-.5mu\sT}\mkern-.5mu f + c_\gamma \nabla_{\!\vz} \boldsymbol{\cdot} f - i(\nabla_{\!\bar\vz}\mkern1.2mu\chi)^{\mkern-.5mu\sT}\mkern-.5mu\bar{f} - c_\gamma \nabla_{\!\bar\vz}\mkern1.2mu \boldsymbol{\cdot} \bar{f} \\
    &\qquad -\sigma^2\frac{ic_\gamma^2}{\Phi^\prime} \nabla_{\!\vz}\boldsymbol{\cdot} \biggl(\frac{i}{c_\gamma} \Phi^\prime \nabla_{\!\bar{\vz}}\chi \biggr) - \frac{\gamma\sigma^2}{2}\frac{\Delta\sqrt{\rho}}{\sqrt{\rho}} + iF\\
    &= i(\nabla_{\!\vz}\mkern1.2mu \chi)^{\mkern-.5mu\sT}\mkern-.5mu f-i(\nabla_{\!\bar\vz}\mkern1.2mu\chi)^{\mkern-.5mu\sT}\mkern-.5mu\bar{f} + c_\gamma\bigl(\nabla_{\!\vz}\boldsymbol{\cdot}f - \nabla_{\!\bar\vz}\boldsymbol{\cdot}\bar{f}\bigr) \\
    &\qquad + i\sigma^2\nabla_{\!\bar\vz}\mkern1.2mu\chi^\sT \nabla_{\!\vz}\mkern1.2mu\chi + \frac{\sigma^2}{4}\Delta(\chi-i\gamma\eta) - \frac{\gamma}{2}\lVert \sigma\nabla_{\!\vz}\mkern1.2mu(\chi+\eta)\rVert^2 + iF.
  \end{aligned}
\]
With the same procedure as for \cref{appeq:cvsde1c}, combining the above formulas yields
\begin{flalign*}
  \qquad \dd \chi &= -i\Bigl(-\sigma^2\nabla_{\!\bar\vz}\mkern1.2mu\chi^\sT\nabla_{\!\vz}\mkern1.2mu\bar\varphi + \sigma^2\nabla_{\!\vz}\mkern1.2mu\chi^\sT\nabla_{\!\bar\vz}\mkern1.2mu\varphi -\sigma^2\nabla_{\!\bar\vz}\mkern1.2mu\chi^\sT\nabla_{\!\vz}\mkern1.2mu\chi &&\\
  &\qquad {}- F + \mfrac{\gamma\sigma^2}{4}\Delta\eta - \mfrac{i\gamma}{2}\lVert \sigma\nabla_{\!\vz}\mkern1.2mu(\chi+\eta)\rVert^2\Bigr) \dd s + &&\\
  &\qquad \bigl(c_\gamma(\nabla_{\!\vz}\boldsymbol{\cdot}f - \nabla_{\!\bar\vz}\boldsymbol{\cdot}\bar{f}) + \mfrac{\sigma^2}{2}\Delta\chi\bigr) \dd s + \sigma(\nabla_{\!\bar\vz}\chi)^\sT\mkern-1.2mu \dd \bar{W}_\ss + \sigma (\nabla_{\!\vz}\chi)^\sT\mkern-1.2mu \dd W_\ss &&\\
  &= -i\Bigl(-\lVert \sigma\nabla_{\!\bar\vz}\mkern1.2mu\eta\rVert^2 - F + \mfrac{\gamma\sigma^2}{4}\Delta\eta + \bigl(1+\mfrac{i\gamma}{2}\bigr)\lVert \sigma\nabla_{\!\vz}\mkern1.2mu(\chi+\eta)\rVert^2\Bigr) \dd s + &&\\
  &\qquad \bigl(c_\gamma(\nabla_{\!\vz}\boldsymbol{\cdot}f - \nabla_{\!\bar\vz}\boldsymbol{\cdot}\bar{f}) + \mfrac{\sigma^2}{2}\Delta\chi\bigr) \dd s + \sigma(\nabla_{\!\bar\vz}\mkern1.2mu\chi)^\sT\mkern-1.2mu \dd \bar{W}_\ss + \sigma(\nabla_{\!\vz}\mkern1.2mu\chi)^\sT\mkern-1.2mu \dd W_\ss &&\\
  &= -i\bigl(-\lVert \check{v}_\ss\rVert^2-F_\ss + \gamma\sigma\nabla_{\!\vz}\boldsymbol{\cdot}\check{v}_\ss + \bigl(1+\mfrac{i\gamma}{2}\bigr)\lVert v_\ss+\bar{v}_\ss\rVert^2\bigr) \dd s + &&\\
  &\qquad \sigma\bigl(\nabla_{\!\vz}\boldsymbol{\cdot}\hat{v}_\ss + \nabla_{\!\bar\vz}\boldsymbol{\cdot}v_\ss\bigr) \dd s + \hat{v}^\ssT_\ss \dd \bar{W}_\ss + v^\ssT_\ss \dd W_\ss, &&
\end{flalign*}
which proves \cref{appeq:cvsde2c}. \QED{}

\subsection{Proof of Theorem~\texorpdfstring{\ref{thm:ipf}}{2} (iterative proportional fitting)}\label{subsec:ipf_new}

We assume the Hopf--Cole reality condition for each law's own marginal density,
with $\bP_{\theta,0}=\rho_0$ and $\bQ_{\phi,T}=\rho_T$, and the regularity and integrability required for Girsanov's theorem, time reversal, and the log-density It\^o formulas. We establish both KL representations. Fix $\theta$ and take expectations under $\bP_\theta$, with $t$-indexed fields in the forward parameterization. Using \cref{thm:sb-cvsde} with $F=0$, the parameterization identities in its proof, the change of measure~\eqref{eq:com}, and the time-reversal calculation of \citet[Proposition 9]{deepgsb}, we obtain
\begingroup
\addtolength{\abovedisplayskip}{-1pt}
\addtolength{\belowdisplayskip}{-1pt}
\begin{align*}
  \KL(\bP_\theta\Vert\,\bQ_\phi)
  &=\bE_{\bP_\theta}\bigl[\log\rho_0(Z_0)-\log\rho_T(Z_T)\bigr]\\[-1pt]
  &\quad+\int_0^T\mkern-3mu\bE_{\bP_\theta}\Bigl[
    \bigl\lVert\hat{v}^\theta_t+\check{v}^\phi_t\bigr\rVert^2
    -2\sigma\Im\bigl\{\nabla_{\!\vz}\boldsymbol{\cdot}\mkern-1.2mu\check{v}^\phi_t\bigr\}
    \Bigr]\!\dd t.
\end{align*}
\endgroup
Indeed, the real diffusion covariance is $(\sigma^2/2)\vI_{2d}$. Applying It\^o's formula to the log marginal density of $\bQ_\phi$ cancels the score terms in the squared drift difference. Its boundary contribution cancels the $\phi$-dependent initial log-density in Girsanov's initial density ratio, leaving logarithms of the fixed boundary densities $\rho_0$ and $\rho_T$. Under the assumed integrability, the stochastic integrals have zero expectation, and the base-drift condition gives $\Im\nabla_{\!\vz}\boldsymbol{\cdot}f=0$.

Since $\nabla_{\!\vz}\boldsymbol{\cdot}\check{v}^\phi_t=\nabla_{\!\bar\vz}\boldsymbol{\cdot}\bar{v}^\phi_t$, taking the imaginary part of $\cE^{\scriptscriptstyle-}_T$ gives
\begingroup
\addtolength{\abovedisplayskip}{-1pt}
\addtolength{\belowdisplayskip}{-1pt}
\begin{align*}
  \Im\bigl[\cE^{\scriptscriptstyle-}_T(\theta;\phi)\bigr]
  &=\int_0^T\mkern-3mu\bE_{\bP_\theta}\Bigl[
    \bigl\lVert\hat{v}^\theta_t+\check{v}^\phi_t\bigr\rVert^2
    -2\sigma\Im\bigl\{\nabla_{\!\vz}\boldsymbol{\cdot}\mkern-1.2mu\check{v}^\phi_t\bigr\}
    \Bigr]\!\dd t
  -\int_0^T\mkern-3mu\bE_{\bP_\theta}\Bigl[
    \lVert v^\theta_t\rVert^2
    -\gamma\sigma\Re\bigl\{\nabla_{\!\vz}\boldsymbol{\cdot}\mkern-1.2mu\hat{v}^\theta_t\bigr\}
    \Bigr]\!\dd t.
\end{align*}
\endgroup
The endpoint expectation and the integral subtracted in the preceding display are independent of $\phi$. Denoting their sum by $C_\theta$, we have $\KL(\bP_\theta\Vert\,\bQ_\phi)=\Im[\cE^{\scriptscriptstyle-}_T(\theta;\phi)]+C_\theta$. Thus this projection updates $\phi$ while keeping $\theta$ and the sampling law $\bP_\theta$ fixed.

For the reverse IPF, take expectations under $\bQ_\phi$, with $s$-indexed fields in the backward parameterization at $t=T-s$ and the path law expressed in the same $\vz$-coordinates. The same argument gives
\begingroup
\addtolength{\abovedisplayskip}{-1pt}
\addtolength{\belowdisplayskip}{-1pt}
\begin{align*}
  \KL(\bQ_\phi\Vert\,\bP_\theta)
  &=\bE_{\bQ_\phi}\bigl[\log\rho_T(Z_T)-\log\rho_0(Z_0)\bigr]\\[-1pt]
  &\quad+\int_0^T\mkern-3mu\bE_{\bQ_\phi}\Bigl[
    \bigl\lVert v^\theta_\ss+\bar{v}^\phi_\ss\bigr\rVert^2
    -2\sigma\Im\bigl\{\nabla_{\!\bar\vz}\boldsymbol{\cdot}\mkern-1.2mu v^\theta_\ss\bigr\}
    \Bigr]\!\dd s.
\end{align*}
\endgroup
Using $\nabla_{\!\vz}\boldsymbol{\cdot}\hat{v}^\theta_\ss=\nabla_{\!\bar\vz}\boldsymbol{\cdot}v^\theta_\ss$ in the definition of $\cE^{\scriptscriptstyle+}_0$ gives
\begingroup
\addtolength{\abovedisplayskip}{-1pt}
\addtolength{\belowdisplayskip}{-1pt}
\begin{align*}
  -\Im\bigl[\cE^{\scriptscriptstyle+}_0(\phi;\theta)\bigr]
  &=\int_0^T\mkern-3mu\bE_{\bQ_\phi}\Bigl[
    \bigl\lVert v^\theta_\ss+\bar{v}^\phi_\ss\bigr\rVert^2
    -2\sigma\Im\bigl\{\nabla_{\!\bar\vz}\boldsymbol{\cdot}\mkern-1.2mu v^\theta_\ss\bigr\}
    \Bigr]\!\dd s\\
  &\quad-\int_0^T\mkern-3mu\bE_{\bQ_\phi}\Bigl[
    \lVert\check{v}^\phi_\ss\rVert^2
    -\gamma\sigma\Re\bigl\{\nabla_{\!\vz}\boldsymbol{\cdot}\mkern-1.2mu\check{v}^\phi_\ss\bigr\}
    \Bigr]\!\dd s.
\end{align*}
\endgroup
The endpoint expectation and the integral subtracted in the preceding display are independent of $\theta$. Denoting their sum by $C_\phi$, we have $\KL(\bQ_\phi\Vert\,\bP_\theta)=-\Im[\cE^{\scriptscriptstyle+}_0(\phi;\theta)]+C_\phi$. Thus this projection updates $\theta$ while keeping $\phi$ and the sampling law $\bQ_\phi$ fixed.

To match the active-parameter labels of the training losses, let $\theta'$ and $\phi'$ denote frozen parameter copies and define
\begin{alignat*}{2}
  \mathcal{L}_{\mathrm{IPF}}^{+}(\phi;\theta')
  &:= \Im[\cE^{\scriptscriptstyle-}_T(\theta';\phi)],
  &\qquad& \text{with expectations under } \bP_{\theta'},\\
  \mathcal{L}_{\mathrm{IPF}}^{-}(\theta;\phi')
  &:= -\Im[\cE^{\scriptscriptstyle+}_0(\phi';\theta)],
  &\qquad& \text{with expectations under } \bQ_{\phi'}.
\end{alignat*}
Here the signs on $\cE^{\scriptscriptstyle\pm}$ specify the integration coordinates ($s$ for $+$ and $t$ for $-$), whereas the signs on $\mathcal{L}_{\mathrm{IPF}}^{\pm}$ follow the active-parameter labels of the training losses. Furthermore, suppose that the change to the backward parameterization and the reality condition for $\bQ_{\phi'}$ give
\[
  \varphi_{\phi'}=\widehat Y_\ss^{\phi'},\qquad
  \psi_\theta=\gamma\log\rho-\overline{Y_\ss^\theta}
             =2\Re\widehat Y_\ss^{\phi'}-\overline{Y_\ss^\theta},
\]
and also $Y_\ss^\theta-\widehat Y_\ss^{\phi'}=\overline{\varphi_{\phi'}-\psi_\theta}$ and $\Im\Delta Y_\ss^\theta=\Im\Delta\psi_\theta$. Substituting these identities into the reverse KL integrand gives the Hyv{\"a}rinen score-matching objective \citep{hyvs}
\begin{equation}
\arg\min_\theta \cL_{\mathrm{IPF}}^{-}(\theta;\phi')
=
\arg\min_\theta \int_0^T
\bE_{\bQ_{\phi'}}\Bigl[
  \bigl\lVert
    \sigma \nabla_{\!\bar\vz}\psi_\theta
    - \sigma \nabla_{\!\bar\vz}\varphi_{\phi'}
  \bigr\rVert^2
  - \mfrac{1}{2}\sigma^2 \Im\Delta\psi_\theta
\Bigr]\,\dd s,
\end{equation}
considering $\varphi_{\phi^\prime}$ as a fixed target. \QED{}

\subsection{Proof of Proposition~\texorpdfstring{\ref{prop:posterior}}{2} (analytic posterior)}

For the variance-time generalization of $\textrm{COT}_\gamma$ in \cref{eq:cot_scheduled}, consider the endpoint-conditioned CVSDE
\begin{equation}\label{eq:bridge}
  \dd Z_t = -c_\gamma\frac{\sigma_t^2}{\int_t^T \sigma_\tau^2 \dd \tau}(Z_t - \vz_T)\dd t + \sigma_t\dd W_t, \qquad Z_0 = \vz_0, \quad 0 \le t < T.
\end{equation}
The posterior formula holds for $0 < s < T$, while the endpoint laws degenerate to point masses at $\bar{\vz}_T$ for $s = 0$ and at $\bar{\vz}_0$ for $s = T$.

For the complex exponential function $h_\gamma(\ell) = e^{c_\gamma\ell}$, the parametrization $\ell = \log(1 - t/T)$ yields a logarithmic spiral traversed at constant Euclidean speed. Consider first the $\textrm{COT}_\gamma$ subproblem with constant $\sigma_t$ for absolutely continuous functions:
\begin{equation}\label{eq:cot_constant}
  \inf_{\substack{u(0) = \vz_0\\u(T) = \vz_T}} \int_0^T\!L_\gamma\bigl(u(t), \dot{u}(t), t\bigr)\dd t, \qquad u \in AC([0, T], \bC^d).
\end{equation}
For paths of finite action, the transformed path $(1 - t/T)^{-i\gamma}(u(t) - \vz_T)$ has a square-integrable derivative and extends absolutely continuously to $t = T$.
We apply the Cauchy--Schwarz inequality to the objective against the constant function of value $1$.
Combining this with the fundamental theorem of calculus and the endpoint conditions \citep{dunajski2021variational}
\[
  (1 - t/T)^{-i\gamma}(u(t) - \vz_T)\big\vert_{t = 0} = \vz_0 - \vz_T, \qquad \lim_{t \uparrow T}(1 - t/T)^{-i\gamma}(u(t) - \vz_T) = \vzero,
\]
we obtain the lower bound
\begin{align*}
  \int_0^T\!L_\gamma\bigl(u(t), \dot{u}(t), t\bigr)\dd t &= \int_0^T\biggl\lVert\frac{\dd}{\dd t} \Bigl[(1 - t/T)^{-i\gamma}(u(t) - \vz_T)\Bigr]\biggr\rVert^2\dd t\\
  &\ge \frac{1}{T}\biggl\lVert\int_0^T\!\frac{\dd}{\dd t} \Bigl[(1 - t/T)^{-i\gamma}(u(t) - \vz_T)\Bigr]\dd t\biggr\rVert^2 = \frac{\lVert\vz_0 - \vz_T\rVert^2}{T}.
\end{align*}
For an optimal forward path $\vec{u} \in AC([0, T], \bC^d)$, the optimality condition
\begin{equation}\label{eq:optim_u}
  \frac{\dd}{\dd t}\Bigl[(1 - t/T)^{-i\gamma}(\vec{u}(t) - \vz_T)\Bigr] = \frac{\vz_T - \vz_0}{T}
\end{equation}
holds almost everywhere. Recall that for $a \in \bR^+$ and $b \in \bC$, complex conjugation satisfies
\[
  \overline{a^b} = \overline{e^{b\log a}} = e^{\bar{b}\log a} = a^{\bar{b}}.
\]
Combining the endpoint conditions with the optimality condition in \cref{eq:optim_u}, we obtain the minimizing forward path and its conjugate time reversal in the form
\begin{subequations}\label{eq:optim_paths}
\begin{empheq}[left={\empheqlbrace}]{align}
  \vec{u}(t) &= \vz_T + (\vz_0 - \vz_T)(1 - t/T)^{c_\gamma}, &&0 \le t < T, \label{eq:optim_u_fwd}\\
  \cev{u}(s) &= \bar{\vz}_T + (\bar{\vz}_0 - \bar{\vz}_T)(s/T)^{\bar{c}_\gamma}, &&0 < s \le T.\label{eq:optim_u_rev}
\end{empheq}
\end{subequations}
These paths extend continuously to $\vec{u}(T) = \vz_T$ and $\cev{u}(0) = \bar{\vz}_T$, with $\cev{u}(T) = \bar{\vz}_0$ and $\vec{u}(t) = \overline{\cev{u}(T - t)}$. In particular, $\bigl\lVert\dot{\vec{u}}(t)\bigr\rVert = \sqrt{1 + \gamma^2}\lVert\vz_0 - \vz_T\rVert/T$, so the forward path is absolutely continuous and has constant speed.

Suppose that $\sigma_t$ is scheduled in a time-dependent manner. We use reverse time for the diffusion coefficient $\sigma$ and define the accumulated and remaining variances as
\[
  \hat{\omega}^2_\ss = \int^s_0 \hat{\sigma}^2_\tau \dd \tau, \qquad \omega^2_\ss \coloneqq \int^T_\ss \hat{\sigma}^2_\tau \dd \tau, \qquad 0 \le s \le T.
\]
Using these accumulated variances, we restate the objective following the diffusion scheduling argument of \citet{i2sb}.

\begin{definition}[Generalization of \cref{eq:cot}]
We set
\begin{equation}\label{eq:variance_time}
  r_\ss \coloneqq \frac{\hat{\omega}_\ss^2}{\hat{\omega}_\ss^2 + \omega_\ss^2} = \frac{\hat{\omega}_\ss^2}{\hat{\omega}_T^2}, \qquad k_\ss = r_\ss^{\bar{c}_\gamma}, \qquad 0 \le s \le T,
\end{equation}
where $k_0 = 0$ is defined by continuity and $\bQ_W$ denotes the reference law with zero drift and diffusion coefficient $\sigma_t$. Replacing $1 - t/T$ by $r_{T - t}$, we generalize the KL objective and redefine the rotation and its induced drift as
\begingroup
\addtolength{\belowdisplayskip}{-1pt}
\begin{equation}\label{eq:cot_scheduled}
\begin{gathered}
  \minimize_{\bP^\gamma_{0,T} = \rho_{0,T}} \mathrm{KL}(\bP^\gamma \Vert \bQ^\gamma_W),
  \enspace\text{s.t.}\enspace\, \bQ^\gamma_W = (\cH_\gamma)_\#({\bQ}_W),
  \\[-3pt]
  (\cH_\gamma \vz)_t \coloneqq \vz_T + \frac{h_\gamma(\log r_{T - t})}{h_0(\log r_{T - t})}(\vz_t - \vz_T),
  \\[-3pt]
  b_\gamma(\vz, t; \vz_T) \coloneqq i\gamma\frac{\dd}{\dd t}\log r_{T - t}\, (\vz - \vz_T) = -\frac{i\gamma\sigma_t^2}{\hat{\omega}_{T - t}^2}(\vz - \vz_T).
\end{gathered}
\end{equation}
\endgroup
These formulas hold for $0 \le t < T$, with $(\cH_\gamma u)_T = \vz_T$ defined by continuity.
\end{definition}
Applying the Cauchy--Schwarz inequality with weight $\sigma_t^2$ to the variance-time action yields
\begin{align*}
  \int_0^T\frac{\bigl\lVert \dot{u}(t) - b_\gamma(u(t), t; \vz_T) \bigr\rVert^2}{\sigma_t^2}\dd t &= \int_0^T\frac{1}{\sigma_t^2}\biggl\lVert\frac{\dd}{\dd t} \Bigl[r_{T - t}^{-i\gamma}(u(t) - \vz_T)\Bigr]\biggr\rVert^2\dd t\\
  &\ge \frac{1}{\hat{\omega}_T^2}\biggl\lVert\int_0^T\!\frac{\dd}{\dd t} \Bigl[r_{T - t}^{-i\gamma}(u(t) - \vz_T)\Bigr]\dd t\biggr\rVert^2 = \frac{\lVert\vz_0 - \vz_T\rVert^2}{\hat{\omega}_T^2}.
\end{align*}
Equality holds when the transformed derivative equals $(\vz_T - \vz_0)\sigma_t^2/\hat{\omega}_T^2$. Given the endpoint conditions, the minimizing forward path is $\vz_T + (\vz_0 - \vz_T)r_{T - t}^{c_\gamma}$, and its conjugate time reversal takes the form
\begin{equation}\label{eq:scheduled_path}
  \cev{u}(s) = \bar{\vz}_T + (\bar{\vz}_0 - \bar{\vz}_T)k_\ss.
\end{equation}
For a constant $\sigma_t = \sigma$, the variance-time action reduces to the action in \cref{eq:cot_constant} divided by $\sigma^2$, and the minimizing paths recover \cref{eq:optim_paths}.

When the minimum in \eqref{eq:cot_scheduled} is finite, the optimizer and the reference process have the same bridge for almost every endpoint pair under the optimal coupling \citep[Proposition~2.3]{sbsurvey}.
Since $\cH_\gamma$ leaves both endpoints fixed, applying it to the ordinary Gaussian bridge gives this shared bridge.
To obtain the posterior distribution, let $Z_t^{(0)}$ denote the ordinary Gaussian bridge obtained from \cref{eq:bridge} with $\gamma = 0$. Gaussian conditioning, as in \citet[Proposition~3.3]{i2sb}, gives
\begin{equation}\label{eq:ordinary_posterior}
  \bar{Z}_\ss^{(0)} \mid \bar{\vz}_T, \bar{\vz}_0 \sim \cC\cN\!\biggl( \bar{\vz}_T + (\bar{\vz}_0 - \bar{\vz}_T)r_\ss, \frac{\omega_\ss^2\hat{\omega}_\ss^2}{\omega_\ss^2 + \hat{\omega}_\ss^2}\vI_d \biggr).
\end{equation}
Applying the rotation in \cref{eq:cot_scheduled} to this bridge yields
\begin{equation}\label{eq:rotated_bridge}
\begin{aligned}
  Z_t &= \vz_T + r_{T - t}^{i\gamma}\bigl(Z_t^{(0)} - \vz_T\bigr), \\
  \bar{Z}_\ss &= \bar{\vz}_T + r_\ss^{-i\gamma}\bigl(\bar{Z}_\ss^{(0)} - \bar{\vz}_T\bigr).
\end{aligned}
\end{equation}
Using It\^o's formula \citep{sde} and $\frac{\dd}{\dd t}\log r_{T - t} = -\sigma_t^2/\hat{\omega}_{T - t}^2$, we obtain
\[
  \dd Z_t = -c_\gamma\frac{\sigma_t^2}{\hat{\omega}_{T - t}^2}(Z_t - \vz_T)\dd t + \sigma_t r_{T - t}^{i\gamma}\dd W_t^{(0)}.
\]
Since $\lvert r_{T - t}^{i\gamma}\rvert = 1$, the stochastic integral $W_t = \int_0^t r_{T - \tau}^{i\gamma}\dd W_\tau^{(0)}$ is a standard proper complex Wiener process. This verifies \cref{eq:bridge}; uniqueness on each compact subinterval of $[0, T)$ identifies its solution with \cref{eq:rotated_bridge}. Moreover, $\lVert Z_t - \vz_T\rVert = \lVert Z_t^{(0)} - \vz_T\rVert \to 0$ almost surely as $t \uparrow T$, which establishes the continuous endpoint extension.

The transformation in \cref{eq:rotated_bridge} is affine and preserves the isotropic variance of the Gaussian noise. The posterior therefore remains proper complex Gaussian, with mean and covariance
\begin{align*}
  \bar{\mu}_\ss &= \bar{\vz}_T + r_\ss^{-i\gamma} \bigl(\bE\bigl[\bar{Z}_\ss^{(0)}\bigr] - \bar{\vz}_T\bigr) = \bar{\vz}_T + (\bar{\vz}_0 - \bar{\vz}_T)k_\ss, \\
  \Sigma_\ss &= \lvert r_\ss^{-i\gamma}\rvert^2 \frac{\omega_\ss^2\hat{\omega}_\ss^2}{\omega_\ss^2 + \hat{\omega}_\ss^2}\vI_d = \frac{\omega_\ss^2\hat{\omega}_\ss^2}{\omega_\ss^2 + \hat{\omega}_\ss^2}\vI_d.
\end{align*}
This proves \cref{eq:posterior}. Furthermore, expanding and rearranging the terms yields the marginal mean
\[
  \bar{\mu}_\ss = k_\ss\bar{\vz}_0 + (1 - k_\ss)\bar{\vz}_T.
\]

For the discrete updates, let $0 = t_0 < \cdots < t_N = T$ with $N \ge 2$, set $s_n = T - t_n$, and evaluate $r_n$, $k_n$, $\omega_n^2$, $\hat{\omega}_n^2$, $\bar{\mu}_n$, and $\Sigma_n$ at $s_n$. Write $\bar{\vz}_n$ for a realization of $\bar{Z}_{s_n}$, so that $\bar{\vz}_N = \bar{\vz}_T$ and $\bar{\vz}_0$ remains the initial forward endpoint. All subsequent laws are conditional on this boundary pair. To sample from the posterior, we run the reverse process from step $n + 1$ to step $n$.

For the ordinary bridge, Gaussian conditioning gives \citep[Appendix~A]{i2sb}
\begin{align*}
  \bE\bigl[\bar{Z}_{s_n}^{(0)} \mid \bar{Z}_{s_{n + 1}}^{(0)}\bigr] &= \frac{\hat{\omega}_n^2 - \hat{\omega}_{n + 1}^2}{\omega_{n + 1}^2}\bar{\vz}_0 + \frac{\omega_n^2}{\omega_{n + 1}^2}\bar{Z}_{s_{n + 1}}^{(0)}, \\
  \Cov\bigl(\bar{Z}_{s_n}^{(0)} \mid \bar{Z}_{s_{n + 1}}^{(0)}\bigr) &= \frac{\omega_n^2(\hat{\omega}_n^2 - \hat{\omega}_{n + 1}^2)}{\omega_{n + 1}^2}\vI_d.
\end{align*}
For $1 \le n \le N - 2$, we substitute the inverse rotation
\[
  \bar{\vz}_{n + 1}^{(0)} = \bar{\vz}_N + r_{n + 1}^{i\gamma}(\bar{\vz}_{n + 1} - \bar{\vz}_N)
\]
and apply the rotation at step $n$. Rearranging the resulting expression yields the discrete posterior transition and its moments:
\begin{subequations}\label{eq:discrete_posterior}
\begin{align}
  q(\bar{\vz}_n \mid \bar{\vz}_{n + 1}, \bar{\vz}_0, \bar{\vz}_N) &= \cC\cN(\bar{\vz}_n \mid \bar{\mu}_{n \mid n + 1}, \Sigma_{n \mid n + 1}), \\
  \bar{\mu}_{n \mid n + 1} &= \bar{\mu}_n + \frac{\omega_n^2}{\omega_{n + 1}^2} \biggl(\frac{r_n}{r_{n + 1}}\biggr)^{-i\gamma} (\bar{\vz}_{n + 1} - \bar{\mu}_{n + 1}), \label{eq:conditional_mean}\\
  \Sigma_{n \mid n + 1} &= \frac{\omega_n^2(\hat{\omega}_n^2 - \hat{\omega}_{n + 1}^2)}{\omega_{n + 1}^2}\vI_d. \label{eq:conditional_covariance}
\end{align}
\end{subequations}
At the initial reverse step $n = N - 1$, the state is sampled directly from \cref{eq:posterior}, since $r_N = 0$ and the endpoint is fixed. At the final step $n = 0$, the distribution reduces to the point mass at $\bar{\vz}_0$.

Taking the expectation of \cref{eq:conditional_mean} and substituting the inductive hypothesis $\bE[\bar{Z}_{s_{n + 1}}] = \bar{\mu}_{n + 1}$ yields
\[
  \bE[\bar{Z}_{s_n}] = \bar{\mu}_n + \frac{\omega_n^2}{\omega_{n + 1}^2} \biggl(\frac{r_n}{r_{n + 1}}\biggr)^{-i\gamma} \bigl(\bE[\bar{Z}_{s_{n + 1}}] - \bar{\mu}_{n + 1}\bigr) = \bar{\mu}_n.
\]
For the covariance, the rotation again has unit modulus. Using
$\hat{\omega}_n^2 - \hat{\omega}_{n + 1}^2 = \omega_{n + 1}^2 - \omega_n^2$ and the fact that $\omega_n^2 + \hat{\omega}_n^2$ is constant, we have
\begin{align*}
  \Sigma_{n \mid n + 1} + \biggl(\frac{\omega_n^2}{\omega_{n + 1}^2}\biggr)^2\Sigma_{n + 1} &= \biggl[ \frac{\omega_n^2(\hat{\omega}_n^2 - \hat{\omega}_{n + 1}^2)}{\omega_{n + 1}^2} + \frac{\omega_n^4\hat{\omega}_{n + 1}^2}{\omega_{n + 1}^2(\omega_{n + 1}^2 + \hat{\omega}_{n + 1}^2)} \biggr]\vI_d\\
  &= \frac{\omega_n^2\hat{\omega}_n^2}{\omega_n^2 + \hat{\omega}_n^2}\vI_d = \Sigma_n.
\end{align*}
These match the exact marginal mean and covariance in \cref{eq:posterior}. Since Gaussian conditional distributions with affine means and constant covariances compose to a Gaussian, the reverse step guarantees by induction that the sampler reproduces the marginal distribution in \cref{eq:posterior} at each step. If $\gamma = 0$, the rotation vanishes and \cref{eq:discrete_posterior} reduces to the ordinary I$^2$SB posterior. \QED{}

%% file: text/D_detail.tex
\section{Implementation Details}\label{sec:detail}

\subsection{Mean-field simulation}
\begin{wrapfigure}[9]{r}{110pt}
  \centering
  \vskip-19pt
  \includegraphics[width=85pt]{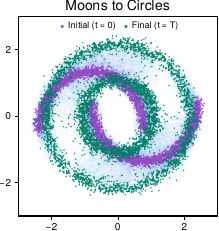}
  \vskip-3pt
  \caption{Moons to circles.}\label{fig:moons_traj}
\end{wrapfigure}
Across all experiments, we employed the AdamW optimizer with a decay factor of $10^{-2}$. SDE integration was performed via the Euler--Maruyama scheme with step size $\Delta t =0.005$, where the total time $T$ varied depending on the tasks \citep{liu2024generalized}. For the marginal distribution on momenta, we used scaled normal distributions $\vv_0 \sim \cN(\mathbf{0}, \varsigma^2 \vI_d)$, setting the scale larger than the local modal dispersion of the position distributions. Specifically, we used $\varsigma = 0.5$ for moons-circles (see \cref{fig:moons_traj} for trajectories), $\varsigma = 2.0$ for GMM-8, $\varsigma = 1.5$ for S-Tunnel, and $\varsigma = 1.0$ for V-Neck.

\textbf{TD Learning.}\enspace The temporal difference estimates $\textrm{TD}^{+}_\ss(\phi;\theta^\prime)$ and $\textrm{TD}^{-}_t(\theta;\phi^\prime)$ are approximated as
\begin{align*} 
   &i \bigl(\lVert \check{v}^\phi_\ss\rVert^2 - \gamma\sigma\nabla_{\!\vz} \!\boldsymbol{\cdot}\mkern-1.2mu \check{v}^\phi_\ss - \bigl(1\!+\!\tfrac{i\gamma}{2}\bigr)\lVert v^{\theta^\prime}_\ss\!+\!\bar{v}^\phi_\ss\rVert^2\!-i\sigma\nabla_{\!\vz} \!\boldsymbol{\cdot}\mkern-1.2mu \hat{v}^{\theta^\prime}_\ss -i\sigma\nabla_{\!\bar\vz} \!\boldsymbol{\cdot}\mkern-1.2mu v^{\theta^\prime}_\ss\bigr) \delta s +  (v^{\theta^\prime}_\ss)^\ssT \delta W_\ss + (\hat{v}^{\theta^\prime}_\ss)^\ssT \delta \bar{W}_\ss \\ 
   -&i \bigl(\lVert v^\theta_t\rVert^2 - \gamma\sigma\nabla_{\!\vz} \!\boldsymbol{\cdot}\mkern-1.2mu \hat{v}^\theta_t - \bigl(1\!+\!\tfrac{i\gamma}{2}\bigr)\lVert \hat{v}^\theta_t\!+\!\check{v}^{\phi^\prime}_t\rVert^2\!-i\sigma\nabla_{\!\vz} \!\boldsymbol{\cdot}\mkern-1.2mu \check{v}^{\phi^\prime}_t -i\sigma\nabla_{\!\bar\vz} \!\boldsymbol{\cdot}\mkern-1.2mu \bar{v}^{\phi^\prime}_t\bigr) \delta t -  (\bar{v}^{\phi^\prime}_t)^\ssT \delta W_{\!t} - (\check{v}^{\phi^\prime}_t)^\ssT \delta \bar{W}_t,
\end{align*}
where $F_\ss$ and $F_t$ can be added to the respective expressions for nonzero external MF costs \citep{deepgsb}. For base drifts, we keep the value constant or zero because the divergence terms produce side effects in TD learning.  
Raw returns are constructed from one or more TD steps, incorporating the appropriate boundary or bootstrap values. Applying the Hopf--Cole relation $\widehat{Y}=Y+i\log\rho$ in the corresponding parameterization, we transform these returns into regression targets: $\bar{G}^+_\ss=R^+_\ss+i\log\rho_\ss$ and $G^-_t=R^-_t-i\log\rho_t$, with the density evaluated at the return's starting state and time.

For each starting point, let $N\ge1$ denote the number of remaining steps to the boundary in the respective time direction, with $0\leq\lambda\leq1$. We define $R^{+}_{\ss,n}$ as the value of $Y_\theta$ at the sampled endpoint $s+n\delta s$ minus the sum of the intervening $\textrm{TD}^{+}$ increments, and define $R^{-}_{t,n}$ analogously from $\widehat{Y}_\phi$, $t+n\delta t$, and $\textrm{TD}^{-}$. The positive endpoint corresponds to physical time $T-s-n\delta s$. Applying the same conversion yields $\bar{G}^{+}_{\ss,n}=R^{+}_{\ss,n}+i\log\rho_\ss$ and $G^{-}_{t,n}=R^{-}_{t,n}-i\log\rho_t$. Following the undiscounted finite-horizon episodic $\lambda$ return formulation of $\textrm{TD}(\lambda)$ \citep{rl}, we obtain
\begin{subequations}
\begin{align}
  \bar{G}^{+,\lambda}_\ss(\phi;\theta) &= (1 - \lambda) \sum_{n=1}^{N-1} \lambda^{n-1} \bar{G}^{+}_{\ss,n}(\phi;\theta) + \lambda^{N-1} \bar{G}^{+}_{\ss,N}(\phi;\theta) \\ 
  G^{-,\lambda}_t(\theta;\phi) &= (1 - \lambda) \sum_{n=1}^{N-1} \lambda^{n-1} G^{-}_{t,n}(\theta;\phi) + \lambda^{N-1} G^{-}_{t,N}(\theta;\phi)  
\end{align}
\end{subequations}
Higher $\lambda$ was often the preferred choice due to the interplay of multiple losses and fast syncing with the EMA estimates. We set $\lambda=1$ for comparison with benchmarks. 

\textbf{Network architecture.}\enspace For the 2D control problems, we adopted the potential and drift neural network architectures from \citet{deepgsb}, augmenting both input and output layers with additional momentum coordinates. We also evaluated fully complex-valued networks with holomorphic activation functions; however, the holomorphy constraint \one{} conflicted with our core assumptions, and \two{} provided no empirical benefits to justify the added computational complexity and memory overhead. This observation likewise held for the subsequent image generation workflows.
   
\subsection{Image-to-image translation} 

\textbf{RGB.}\enspace For MNIST and CIFAR-10 data, we deployed a simplified custom NCSN$_{++}$ architecture \citep{sgmsde}\footnote{\url{https://github.com/yang-song/score_sde}} featuring two residual blocks. We adopted a symmetric parabolic diffusion schedule $\sigma_t = 4 \cdot 0.012 \cdot t (1 - t) + 10^{-5}$ for both cases, and set the initial momentum standard deviation to $0.01$, except for the CIFAR-10 inpainting task, for which we used $0.003$. For ImageNet256, modeling and evaluation relied on the official I$^2$SB repository\footnote{\url{https://github.com/NVlabs/I2SB}}. Across all settings, the network architectures remained virtually identical, varying only in their input and output channel dimensions. 

\textbf{MRI.}\enspace We simulated complex-valued brain MRI slices at 64×64 resolution using the open-source package \texttt{KomaMRI.jl} \citep{castillo2023komamri}. Given a source-target separation of 24\,mm between fixed regions of the brain, the objective was to reconstruct the correct amplitude and colorization of a subject's brain image from a source image. In total, we generated 24 synthetic subjects whose characteristics were controlled through anatomical and tissue-property perturbations, yielding 10,000 training pairs (20 subjects) for the SB models and 2,000 validation pairs (4 subjects). For this synthetic benchmark, we employed the standard NCSN$_{++}$ architecture which has a larger total parameter count than the model in the MNIST and CIFAR experiments.

\subsection{Circuit quantum electrodynamics simulation}

To simulate a closed system, we employed the open-source package \texttt{QuantumToolbox.jl} \citep{quantumtoolbox} to numerically integrate the Schr\"odinger equation for a harmonic oscillator with $d=2$ modes with high precision. The system dynamics are governed by a Hamiltonian incorporating local cavity detuning and inter-mode hopping:
\[
  \hat{H}_\mathrm{osc} = D(\hat{a}^\dagger_1 \hat{a}_1) + D(\hat{a}^\dagger_2 \hat{a}_2) + J (\hat{a}^\dagger_1 \hat{a}_2 + \hat{a}_1 \hat{a}_2^\dagger)
\]
where $\hat{a}_j$ and $\hat{a}_j^\dagger$ denote the bosonic annihilation and creation operators for mode $j\in\{1,2\}$, respectively. We set the detuning parameter to $D=10\pi$, and the inter-mode coupling strength to $J=4\pi$. The system was initialized in a pure tensor product state representing a coherent state with an intensity of $4.0$ in the first mode and a vacuum state in the second. Deterministic unitary evolution was computed over a total duration $T=4$. At each step, we extracted the exact expectation values for the position and momentum quadratures, defined respectively as $\hat{x}_i = (\hat{a}_i + \hat{a}_i^\dagger) / 2$ and $\hat{p}_i = (\hat{a}_i - \hat{a}_i^\dagger) / (2i)$, which served as our main observation inputs. For data generation, the injected Gaussian noise was non-stationary and explicitly scaled by the instantaneous quantum standard deviation of each observable, \eg{}, $\Delta x=\sqrt{\langle x^2\rangle - \langle x\rangle^2}$. We added Gaussian noise proportional to these quantum fluctuations, with a noise scalar $\xi$.

We tested tracking these time evolutions with Wirtinger gradients of a potential network and compared the potential network's accuracy over $5$ seconds of simulated time with that of the Hamiltonian neural net \citep{greydanus2019hamiltonian}, a Neural ODE \cite{node} model and a neural SDE (NFE $= 1,600$). The neural SDE model was trained using It\^o drift, whereas the other models were trained with the mean drift obtained by averaging both drifts. For training, we used $5,000$ episodes, and we evaluated using an average over $1,000$ particles. \cref{tab:harfit} shows that estimating drift with Wirtinger gradients gave us solid results in predicting the next step in a system governed by the Schr\"odinger equation. Even though the neural SDE got closer as $\xi$ diminished, it failed to capture the structure of the dynamics, as seen in \cref{fig:qt}~(b).

\begin{table}[t]
  \centering
  \caption{Fitting a harmonic oscillator trajectory from noisy quantum observations (5 seeds).}
  \label{tab:harfit}
  \begin{tabular}{lccc}
    \toprule
    \textbf{Method} & $\xi =10^{-3}$ & $\xi =10^{-2}$  & $\xi =10^{-1}$ \\
    \midrule
    NODE \citep{node} & 4027144 & 4197223 & 234321 \\
    HNN \citep{greydanus2019hamiltonian} & 8.37$_{\pm 0.55}$ & 7.79$_{\pm 1.52}$ & 9.35$_{\pm 0.77}$\\         
    Neural SDE & $1.39_{\pm 2.47}$ & $1.54_{\pm 3.19}$ & $2.92_{\pm 2.63}$ \\
    Ours (potential net) & 0.16$_{\pm 0.10}$ & 0.35$_{\pm 0.33}$ & $1.61_{\pm 1.40}$\\
    \bottomrule
  \end{tabular}
\end{table}

\textbf{3D qubit gate transform.}\enspace Building on the harmonic oscillator benchmarks, we extended our framework to simulate the dynamics of a tripartite hybrid quantum system using \texttt{QuantumToolbox.jl}. Specifically, we modeled a three-mode harmonic cavity coupled to a three-mode Josephson junction, which in turn interacted with a two-level qubit \citep{nakamura1999coherent}. Dynamics are governed by the rotating-frame Hamiltonian:
\[
  \hat{H} = \hat{H}_\mathrm{osc}+  \hat{H}_\mathrm{jj} +  \hat{H}_\mathrm{qubit} + \hat{H}_\mathrm{int}.
\]
We framed these experiments as source-to-target state mapping tasks. The system was initialized with the cavity in coherent states spanning multiple intensities (mean photon numbers in $\{0.5,1.0,2.0\}$), the JJ in the vacuum state, and the qubit in canonical states on the Bloch sphere (\eg, north pole, equator). For each configuration, deterministic unitary trajectories were integrated over a duration of $T=5.0$ across 2,001 time steps. Target states were simulated analogously after first transforming the initial qubit state through ideal unitary gates. Models were trained on 5,000 episodes and evaluated across 1,000 trajectories in a fully unpaired setting. As a baseline, we trained an MLP on the complete oscillator data using batchwise cost matching, incorporating no physical priors beyond the projection layer. For OT-CFM and SB-CFM \citep{tong2023improving}, we provided an equivalent data budget, a trainable Kerr-nonlinear Hamiltonian for the Josephson junction, and exact cost matrices for target matching. Assuming a flat space, we trained the network with the conditional CSB algorithm, using an OT solver for batchwise couplings where the distances were calculated from conformal geodesics.

\subsection{Computational costs}

Throughout the experiments, we controlled the internal neural network architectures (\eg{}, maintaining an identical number of hidden layers) to ensure a fair comparison. Under this setup, the only difference in parameter count lies in the input and output projection layers, which were adjusted to match the dimensionalities of the respective input/output specifications. This discussion concerns comparisons with fixed internal architectures; the two-qubit experiment below also examines a change in architecture. We clarify the computational costs separately for the general CSB formulation and the conditional (including Image-to-Image) CSB algorithms below:
\begin{itemize}[leftmargin=*, noitemsep, partopsep=0pt, topsep=0pt, parsep=0pt]
  \item \textbf{General CSB}
  \begin{itemize}[leftmargin=1.5em]
    \item \textbf{Forward pass and particle forward CVSDE.}\enspace While particle simulation based on the complex It\^o lemma double-counts operations over the real and imaginary components, this overhead is negligible compared to the forward pass through the non-linear neural network. Consequently, trajectory generation at inference remains highly efficient once the model is trained.
    \item \textbf{Potential backward CVSDE.}\enspace Due to the nature of automatic differentiation (vector-Jacobian products), temporal difference computations for the backward CVSDE double the gradient evaluations, while exact Laplacian calculations add non-trivial overhead.
    To mitigate the cost of second-order differentiation, gradient caching with neural networks \citep{deepgsb} can be applied.
  \end{itemize}
  \item \textbf{Conditional CSB}
  \begin{itemize}[leftmargin=1.5em]
    \item \textbf{Forward and backward updates.}\enspace In image-based setups, input/output projection differences are practically negligible. Crucially, the costly backward potential updates are replaced with analytic posterior representations, bypassing explicit Laplacian evaluations and drastically reducing sampling latency to match that of standard real-valued diffusion bridge baselines.
  \end{itemize}
\end{itemize}

%% file: text/E_expr.tex
\section{Additional Experimental Results}\label{sec:add_expr}

\subsection{Empirical validation of the analytic posterior and hyperparameter selection}
 
\textbf{Choice of the hyperparameter $\gamma$.}\enspace
The benefits of incorporating Hamiltonian dynamics into stochastic processes are well established \citep{neal2011mcmc}, and our empirical results generally confirm that our approach is consistently effective across datasets. For its validation, we conducted a grid search based on validation performance and empirically observed that a higher value of $\gamma$ is particularly effective when our theoretical assumptions explicitly hold and nonconservative biases between samples are minimal. To precisely validate this argument, let us define a stochastic interpolant \citep{albergo2025stochastic} $z_t \sim \mathbb{P}_t^\gamma$ using the analytic posterior curve parameterized by $\gamma$. The $\gamma$-interpolant is defined as
\begin{equation}
  z_t = z_1 + (z_0-z_1)(1-t)^{1+i\gamma}
  + 2\sigma\sqrt{t(1-t)}\,\hat{\xi}, \qquad t\in[0,1],
\end{equation}
where $\hat{\xi}$ denotes the interpolation noise. Now, using this construction, we considered the remaining area under the curve (rAUC), which measures how close the path measure stays to the target. For instance, the 2-Wasserstein rAUC ($W_2$-rAUC) metric is defined as
\begin{equation}
  W_2\text{-rAUC}(\tau)
  = \frac{1}{T}\int_{\tau}^{T} W_2(\mathbb{P}_t^\gamma,\mathbb{Q}_T)\,\mathrm{d}t,
\end{equation}
where $\mathbb{Q}_T$ denotes the target marginal, and $T=1$ for the normalized time interval used above. We approximated the integration by partitioning the time span into discrete intervals. 

\begin{table}
  \centering
  \caption{2-Wasserstein saturation of the analytic posterior for oscillator states of the Hadamard operation. Bold and underlined values denote the best and second-best results.} \label{tab:r3}
  \resizebox{\linewidth}{!}{
    \begin{tabular}{rcccccccc}
      \toprule
      $\gamma$ & $W_2$-rAUC(.94) $\downarrow$ & $\mathrm{MMD}^{2}$-rAUC(.94) $\downarrow$ & $W_2$-rAUC(.95) $\downarrow$ & $\mathrm{MMD}^{2}$-rAUC(.95) $\downarrow$ & $W_2$-rAUC(.97) $\downarrow$ & $\mathrm{MMD}^{2}$-rAUC(.97) $\downarrow$ & $W_2$-rAUC(.99) $\downarrow$ & $\mathrm{MMD}^{2}$-rAUC(.99) $\downarrow$ \\
      \midrule
      $0$ & $\mathbf{0.202734}$ & $\mathbf{0.051280}$ & $\mathbf{0.154899}$ & $\mathbf{0.039132}$ & $\mathbf{0.063955}$ & $\mathbf{0.015614}$ & $\underline{0.007683}$ & $\mathbf{0.001424}$ \\
      $1$ & $0.203424$ & $0.051459$ & $0.155543$ & $0.039302$ & $0.064253$ & $0.015706$ & $0.007710$ & $\underline{0.001433}$ \\
      $3$ & $0.203286$ & $0.051405$ & $0.155257$ & $\underline{0.039217}$ & $\underline{0.064030}$ & $\underline{0.015637}$ & $0.007709$ & $\underline{0.001433}$ \\
      $5$ & $0.203372$ & $0.051429$ & $\underline{0.155256}$ & $0.039224$ & $0.064083$ & $0.015658$ & $0.007723$ & $0.001437$ \\
      $10$ & $\underline{0.203020}$ & $\underline{0.051384}$ & $0.155370$ & $0.039272$ & $0.064292$ & $0.015720$ & $0.007733$ & $0.001441$ \\
      $30$ & $0.203512$ & $0.051452$ & $0.155503$ & $0.039270$ & $0.064053$ & $0.015648$ & $\mathbf{0.007682}$ & $\mathbf{0.001424}$ \\
      \bottomrule
    \end{tabular}
  }
\end{table}

In \cref{tab:r3}, we report the 2-Wasserstein rAUC metric and the squared MMD metric ($\mathrm{MMD}^{2}$-rAUC) for a representative instance in the cQED simulation task. We observed that $\gamma=0$ gives the lowest $W_2$-rAUC and $\mathrm{MMD}^{2}$-rAUC for $\tau\in\{0.94,0.95,0.97\}$ among the tested values. At $\tau=0.99$, $\gamma=30$ gives the lowest $W_2$-rAUC, while $\gamma=0$ and $\gamma=30$ tie for the lowest $\mathrm{MMD}^{2}$-rAUC at the reported precision. We also report FID scores (noise-to-image) in \cref{tab:fid-gamma} with similar tendencies and different scales. As a result, the optimal choice of $\gamma$ depends largely on data characteristics and the underlying relationship between marginal distributions.  

\begin{table}
  \centering
  \caption{CIFAR-10 posterior FID scores for different values of $\gamma$.}
  \label{tab:fid-gamma}
  \begin{tabular}{rrrrr}
    \toprule
    $\gamma$ & FID at $0.9$ & FID at $0.95$ & FID at $0.99$ & FID at $0.995$ \\
    \midrule
    $0$ & $31.316002$ & $\underline{2.152040}$ & $\underline{0.081899}$ & $0.094650$ \\
    $0.0001$ & $\underline{31.315914}$ & $2.152060$ & $0.081906$ & $0.094652$ \\
    $0.001$ & $\mathbf{31.315853}$ & $\mathbf{2.151930}$ & $0.081904$ & $\underline{0.094637}$ \\
    $0.01$ & $31.321283$ & $2.152645$ & $\mathbf{0.081816}$ & $\mathbf{0.094604}$ \\
    $0.1$ & $31.520388$ & $2.184362$ & $0.081928$ & $0.094841$ \\
    $1$ & $47.182999$ & $5.063271$ & $0.094534$ & $0.100584$ \\
    $3$ & $97.487869$ & $23.163961$ & $0.109886$ & $0.110451$ \\
    $9$ & $208.948288$ & $75.324211$ & $0.225691$ & $0.114714$ \\
    \bottomrule
  \end{tabular}
\end{table}

\subsection{Two-dimensional control benchmarks}

Since $\bC^d\cong\bR^{2d}$, there is no mathematical distinction between using the space $\bC^{d}$ and using an unconstrained model in $\bR^{2d}$ per se.\footnote{Nonetheless, this work introduces a compact presentation of the dynamics for a coherent geometric prior and an inductive bias for dealing with coupled position-momentum and magnitude-phase dynamics with useful complex operations.} Therefore, the ablation study was conducted by augmenting the state space of DeepGSB \citep{deepgsb} to $\bR^{2d}$, achieving the same dimensionality as CSB but without the Hamiltonian prior. As shown in Tables~\ref{tab:r1}~and~\ref{tab:r2}, CSB shows substantial improvement over the DeepGSB formulation while being competitive with GSBM \citep{liu2024generalized} in terms of endpoint matching. Notably, CSB obtained the lowest total trajectory costs (both kinetic and potential) on all three tasks, verifying our claim about higher-order dynamics for diffusion bridges.

\begingroup
\captionsetup{hypcap=false}
\begin{table}[H]
  \centering
  \begin{minipage}[t]{0.49\linewidth}
  \centering
  \caption{2-Wasserstein 2D control benchmark results (forward direction, 5 different seeds).}
  \label{tab:r1}
  \resizebox*{!}{2cm}{
  \begin{tabular}{lrrr}
    \toprule
    \textbf{Method} & GMM-8 & S-Tunnel & V-Neck \\
    \midrule
    DeepGSB ($\mathbb{R}^{2d}$) & $3.92 \pm 2.15$ & $0.56 \pm 0.04$ & $0.81 \pm 0.03$ \\
    GSBM ($\mathbb{R}^d$) & $2.65 \pm 0.41$ & $\mathbf{0.17 \pm 0.03}$ & $\mathbf{0.09 \pm 0.01}$ \\
    CSB ($\gamma=0.001$, $\mathbb{C}^d$) & $\underline{1.92 \pm 0.83}$ & $0.21 \pm 0.07$ & $0.61 \pm 0.03$ \\
    CSB ($\gamma=0.01$, $\mathbb{C}^d$) & $\mathbf{1.91 \pm 0.25}$ & $\underline{0.18 \pm 0.05}$ & $0.63 \pm 0.04$ \\
    CSB ($\gamma=0.1$, $\mathbb{C}^d$) & $1.99 \pm 0.77$ & $0.20 \pm 0.08$ & $0.65 \pm 0.04$ \\
    CSB ($\gamma=1$, $\mathbb{C}^d$) & $3.67 \pm 1.92$ & $0.23 \pm 0.06$ & $\underline{0.52 \pm 0.19}$ \\
    \bottomrule
  \end{tabular}
  }
  \end{minipage}\hfill
  \begin{minipage}[t]{0.49\linewidth}
  \centering
  \caption{Total trajectory cost calculated from a sum of kinetic energy and obstacle costs (5 runs).}
  \label{tab:r2}
  \resizebox*{!}{2cm}{
  \begin{tabular}{lrrr}
    \toprule
    \textbf{Method} & GMM-8 & S-Tunnel & V-Neck \\
    \midrule
    DeepGSB ($\mathbb{R}^{2d}$) & $228.86 \pm 0.61$ & $510.16 \pm 4.93$ & $12.00 \pm 1.87$ \\
    GSBM ($\mathbb{R}^d$) & $121.48 \pm 14.40$ & $484.98 \pm 2.30$ & $144.72 \pm 0.37$ \\
    CSB ($\gamma=0.001$, $\mathbb{C}^d$) & $116.06 \pm 0.32$ & $\mathbf{295.66 \pm 36.64}$ & $\underline{8.08 \pm 0.68}$ \\
    CSB ($\gamma=0.01$, $\mathbb{C}^d$) & $\underline{115.97 \pm 0.23}$ & $\underline{300.97 \pm 46.00}$ & $8.14 \pm 0.56$ \\
    CSB ($\gamma=0.1$, $\mathbb{C}^d$) & $\mathbf{115.64 \pm 0.23}$ & $303.22 \pm 41.43$ & $8.10 \pm 0.73$ \\
    CSB ($\gamma=1$, $\mathbb{C}^d$) & $118.80 \pm 1.28$ & $306.32 \pm 45.83$ & $\mathbf{4.00 \pm 2.78}$ \\
    \bottomrule
  \end{tabular}
  }
  \end{minipage}
\end{table}
\endgroup

\subsection{Two-qubit gate experiments}

\begin{wraptable}[9]{r}{0.45\textwidth}
  \vskip-8pt
  \centering
  \caption{Qubit fidelity of CNOT and SWAP 2-qubit logic gates.}
  \vskip-3pt
  \label{tab:r4}
  \begin{tabular}{lrr}
    \toprule
     & CNOT & SWAP \\
    \midrule
    OT-CFM & $0.9993$ & $0.9448$ \\
    SB-CFM & $0.9062$ & $0.9991$ \\
    Ours ($\gamma=1$) & $\mathbf{0.9994}$ & $\mathbf{0.9998}$ \\
    \bottomrule
  \end{tabular}
\end{wraptable}

Building upon our single-qubit experiment setups, we expanded our evaluation to two-qubit gates by comparing a 4-layer MLP against a 3-block residual network, representing a roughly fourfold increase in raw trainable parameters (total 800K). Notably, we applied weak pairing to faithfully replicate the output of each gate's truth table, and this indicates a single neural network handled the four input cases in each two-qubit gate's truth table. Then, we measured \textit{qubit fidelity}, a measure of accuracy and reliability that quantifies how closely an actual quantum state or operation aligns with an ideal target state or operation. The results are summarized in Table~\ref{tab:r4}. In both quantum logic gates (CNOT and SWAP operations), our approach showed stable performance and near-perfect transformation fidelity in conditional SB learning and also demonstrated scalability. The result demonstrates that our OT application for 3D qubits is scalable to larger cQED applications.

\subsection{RGB image translation}

\begin{wraptable}[8]{r}{0.25\textwidth}
  \vskip-10pt
  \caption{MNIST FID.} \label{tab:mnifid}
  \vskip-3pt
  \centering
  \begin{tikzpicture}[tight background]
    \node[scale=0.7] at (0,0) {
      \begin{tabular}{lc}
        \toprule
        Method & FID ($\downarrow$) \\
        \midrule
        I$^2$SB & 3.7557 \\
        Ours ($\gamma=0.0001$) & {2.7138} \\
        Ours ($\gamma=0.001$) & {2.881} \\
        Ours ($\gamma=0.01$) & \textbf{2.4791} \\
        Ours ($\gamma=0.1$) & \underline{2.5536}\\
        Ours ($\gamma=1.0$) & {17.0554}\\
        \bottomrule
      \end{tabular}
    };
  \end{tikzpicture}
\end{wraptable}

In \cref{tab:mnifid}, the FID trends for unpaired MNIST generation mirror those observed for CIFAR-10 in \cref{fig:rgb}. Because this behavior recurred across image benchmarks, we conclude that our assumption regarding Hamiltonian structures holds broadly, even for static visual scenes. \cref{fig:mnicifar_additional,fig:img256_additional} show qualitative image-to-image translation results, illustrating how each dataset-specific model configuration contributes to FID improvements. On MNIST, gains are driven by increased mode diversity, whereas on higher-dimensional images, improvements stem primarily from contextual fidelity, followed by sharper object contours.

\subsection{Brain MRI image translation}

MRI preserves both the amplitude and the phase of an oscillating signal; complex numbers are a convenient way to retain both pieces of information. We conducted experiments on a complex-valued brain MRI dataset created with the open-source package \texttt{KomaMRI.jl} \citep{castillo2023komamri} and comprising 24 simulated subjects generated through controlled anatomical and tissue-property perturbations. As shown in Table~\ref{tab:r5}, rotational dynamics improved the transport of complex MRI images across both real and imaginary signal components, consistent with our theoretical assumptions. Furthermore, all four tested values of $\gamma$ achieved lower FID and higher magnitude SSIM than I$^2$SB, with $\gamma=0.1$ yielding the best performance across both metrics.

\begin{figure}[p]
  \centering
  \makebox[\linewidth][c]{
  \begin{tikzpicture}[x=1cm,y=1cm,tight background,
      every node/.style={inner sep=0cm,outer sep=0cm}]
    \foreach \method/\col/\title in {
        source/0/{Source},
        dsbm/1/{DSBM},
        sbflow/2/{SB Flow},
        i2sb/3/{I$^2$SB},
        {csb/gamma_0.0001}/4/{{\font\mrigreek=gsmn1000 at 10pt\mrigreek g} = 10\textsuperscript{-4}},
        {csb/gamma_0.001}/5/{{\font\mrigreek=gsmn1000 at 10pt\mrigreek g} = 10\textsuperscript{-3}},
        {csb/gamma_0.01}/6/{{\font\mrigreek=gsmn1000 at 10pt\mrigreek g} = 10\textsuperscript{-2}},
        {csb/gamma_0.1}/7/{{\font\mrigreek=gsmn1000 at 10pt\mrigreek g} = 0.1},
        {csb/gamma_1}/8/{{\font\mrigreek=gsmn1000 at 10pt\mrigreek g} = 1},
        target/9/{Target}} {
      \node[scale=.5,anchor=base] at (0.647+1.354*\col,0.08) {\textsf{\title}};
      \foreach \sample/\top in {0085/0,1006/-5.636,1991/-11.272} {
        \foreach \view/\row in {magnitude/0,phase/1,real/2,imag/3} {
          \node[anchor=north west] at (1.354*\col,\top-1.354*\row)
            {\includegraphics[width=1.294cm]{fig/mri/\method/sample_\sample_\view.png}};
        }
      }
    }
    \node[scale=.7,anchor=base] at (8.771,0.40) {\textsf{Ours}};
    \draw[line width=0.012cm] (5.456,0.35) -- (12.086,0.35);
    \foreach \top in {0,-5.636,-11.272} {
      \foreach \label/\row in {$|z|$/0,$\arg z$/1,\textsf{real}/2,\textsf{imag.}/3} {
        \node[scale=.5,rotate=90,anchor=south] at (-0.05,\top-0.647-1.354*\row) {\label};
      }
    }
    \node[scale=.43,anchor=east] at (1.27,-16.928) {$|z|$};
    \node[anchor=west] at (1.42,-16.928)
      {\includegraphics[width=1.85cm,height=0.12cm]{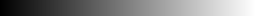}};
    \node[scale=.43,anchor=north] at (1.42,-17.038) {$0$};
    \node[scale=.43,anchor=north] at (3.27,-17.038) {$1$};
    \node[scale=.43,anchor=east] at (5.59,-16.928) {$\arg z$};
    \node[anchor=west] at (5.74,-16.928)
      {\includegraphics[width=1.85cm,height=0.12cm]{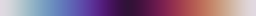}};
    \node[scale=.43,anchor=north] at (5.74,-17.038) {$-\pi$};
    \node[scale=.43,anchor=north] at (7.59,-17.038) {$\pi$};
    \node[scale=.43,anchor=east] at (10.02,-16.928) {$\Re z,\Im z$};
    \node[anchor=west] at (10.17,-16.928)
      {\includegraphics[width=1.85cm,height=0.12cm]{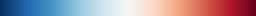}};
    \node[scale=.43,anchor=north] at (10.17,-17.038) {$-1$};
    \node[scale=.43,anchor=north] at (12.02,-17.038) {$1$};
    \path let \p1=(current bounding box.west) in
      ([xshift=-\x1]current bounding box.east);
  \end{tikzpicture}
  }
  \caption{Complex MRI slice-transport results for fixed validation
    examples. Each block shows magnitude, phase, real, and
    imaginary components with shared display ranges.} \label{fig:mri_additional}
\end{figure}

\begin{wraptable}{r}{0.46\textwidth}
  \vskip-15pt
  \centering
  \caption{Magnitude FID and SSIM metrics on the paired $64 \times 64$ synthetic brain MRI validation dataset for unseen subjects (source-target slice separation: 24 mm).} \label{tab:r5}
  \begin{tabular}{lrr}
    \toprule
    Method & FID $\downarrow$ & SSIM $\uparrow$ \\
    \midrule
    DSBM & $23.1730$ & $0.6126$ \\
    SB Flow & $21.4061$ & $0.6312$ \\ 
    I$^2$SB & $15.2626$ & $0.6597$ \\
    Ours ($\gamma=0.001$) & $12.0784$ & $0.7130$ \\
    Ours ($\gamma=0.01$) & $12.4734$ & $0.7104$ \\
    Ours ($\gamma=0.1$) & $\mathbf{12.0135}$ & $\mathbf{0.7139}$ \\
    Ours ($\gamma=1.0$) & $12.8268$ & $0.7129$ \\
    \bottomrule
  \end{tabular}
\end{wraptable}
For qualitative evaluation on the MRI dataset, \cref{fig:mri_additional} displays image generation results across three distinct subjects for DSBM \citep{dsbm}, SB Flow \citep{de2024schrodinger}, I$^2$SB \citep{i2sb}, and our method. In this comparison, CSB recovered plausible target images in terms of shape, magnitude, and angle. We observed that the improvements in image quality were more pronounced when handling explicit complex-valued data than when preprocessing RGB images with zero-centered momenta. Combined with the quantitative evaluation on the validation set in \cref{tab:r5}, these results on the complex-valued dataset demonstrated the broader applicability of our theoretical framework to scientific data.

\clearpage

\begingroup
\captionsetup{hypcap=false}
\begin{figure}[p]
  \centering
  \makebox[\linewidth][c]{
  \begin{tikzpicture}[x=1cm,y=1cm,tight background,
      every node/.style={inner sep=0pt,outer sep=0pt}]
    \node[scale=.65,anchor=base] at (8.23,0.55) {\textsf{I$^2$CSB}};
    \draw[line width=0.012cm] (2.62,0.44) -- (13.84,0.44);
    \foreach \file/\col/\title in {
        i2sb/0/{I$^2$SB},
        i2csb_gamma_0.0001/1/{$\gamma=10^{-4}$},
        i2csb_gamma_0.001/2/{$\gamma=10^{-3}$},
        i2csb_gamma_0.01/3/{$\gamma=10^{-2}$},
        i2csb_gamma_0.1/4/{$\gamma=10^{-1}$},
        i2csb_gamma_1.0/5/{$\gamma=1$}} {
      \node[scale=.57,anchor=base] at (1.45+2.26*\col,0.13) {\textsf{\title}};
      \node[anchor=north west] at (0.36+2.26*\col,0)
        {\includegraphics[width=2.18cm]{fig/mnist_additional/\file.png}};
      \node[anchor=north west] at (0.36+2.26*\col,-2.30)
        {\includegraphics[width=2.18cm]{fig/cifar_additional/\file.png}};
    }
    \node[scale=.58,rotate=90] at (0.12,-1.09) {\textsf{MNIST}};
    \node[scale=.58,rotate=90] at (0.12,-3.39) {\textsf{CIFAR-10 inpainting}};
    \path ([xshift=9.4bp]current bounding box.east);
  \end{tikzpicture}
  }
  \caption{MNIST and CIFAR-10 image-to-image translation} \label{fig:mnicifar_additional}
\par\vspace{12pt}
  \makebox[\linewidth][c]{
  \begin{tikzpicture}[x=1cm,y=1cm,tight background,
      every node/.style={inner sep=0pt,outer sep=0pt}]
    \foreach \title/\block in {Center/0,FreeForm1020/1,FreeForm2030/2} {
      \node[scale=.75,rotate=90,anchor=south] at (-0.12,-2.22-5.1525*\block) {\textsf{\title}};
    }
    \foreach \mask/\block/\instance/\row/\pair/\g in {
        center/0/1/0/0/{10^{-1}},
        center/0/2/0/1/{10^{-4}},
        center/0/3/1/0/{10^{-2}},
        center/0/4/1/1/{10^{-2}},
        ff1020/1/1/0/0/{10^{-2}},
        ff1020/1/2/0/1/{10^{-1}},
        ff1020/1/3/1/0/{10^{-2}},
        ff1020/1/4/1/1/{10^{-1}},
        ff2030/2/1/0/0/{10^{-1}},
        ff2030/2/2/0/1/{10^{-1}},
        ff2030/2/3/1/0/{10^{-1}},
        ff2030/2/4/1/1/{10^{-4}}} {
      \foreach \view/\col/\title in {
          source/0/{Source},
          reconstruction/1/{Ours},
          target/2/{Target}} {
        \ifnum\row=0
          \node[scale=.57,anchor=base] at (1.09+6.78*\pair+2.26*\col,0.12-5.1525*\block)
            {\textsf{\title}};
        \fi
        \node[anchor=north west] at (6.78*\pair+2.26*\col,-5.1525*\block-2.26*\row)
          {\includegraphics[width=2.18cm]{fig/imgnet_additional/\mask/example_\instance/\view.png}};
      }
    }
    \path ([xshift=9.05bp]current bounding box.east);
  \end{tikzpicture}
  }
  \caption{Image recovery from three different types of corruption on the ImageNet256 validation dataset.}
  \label{fig:img256_additional}
\end{figure}
\endgroup

\clearpage

